\documentclass[10pt,twocolumn,letterpaper]{article}

\usepackage{iccv}
\usepackage{times}
\usepackage{epsfig}
\usepackage{graphicx}
\usepackage{amsmath}
\usepackage{amssymb}
\usepackage[accsupp]{axessibility}

\usepackage{bm}
\usepackage{bbding}
\usepackage{tabularx}
\usepackage{enumitem}
\usepackage{float}
\usepackage{booktabs}
\usepackage[noend]{algpseudocode}
\usepackage{algorithmicx,algorithm}
\usepackage[usenames,dvipsnames]{color}
\definecolor{mygray}{gray}{.9}
\usepackage{colortbl}
\usepackage[dvipsnames]{xcolor}
\usepackage{eso-pic}

\usepackage[breaklinks=true,bookmarks=false]{hyperref}

\iccvfinalcopy 

\def\iccvPaperID{****} 

\ificcvfinal\fi

\begin{document}
\def\mA{\mathcal{A}}
\def\mB{\mathcal{B}}
\def\mC{\mathcal{C}}
\def\mD{\mathcal{D}}
\def\mE{\mathcal{E}}
\def\mF{\mathcal{F}}
\def\mG{\mathcal{G}}
\def\mH{\mathcal{H}}
\def\mI{\mathcal{I}}
\def\mJ{\mathcal{J}}
\def\mK{\mathcal{K}}
\def\mL{\mathcal{L}}
\def\mM{\mathcal{M}}
\def\mN{\mathcal{N}}
\def\mO{\mathcal{O}}
\def\mP{\mathcal{P}}
\def\mQ{\mathcal{Q}}
\def\mR{\mathcal{R}}
\def\mS{\mathcal{S}}
\def\mT{\mathcal{T}}
\def\mU{\mathcal{U}}
\def\mV{\mathcal{V}}
\def\mW{\mathcal{W}}
\def\mX{\mathcal{X}}
\def\mY{\mathcal{Y}}
\def\mZ{\mathcal{Z}} 

\def\bbN{\mathbb{N}} 
\def\bbR{\mathbb{R}} 
\def\bbP{\mathbb{P}} 
\def\bbQ{\mathbb{Q}} 
\def\bbE{\mathbb{E}}

\def\1n{\mathbf{1}_n}
\def\0{\mathbf{0}}
\def\1{\mathbf{1}}

\def\A{{\bf A}}
\def\B{{\bf B}}
\def\C{{\bf C}}
\def\D{{\bf D}}
\def\E{{\bf E}}
\def\F{{\bf F}}
\def\G{{\bf G}}
\def\H{{\bf H}}
\def\I{{\bf I}}
\def\J{{\bf J}}
\def\K{{\bf K}}
\def\L{{\bf L}}
\def\M{{\bf M}}
\def\N{{\bf N}}
\def\O{{\bf O}}
\def\P{{\bf P}}
\def\Q{{\bf Q}}
\def\R{{\bf R}}
\def\S{{\bf S}}
\def\T{{\bf T}}
\def\U{{\bf U}}
\def\V{{\bf V}}
\def\W{{\bf W}}
\def\X{{\bf X}}
\def\Y{{\bf Y}}
\def\Z{{\bf Z}}

\def\a{{\bf a}}
\def\b{{\bf b}}
\def\c{{\bf c}}
\def\d{{\bf d}}
\def\e{{\bf e}}
\def\f{{\bf f}}
\def\g{{\bf g}}
\def\h{{\bf h}}
\def\i{{\bf i}}
\def\j{{\bf j}}
\def\k{{\bf k}}
\def\l{{\bf l}}
\def\m{{\bf m}}
\def\n{{\bf n}}
\def\o{{\bf o}}
\def\p{{\bf p}}
\def\q{{\bf q}}
\def\r{{\bf r}}
\def\s{{\bf s}}
\def\t{{\bf t}}
\def\u{{\bf u}}
\def\v{{\bf v}}
\def\w{{\bf w}}
\def\x{{\bf x}}
\def\y{{\bf y}}
\def\z{{\bf z}}

\def\balpha{\mbox{\boldmath{$\alpha$}}}
\def\bbeta{\mbox{\boldmath{$\beta$}}}
\def\bdelta{\mbox{\boldmath{$\delta$}}}
\def\bgamma{\mbox{\boldmath{$\gamma$}}}
\def\blambda{\mbox{\boldmath{$\lambda$}}}
\def\bsigma{\mbox{\boldmath{$\sigma$}}}
\def\btheta{\mbox{\boldmath{$\theta$}}}
\def\bomega{\mbox{\boldmath{$\omega$}}}
\def\bxi{\mbox{\boldmath{$\xi$}}}
\def\bnu{\mbox{\boldmath{$\nu$}}}                                  
\def\bphi{\mbox{\boldmath{$\phi$}}}
\def\bmu{\mbox{\boldmath{$\mu$}}}

\def\bDelta{\mbox{\boldmath{$\Delta$}}}
\def\bOmega{\mbox{\boldmath{$\Omega$}}}
\def\bPhi{\mbox{\boldmath{$\Phi$}}}
\def\bLambda{\mbox{\boldmath{$\Lambda$}}}
\def\bSigma{\mbox{\boldmath{$\Sigma$}}}
\def\bGamma{\mbox{\boldmath{$\Gamma$}}}
                                  
\newcommand{\myprob}[1]{\mathop{\mathbb{P}}_{#1}}

\newcommand{\myexp}[1]{\mathop{\mathbb{E}}_{#1}}

\newcommand{\mydelta}[1]{1_{#1}}

\newcommand{\myminimum}[1]{\mathop{\textrm{minimum}}_{#1}}
\newcommand{\mymaximum}[1]{\mathop{\textrm{maximum}}_{#1}}    
\newcommand{\mymin}[1]{\mathop{\textrm{minimize}}_{#1}}
\newcommand{\mymax}[1]{\mathop{\textrm{maximize}}_{#1}}
\newcommand{\mymins}[1]{\mathop{\textrm{min.}}_{#1}}
\newcommand{\mymaxs}[1]{\mathop{\textrm{max.}}_{#1}}  
\newcommand{\myargmin}[1]{\mathop{\textrm{argmin}}_{#1}} 
\newcommand{\myargmax}[1]{\mathop{\textrm{argmax}}_{#1}} 
\newcommand{\myst}{\textrm{s.t. }}

\newcommand{\denselist}{\itemsep -1pt}
\newcommand{\sparselist}{\itemsep 1pt}

\definecolor{pink}{rgb}{0.9,0.5,0.5}
\definecolor{purple}{rgb}{0.5, 0.4, 0.8}   
\definecolor{gray}{rgb}{0.3, 0.3, 0.3}
\definecolor{mygreen}{rgb}{0.2, 0.6, 0.2}

\newcommand{\cyan}[1]{\textcolor{cyan}{#1}}
\newcommand{\red}[1]{\textcolor{red}{#1}}  
\newcommand{\blue}[1]{\textcolor{blue}{#1}}
\newcommand{\magenta}[1]{\textcolor{magenta}{#1}}
\newcommand{\pink}[1]{\textcolor{pink}{#1}}
\newcommand{\green}[1]{\textcolor{green}{#1}} 
\newcommand{\gray}[1]{\textcolor{gray}{#1}}    
\newcommand{\mygreen}[1]{\textcolor{mygreen}{#1}}    
\newcommand{\purple}[1]{\textcolor{purple}{#1}}       

\definecolor{greena}{rgb}{0.4, 0.5, 0.1}
\newcommand{\greena}[1]{\textcolor{greena}{#1}}

\definecolor{bluea}{rgb}{0, 0.4, 0.6}
\newcommand{\bluea}[1]{\textcolor{bluea}{#1}}
\definecolor{reda}{rgb}{0.6, 0.2, 0.1}
\newcommand{\reda}[1]{\textcolor{reda}{#1}}

\def\changemargin#1#2{\list{}{\rightmargin#2\leftmargin#1}\item[]}
\let\endchangemargin=\endlist
                                               
\newcommand{\cm}[1]{}

\newcommand{\mhoai}[1]{{\color{magenta}\textbf{[MH: #1]}}}
\newcommand{\yifeng}[1]{{\color{red}\textbf{[yifeng: #1]}}}
\newcommand{\yftodo}[1]{{\color{blue}$\blacksquare$\textbf{[TODO: #1]}}}

\newcommand{\mtodo}[1]{{\color{red}$\blacksquare$\textbf{[TODO: #1]}}}
\newcommand{\myheading}[1]{\vspace{1ex}\noindent \textbf{#1}}
\newcommand{\htimesw}[2]{\mbox{$#1$$\times$$#2$}}


\newif\ifshowsolution
\showsolutiontrue

\ifshowsolution  
\newcommand{\Solution}[2]{\paragraph{\bf $\bigstar $ SOLUTION:} {\sf #2} }
\newcommand{\Mistake}[2]{\paragraph{\bf $\blacksquare$ COMMON MISTAKE #1:} {\sf #2} \bigskip}
\else
\newcommand{\Solution}[2]{\vspace{#1}}
\fi

\newcommand{\truefalse}{
\begin{enumerate}
	\item True
	\item False
\end{enumerate}
}

\newcommand{\yesno}{
\begin{enumerate}
	\item Yes
	\item No
\end{enumerate}
}

\newcommand{\Sref}[1]{Sec.~\ref{#1}}
\newcommand{\Eref}[1]{Eq.~(\ref{#1})}
\newcommand{\Fref}[1]{Fig.~\ref{#1}}
\newcommand{\Tref}[1]{Table~\ref{#1}}

\title{Interactive Class-Agnostic Object Counting}

\author{
Yifeng Huang\textsuperscript{1},\quad Viresh Ranjan\textsuperscript{2}\thanks{Work done at Stony Brook University, prior to joining Amazon},\quad Minh Hoai\textsuperscript{1,3}\\
\textsuperscript{1}Stony Brook University, Stony Brook, NY, USA \\  \textsuperscript{2}Amazon, USA \quad 
\textsuperscript{3}VinAI Research, Hanoi, Vietnam
}

\maketitle
\ificcvfinal\thispagestyle{empty}\fi

\begin{abstract}
   We propose a novel framework for interactive class-agnostic object counting, where a human user can interactively provide feedback to improve the accuracy of a  counter. Our framework consists of two main components: a user-friendly visualizer to gather feedback and an efficient mechanism to incorporate it. In each iteration, we produce a density map to show the current prediction result, and we segment it into non-overlapping regions with an easily verifiable number of objects. The user can provide feedback by selecting a region with obvious counting errors and specifying the range for the estimated number of objects within it. To improve the counting result, we develop a novel adaptation loss to force the visual counter to output the predicted count within the user-specified range. For effective and efficient adaptation, we propose a refinement module that can be used with any density-based visual counter, and only the parameters in the refinement module will be updated during adaptation. Our experiments on two challenging class-agnostic object counting benchmarks, FSCD-LVIS and \mbox{FSC-147}, show that our method can reduce the mean absolute error of multiple state-of-the-art visual counters by roughly 30\% to 40\% with minimal user input. Our project can be found at \href{https://yifehuang97.github.io/ICACountProjectPage/}{https://yifehuang97.github.io/ICACountProjectPage/}.
\end{abstract}

\section{Introduction}
\label{sec:INTRO}

\begin{figure}[t]
  \centering
  \includegraphics[width=0.95\linewidth]{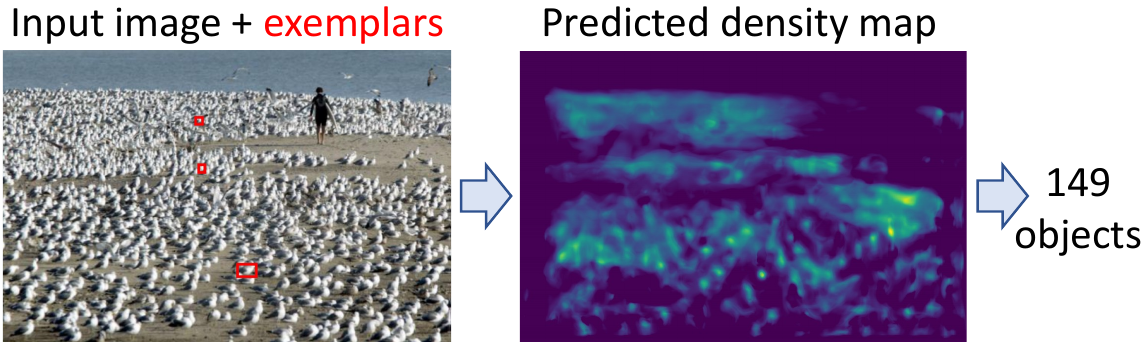}
  \caption{
  Given an input image and several exemplar objects, a class-agnostic counter will output a density map and the total object count. It is often challenging to validate these outputs, making it difficult to adopt automatic visual counting in practice.  To improve the practicality of a visual counter, we propose an interactive framework that allows a human user to quickly detect mistakes and improve performance based on the identified errors.
  }
  \label{fig:TEASER_FIGURE}
  \vspace{-0.6cm}
\end{figure}

The need for counting objects in images arises in many applications, and significant progress has been made for both class-specific \cite{li2018csrnet,miao2020shallow,jiang2020attention,wang2020DMCount,xu2019learn,hu2020count,bai2020adaptive,liu2020adaptive,song2021rethinking,liu2019point,lian2019density,sam2020locate,liu2019recurrent,laradji2018blobs,m_Ranjan-etal-ECCV18,m_Ranjan-etal-ACCV20,m_Abousamra-etal-AAAI21} and class-agnostic~\cite{yang2021class, ranjan2021learning, ranjan2022vicinal, shi2022represent, Zhiyuan2023safecount, lu2018class, nguyen2022few, m_Ranjan-Hoai-ACCV22} counting. However, unlike in many other computer vision tasks  where the predicted results can be verified for reliability, visual counting results are difficult to validate, as illustrated in \Fref{fig:TEASER_FIGURE}. Mistakes can be made, and often there are no mechanisms to correct them. To enhance the practicality of visual counting methods, the results need to be more intuitive and verifiable, and feedback mechanisms should be incorporated to allow errors to be corrected. This necessitates a human-in-the-loop framework that can interactively display the predicted results, collect user feedback, and adapt the visual counter to reduce counting errors. 

It is, however, challenging to develop an interactive framework for visual counting. The first challenge is to provide the user with an intuitive visualizer for the counting result. Current state-of-the-art visual counting methods typically generate a density map and then sum the density values to obtain the final count. However, as shown in \Fref{fig:TEASER_FIGURE}, verifying the final predicted count can be difficult, as can verifying the intermediate density map, due to the mismatch between the continuous nature of the density map and the discrete nature of the objects in the image. The second challenge is to design an appropriate user interaction method that requires minimal user effort while being suited for providing feedback on object counting. The third challenge is developing an effective adaptation scheme for the selected interaction type that can incorporate user feedback and improve the performance of visual counters. In this paper, we address all three aforementioned challenges to develop an interactive framework for visual counting. 

For the first challenge, we propose a novel segmentation method that segments a density map into non-overlapping regions, where the sum of density values in each region is a near-integer value that can be easily verified. This provides the user with a more natural and understandable interpretation of the predicted density map. Notably, developing such an algorithm that must also be suitably fast for an interactive system is challenging, which constitutes a technical contribution of our paper.

For the second challenge, we propose a novel type of interaction that enables the user to provide feedback with just two mouse clicks: the first click selects the region, and the second click selects the appropriate range for the number of objects in the chosen region. The proposed user interaction method is unique as it is specifically tailored for object counting and requires minimal user effort. Firstly, the auto-generated segmentation map allows the user to select an image region using just one mouse click, which is faster compared to drawing a polygon or scribbles. Secondly, by leveraging the humans' subitizing ability, which allows them to estimate the number of objects in a set quickly without counting them individually, we can obtain an approximate count with just another mouse click, which is quicker than one by one counting using dot annotations. 

For the third challenge, we develop an interactive adaptation loss based on range constraints. To update the visual counter efficiently and effectively and to reduce the disruption of the learned knowledge in the visual counter, we propose the refinement module that directly refines the spatial similarity feature in the regression head. Furthermore, we propose a technique to estimate the user's feedback confidence and use this confidence to adjust the learning rate and gradient steps during the adaptation process.

In this paper, we primarily focus on class-agnostic counting, and we demonstrate the effectiveness of our framework with experiments on FSC-147~\cite{ranjan2021learning} and FSCD-LVIS~\cite{nguyen2022few}. However, our framework can be extended to category-specific counting, as will be seen in our experiments on several crowd-counting and car-counting benchmarks, including ShanghaiTech~\cite{zhang2016single}, UCF-QNRF~\cite{idrees2018composition}, and CARPK~\cite{hsieh2017drone}. We also conduct a user study to investigate the practicality of our method in a real-world setting.

In short, the main contribution of our paper is a framework that improves the accuracy and practicality of visual counting. Our technical contributions include: (1) a novel segmentation method that quickly segments density maps into non-overlapping regions with near-integer density values, which enhances the interpretability of predicted density maps for users; (2) an innovative user feedback scheme that requires minimal user effort for object counting by utilizing subitizing ability and auto-generated segmentation maps; and (3) an effective adaptation approach that incorporates the user's feedback into the visual counter through a refinement module and a confidence estimation method.

\section{Related Works}
\label{sec:RELWORKS}
\myheading{Visual counting.} Various visual counting methods have been proposed, e.g.,~\cite{li2018csrnet,miao2020shallow,jiang2020attention,wang2020DMCount,xu2019learn,hu2020count,bai2020adaptive,liu2020adaptive,song2021rethinking,liu2019point,lian2019density,sam2020locate,liu2019recurrent,laradji2018blobs}, but most of them are class-specific counters, requiring large amounts of training data with hundreds of thousands of annotated objects. To address this limitation and enable counting of objects across multiple categories, several class-agnostic counters have been proposed~\cite{yang2021class, ranjan2021learning, ranjan2022vicinal, shi2022represent, Zhiyuan2023safecount, lu2018class, nguyen2022few}. These methods  work by regressing the object density map based on the spatial correlation between the input image and the provided exemplars. However, in many cases, a limited number of exemplars are insufficient to generalize over object instances with varying shapes, sizes, and appearances. 

\myheading{Interactive counting.} There exists only one prior method for interactive counting~\cite{arteta2014interactive}. This method uses low-level features and ridge regression to predict the density map. To visualize the density map, it uses MSER~\cite{matas2004robust} or spectral clustering~\cite{shi2000normalized} to generate some candidate regions, then seeks a subset of candidate regions that can keep the integrality of each region in the subset.  At each iteration, the user must draw a region of interest and mark the objects in this region with dot annotations. Additionally for the first iteration, the user has to specify the diameter of a typical object.  This method~\cite{arteta2014interactive} has two drawbacks. First, it requires significant effort from the user to draw a region of interest, specify the typical object size, and mark all objects in the region.  Second, MSER and spectral clustering may not generate suitable candidate regions for dense scenes, as will be shown in \Sref{sec:segalgo}. To alleviate the user's burden, the counting results should be easy to verify, and feedback should be simple to provide.
In this paper, we propose a density map visualization method that can generate regions by finding and expanding density peaks. Unlike MSER and spectral clustering, our approach works well on dense density maps. 

\myheading{Interactive methods for other computer vision tasks.} Various interactive methods have been developed for other computer vision tasks, such as object detection~\cite{yao2012interactive}, tracking~\cite{shen2015interactive}, and segmentation~\cite{jang2019interactive,sofiiuk2020f,gabbur2021probabilistic,lin2020interactive,liew2017regional,hu2019fully,majumder2019content,mahadevan2018iteratively,li2018interactive,liew2021deep,xu2016deep,song2018seednet}. While the success of these methods is inspiring, none of them are directly applicable to visual counting due to unique technical challenges. Unlike object detection, tracking, and segmentation, the immediate and final outputs of visual counting are difficult to visualize and verify. Designing an interactive framework for visual counting requires addressing the technical challenges discussed in the introduction section, none of them has been considered in previous interactive methods.

\begin{figure}[t]
  \centering
  \includegraphics[width=1.0\linewidth]{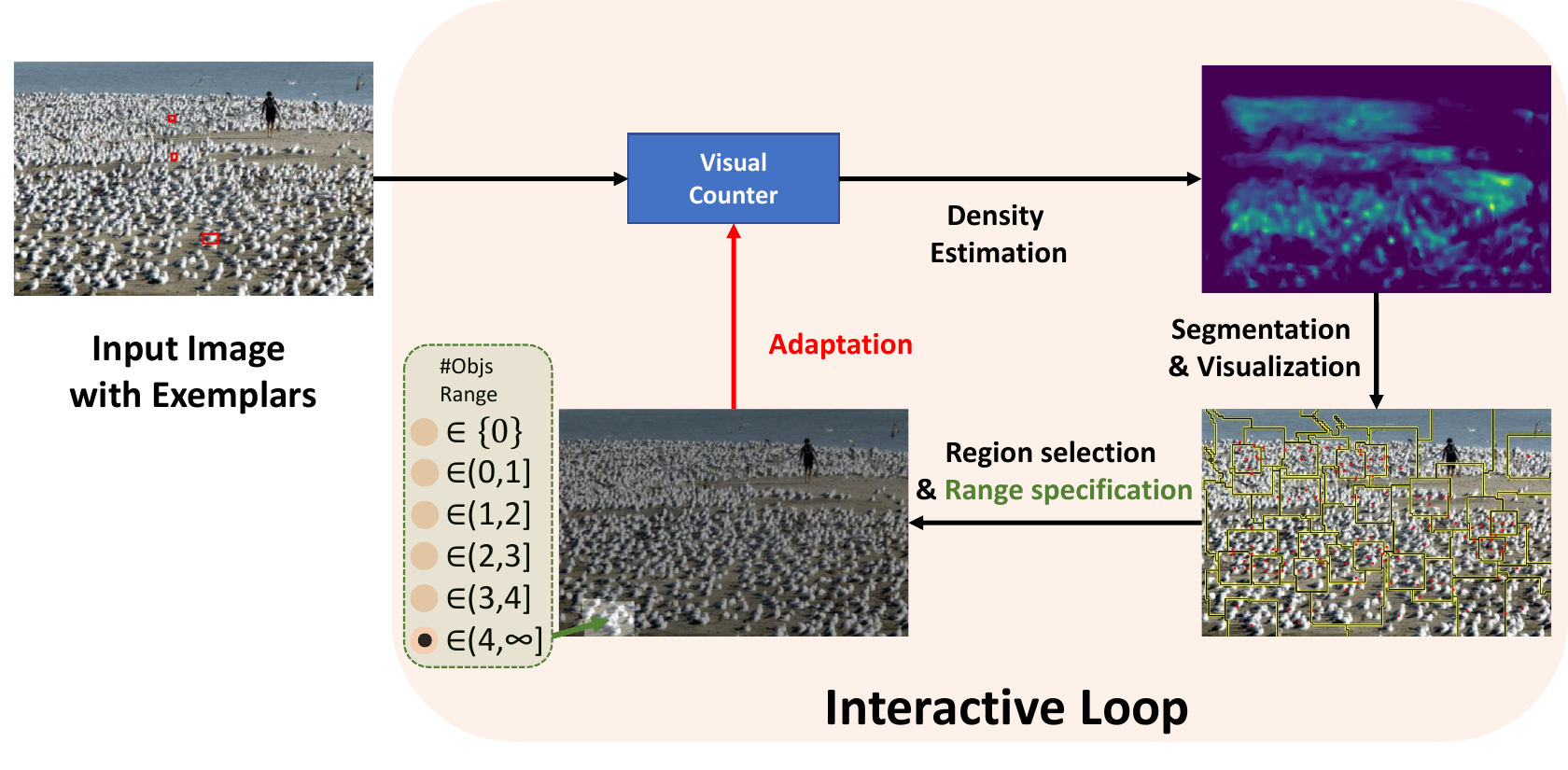}
  \caption{We propose a practical approach for visual counting based on interactive user's feedback. In each iteration: (1) the visual counter estimates the density map for the input image; (2) the density map is segmented and visualized; (3) the user selects a region and provides a range for the number of objects in the region; (4) an objective function is defined based for the provided region and count range, and the parameters of a refinement module are updated by optimizing this objective function. }
  \label{fig:INTER_LOOP}
\end{figure}

\section{Proposed Approach}
\label{sec:APPROACH}
We propose an interactive framework for visual counting, as illustrated in \Fref{fig:INTER_LOOP}. Each interactive iteration consists of two phases. The first phase visualizes the predicted density map to collect user feedback. The second phase uses the provided feedback to improve the visual counter.

\subsection{Overview of the two phases \label{sec:overview}}
In the first phase, we will visualize the density map by segmenting it into  regions $\{R_1, \cdots, R_n\}$ with the following desiderata:  
\begin{enumerate}[label=C\arabic*.] \denselist 
    \item Non-overlapping: $R_{i} \cap R_{j} = \emptyset$ for all $i \neq j$,
    \item Total coverage:  $\mathop{\cup}\limits_{i=1}^n{R_i} = $ the predicted density map,
    \item Moderate size: each region is not too big or too small, 
    \item Near-integer and small integral: the sum of density values within each region should be close to an integer and smaller than a verifiable counting limit. 
\end{enumerate}
The above desiderata are for visualization and easy verification of the results. The last desideratum is motivated by humans' subitizing ability, which is the ability to identify the number of objects in an image simply by quickly looking at them, not by counting them one by one. 

In the second phase of each iteration, the user is prompted to pick a region and specify the range for the number of objects in that region. Let $R$ denote the selected region and $c = (c_l, c_u]$ the range specified by the user for the number of objects in $R$,  a loss $\mathcal{L}(R, c)$ will be generated and used to adapt the counting model. For efficient and effective adaptation, instead of adapting the whole counting network, we propose a refinement module that directly refines the feature map in the regression head and we only adapt the parameters of  this module using gradient descent.

\subsection{Density map segmentation algorithm \label{sec:segalgo} }
One technical contribution of this paper is the development of a fast segmentation algorithm called Iterative Peak Selection and Expansion (IPSE) that satisfies the desiderata described in \Sref{sec:overview}. The input to this algorithm is a smoothed density map, and the output is a set of non-overlapping regions. IPSE is an iterative algorithm where the output set will be grown one region at a time, starting from an empty set. To yield a new region for the output, it starts from the pixel $p$ with the highest density value (called the peak) among the remaining pixels that have not been included in any previously chosen region. IPSE seeks a region $R$ containing $p$ that minimizes the below objective:
\begin{align}
 h(R) = & \frac{|R_s - \lceil R_s - \frac{1}{2} \rceil|}{\max(1, \lceil R_s - \frac{1}{2} \rceil)} + \frac{\max(0, T_{l} - R_a)}{T_{l}} \nonumber \\
 & + \lceil \max(0, R_s - C) \rceil, 
 \label{ObjectiveFunction}
\end{align}
where $R_s$ denote the sum of density values in $R$ and $R_a$ the area of $R$. $T_{l}$ is the preferred minimum area for the region $R$, and $C$ is the preferred maximum number of objects in the region. The first term of the objective function encourages the sum of the densities to be close to an integer. It also encourages the region not to have too small count. The second term penalizes small regions. The region cannot be too big either. The expansion algorithm will stop when the area reaches a predefined upper bound. The last term penalizes regions with a total density greater than $C$.

Because there are exponentially many regions containing a given peak $p$, finding the optimal region $R$ that minimizes $h(R)$ is an intractable problem. Fortunately, we only need to obtain a sufficiently good solution. We therefore restrict the search space to a smaller list of expanding regions $S_0 \subset S_1 \subset \cdots \subset S_m$ and perform an exhaustive search among this list. This list can be constructed by starting from the seed $p$, i.e., $S_0 = \{p\}$ and constructing $S_{i+1}$ from $S_i$ by adding a pixel $q$ selected from neighboring pixels of $S_i$ that have not been part of any existing output regions. If multiple such neighboring pixels are available, we prioritize pixels with positive density values and select the one closest to $p$. The process terminates when any of the following conditions is met: (1) all neighboring pixels of $S_i$ have been included in an output region; (2) the area or the sum of density values in $S_i$ has reached a predefined limit; or (3) the proportion of zero-density pixels in $S_i$ has exceeded a predefined threshold.

The above peak selection and expansion process  is used repeatedly to segment the density map, with each iteration commencing from the seed pixel $p$ that has the highest density value among the remaining pixels that have not been included in any prior output region. If all the remaining pixels have zero-density values, a random seed location is selected. The process continues until complete coverage of the density map is achieved, at which point all small regions are merged into their neighboring regions.

\myheading{Comparison with other density segmentation methods.} 
Our algorithm for segmenting the density map bares some similarity to gerrymandering or political redistricting. 

Gerrymandering involves dividing a large region into several smaller regions while adhering to certain constraints related to the population and contiguity of the regions.
Most methods for this problem are based on weighted graph partitioning or heuristic algorithms~\cite{gliesch2020heuristic,rincon2017comparative,kim2018multiobjective,forman2003congressional,goderbauer2018political}. However, an object density map contains several hundred thousand pixels, making these methods too slow for interactive systems. For example, the time and iteration limit of~\cite{gliesch2020heuristic} is 600 seconds and 1000 iterations, which cannot meet the real-time requirements of interactive systems. In contrast, our method takes less than one second, as reported in~\Sref{sec:user_study}.

Another approach for visualizing a density map is to use MSER~\cite{matas2004robust} or spectral clustering\cite{shi2000normalized} to generate some candidate regions, as used in \cite{arteta2014interactive}. MSER and spectral clustering, however, often fail to generate suitable candidate regions for dense scenes, as shown in \Fref{fig:VISComparison}.

\subsection{Interactive feedback and adaptation}
Upon presenting the segmented density map, the user will be prompted to pick a region $R$ and choose a numeric range $c$ for the number of objects in $R$, from a list of range options, $c \in \{(-\infty, 0], (0, r], (r, 2r],\ldots, (C-r, C], (C, \infty)\}$, where $r$ is the range interval and $C$ is the counting limit. This method of user interaction is innovative and specifically tailored for object counting, necessitating only two mouse clicks per iteration. The reason for using a range instead of an exact number is because it can be ambiguous to determine an exact number for a region. Despite our segmentation efforts to ensure that each region contains an integral number of objects, some regions may still contain partial objects, making it more challenging for the user to provide an accurate number quickly.

An important technical contribution of our paper is the creation of an adaptation technique capable of leveraging the user's weak supervision feedback. This feedback is weak in terms of both object localization and count. Specifically, it does not provide exact locations of object instances, only indicating the number of objects present in an image region. Additionally, the number of objects is provided only as a range, rather than an exact count. Below, we will detail our adaptation technique, starting from the novel refinement module for effective and efficient adaptation. 

\def\subFigSz{0.48\linewidth}
\begin{figure}[t] 
\centering

\makebox[\subFigSz]{\footnotesize{MSER~\cite{arteta2014interactive}}} \hfill 
\makebox[\subFigSz]{\footnotesize{IPSE (Proposed)}} 

\includegraphics[width=\subFigSz]{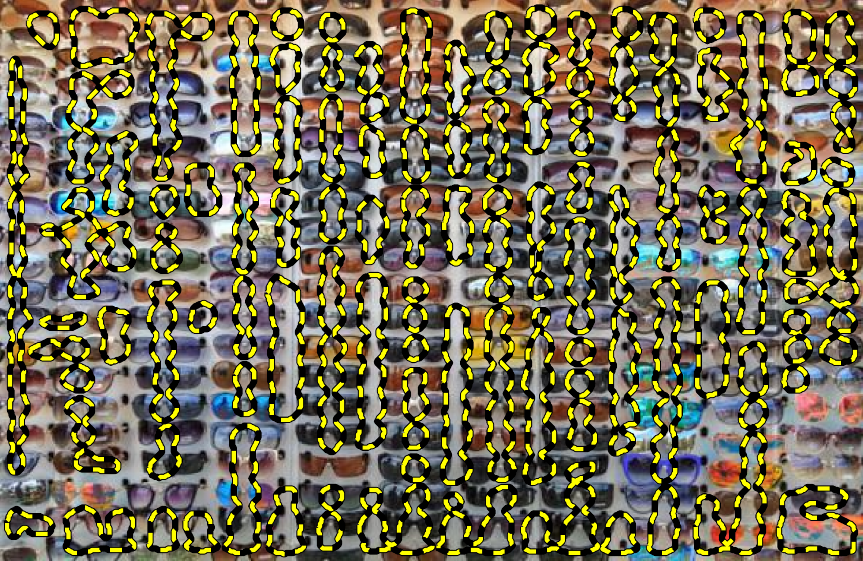} \hfill
\includegraphics[width=\subFigSz]{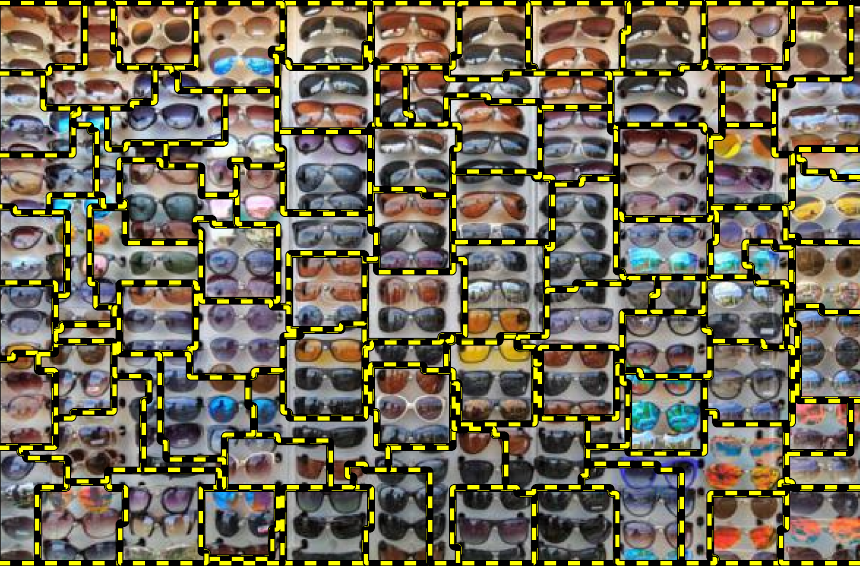}\\
\includegraphics[width=\subFigSz]{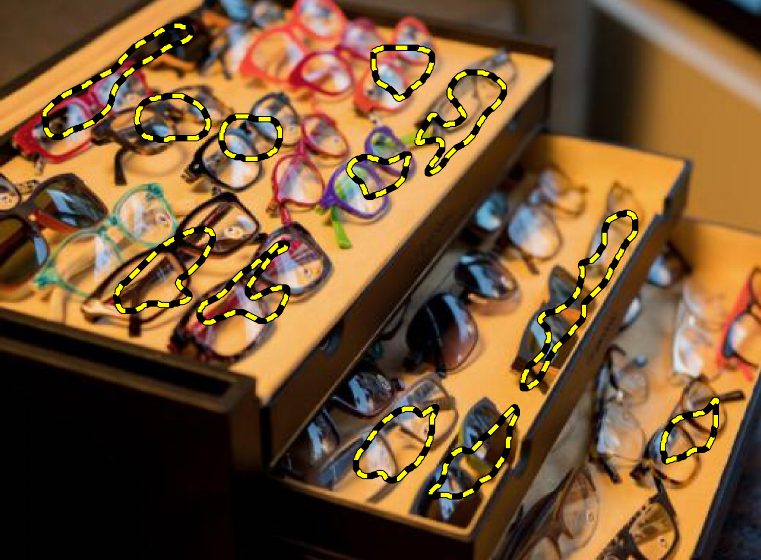} \hfill
\includegraphics[width=\subFigSz]
{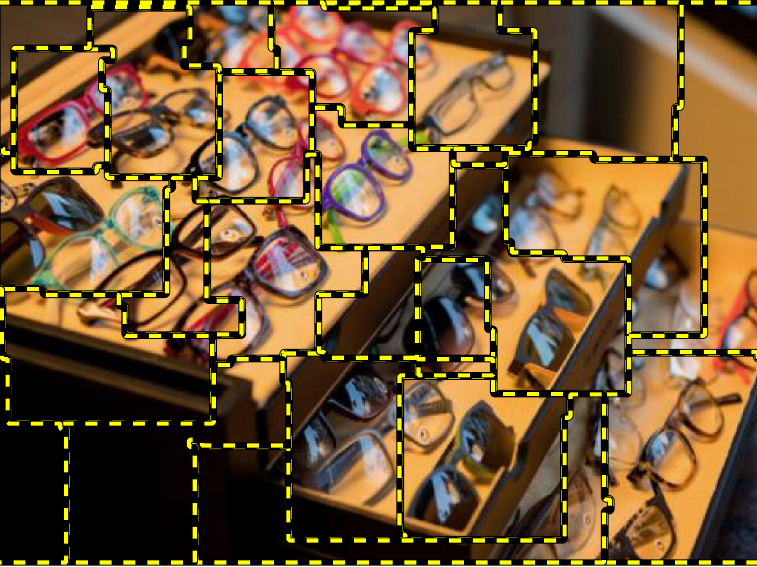} 
\caption{Density map segmentation comparison with MSER~\cite{arteta2014interactive}. These two examples are from the FSC-147 dataset. \label{fig:VISComparison}}
\end{figure}

\subsubsection{Refinement module}

We aim for an adaptation method that works for all class-agnostic counting networks~\cite {yang2021class, ranjan2021learning, ranjan2022vicinal, shi2022represent, Zhiyuan2023safecount}. Most of them contain three components: a feature extractor $f$, a spatial-similarity module $g$ (e.g., convolution~\cite{ranjan2021learning} or learnable bilinear transform~\cite{shi2022represent}), and a regression head $h$ consisting of upsampling layers and convolution layers. Let $\I$ be the input image, $\E$ the exemplars, $\S = g(f(\I), f(\E))$ the correlation map for the spatial similarity between the input image and exemplars, and $\D = h(\bm{S})$ the predicted density map. The predicted count is obtained by summing over the predicted density map $\D$. 

The correlation map serves as input to the regression head, which applies several convolution and upsampling layers to generate the output object density map. 
We observe that if the correlation map or any intermediate feature map between the input and output maps accurately represents the spatial similarity between the input image and exemplar objects, the final output density map and the predicted count will be correct. Therefore, the adaptation process need not revert to layers earlier than the correlation map. To minimize disruption to learned knowledge and accelerate adaptation for user interaction, we propose a lightweight refinement module that is integrated only within the regression head.

\begin{figure*}[t]
  \centering
  \includegraphics[width=0.95\linewidth]{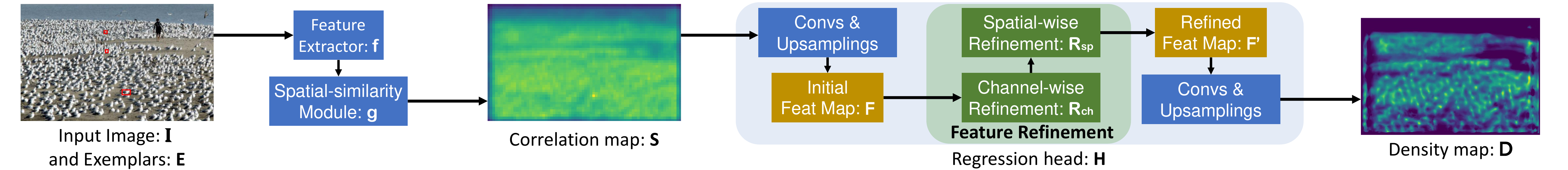}
  \caption{The refinement module can be integrated into the \textbf{regression head} of any density estimation visual counter.}
  \label{FRM_Framework}
\end{figure*}

The refinement module, depicted in \Fref{FRM_Framework}, can be applied to any intermediate feature map $\F$ between the input correlation map $\S$ and the output density map $\D$: $\bm{F^\prime} = \mR(\bm{F})$, where $\bm{F^\prime}$ is the refined feature map. Our refinement module consists of two components: channel-wise refinement and spatial-wise refinement.

\begin{figure}[t]
  \centering
  \includegraphics[width=0.9\linewidth]{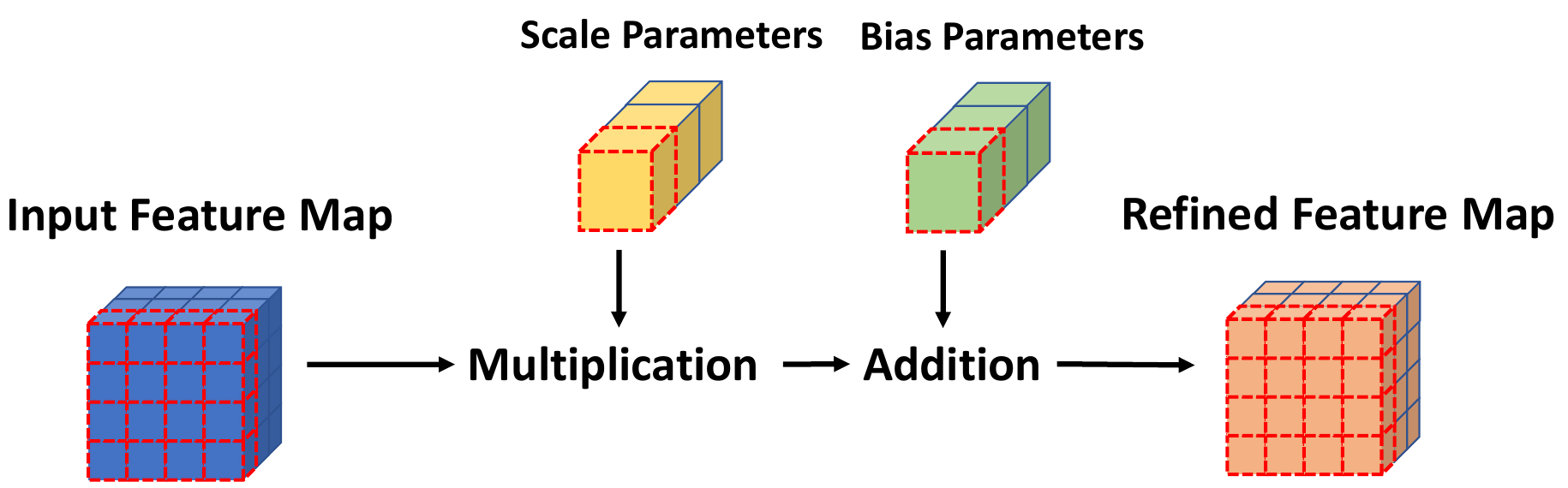}
  \caption{Channel-wise refinement refines the feature of each channel by multiplying with scale and adding with a bias.}
  \label{FRM_CH}
\end{figure}

\myheading{The channel-wise refinement} is illustrated in \Fref{FRM_CH}, and it refines the feature of each channel by multiplying with a scale parameter and adding with a bias parameter: ${\mR_{ch}}(\bm{F}) = \theta_{ch}^{scale} \odot \bm{F} + \theta_{ch}^{bias}$, where $\bm{F}\in\mathbb{R}^{H\times W\times C}$ is the input feature map, $\theta_{ch}^{scale}\in\mathbb{R}^C$ is the vector of scale parameters, $\theta_{ch}^{bias}\in\mathbb{R}^C$ is the vector of bias parameters. 

\begin{figure}[t]
  \centering
  \includegraphics[width=0.9\linewidth]{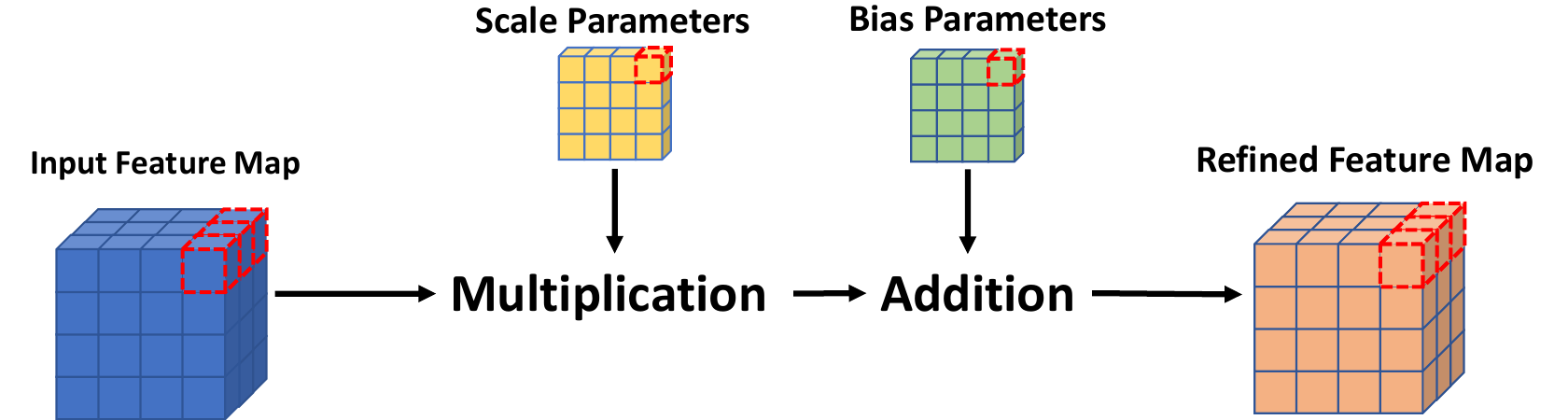}
  \caption{Spatial-wise refinement refines the feature at each spatial position by multiplying with scale and adding with a bias.}
  \label{FRM_SP}
\end{figure}

\myheading{The spatial-wise refinement} is illustrated in \Fref{FRM_SP}, and it also refines the feature at each spatial position: ${\mR_{sp}}(\bm{F}) = \theta_{sp}^{scale} \odot \bm{F} + \theta_{sp}^{bias}$, where  $\theta_{sp}^{scale}\in\mathbb{R}^{H\times W}$, $\theta_{sp}^{bias}\in\mathbb{R}^{H\times W}$. 

\myheading{The overall refinement module} is the successive application of channel-wise refinement and spatial-wise refinement:  $\bm{F^\prime} = \mR(\bm{F}) = \mR_{sp}(\mR_{ch}(\bm{F}))$. The set of adaptable parameters of the two refinement modules are: 
$\theta^{scale} =[ \theta_{ch}^{scale}; \theta_{sp}^{scale}]$ and $\theta^{bias} = [ \theta^{bias}_{ch}; \theta^{bias}_{sp}]$. 
At the beginning of each adaptation iteration, the scale parameters are reset to one and the bias parameters to zero, so that the refined feature map $\F^\prime$ is the same as the input feature map $\F$. 

\subsubsection{Adaptation schemes}
Given a selected region $R$ and a specified range $c=(c_l, c_u]$ for the number of objects in region $R$, a weekly-supervised adaptation loss is defined as: 
\begin{equation}
    \label{BasicInteractiveLoss}
    \mathcal{L}_I(R, c) = ReLU(c_l- R_s) + ReLU(R_s - c_u),
\end{equation}
If the sum of predicted density values in the selected region is outside the count range provided by the user, the above loss  will be positive. 

To account for the scenario where the user can provide feedback for multiple regions either in a single iteration or through multiple iterations, we extend the adaptation loss to use multiple regions to update the counter. Let $\Omega = {(R, c)}$ 
denote the user's selected regions and their corresponding specified count ranges. We use the following combined loss for adaptation:

\begin{equation}
\mathcal{L}(\Omega) = \mathcal{L}_L(\Omega) + \mathcal{L}_G(\Omega) + \eta(||\theta^{scale} - 1|| + ||\theta^{bias}||), \label{eqn:loss}
\end{equation}
where $\mathcal{L}_L(\Omega)$ is the sum of regional losses, with each loss corresponding to an individual region separately: 

\begin{equation}
\mathcal{L}_L(\Omega) = \sum_{(R,c) \in \Omega} \mathcal{L}_I(R, c), \label{scheme1}
\end{equation}
and $\mathcal{L}_G(\Omega)$ is the single loss for all the regions combined: 

\begin{equation}
    \label{GlobalInteractiveLoss}
    \mathcal{L}_G(\Omega) = \mathcal{L}_I(\sum_{(R, c) \in \Omega } R_s, \sum_{(R,c) \in \Omega}{c}).
\end{equation}
Here, we combine all the selected regions and view them as one big region, and then use \Eref{BasicInteractiveLoss} on the big region to adapt the visual counter. Hereafter, we will refer to $\mathcal{L}_L(\Omega)$ as Local Loss and $\mathcal{L}_G(\Omega)$ as Global Loss. 

The last term of \Eref{eqn:loss} is a regularization term to discourage large changes to the scale and bias parameters of the refinement module and $\eta$ is the weight for the regularization term. In our experiments,  $\eta = 0.002$. 

Our adaptation loss is based on the sum of predicted density values within one or multiple regions, which provides a weak supervision signal as it lacks penalty terms related to individual values. This type of supervision signal can be problematic if a large learning rate is used. Meanwhile, using a small learning rate would require numerous gradient steps, leading to a prolonged convergence process. To overcome this issue, we propose determining the impact value of the user's feedback and using it to adjust the learning rate and gradient steps, resulting in a smoother adaptation process. When the uncertainty level is higher, such as having too few regions or inconsistent error types (i.e., under-counting in some regions and over-counting in others), we use a lower learning rate with more gradient steps. Conversely, we apply a larger learning rate with fewer gradient steps to shorten the adaptation process when the uncertainty level is lower. We define the user feedback value as follows:
\begin{equation}
F_C(\Omega) = 0.5 F_i(\Omega) + 0.5 F_s(\Omega),
\end{equation}
The first term of the above measures the informativeness of the feedback while the second term measures the inconsistency of the feedback for multiple regions. The informativeness term can be defined as: 
\begin{equation}
F_i(\Omega) = \min(1, \exp(\frac{|\Omega| - t}{T})),
\end{equation}
where $|\Omega|$ is the size of the set $\Omega$, $T$ is the temperature, and $t$ is the informativeness threshold. Specifically, in our experiments $t=3, T=2$. 

Let $\Omega_o$ and $\Omega_u$ be the sets of over-counting and under-counting regions. Let $p = \frac{|\Omega_o|}{|\Omega_o| + |\Omega_u|}$, the feedback consistency value is defined based on negative entropy: 
\begin{align}
F_s(\Omega) = 1 + p\log p + (1-p) \log (1 - p ).
\end{align}

Based on the estimated value of the feedback, the learning rate and the number of gradient steps will be scaled accordingly as follows: $\gamma^\prime = \gamma F_C(\Omega)$, $N^\prime = \frac{N}{F_C(\Omega)}$, where $\gamma$ and $N$ are the default values for the learning rate and the number of gradient steps.

\section{Experiments}
\label{sec:Exp}
\subsection{Class-agnostic counting \label{sec:cac}} 
\subsubsection{Experiment settings}

\myheading{Datasets.} We evaluate our approach on two challenging class-agnostic counting benchmarks: FSC-147~\cite{ranjan2021learning} and FSCD-LVIS~\cite{nguyen2022few}.

\myheading{Class-agnostic visual counters.} Our interactive framework is applicable to many different class-agnostic counters, and we experiment with  FamNet~\cite{ranjan2021learning}, SAFECount~\cite{Zhiyuan2023safecount}, and BMNet+~\cite{shi2022represent} in this paper.

\myheading{Evaluation metrics.} We use Mean Absolute Error (MAE) and Root Mean Squared Error (RMSE) as performance metrics, which are widely used for evaluating visual counting performance~\cite{yang2021class, ranjan2021learning, ranjan2022vicinal, shi2022represent, Zhiyuan2023safecount, lu2018class, nguyen2022few}.

\myheading{User feedback simulation.} 
To simulate user feedback, we randomly select a displayed region and provide the counting range for that region. We repeat each experiment five times with different random seeds (5, 10, 15, 20, 25) and report the \textbf{average} and \textbf{standard error}.

\myheading{Count limit and pre-defined ranges.} According to~\cite{trick1994small}, people can quickly and correctly estimate the number of objects without one-to-one counting if the number of objects is less or equal to four. Therefore, we set the count limit $C$ to 4 and the pre-defined ranges to $\{(-\infty, 0], (0,1], (1,2], (2,3], (3,4], (4, \infty)\}$.

\myheading{Implementation details.} We insert the refinement module after the \textbf{first convolution layer} in the regression head. For faster computation, we first downsample the density map by a factor of four, then perform the IPSE, and finally upsample the density map segmentation result to its original size. We adapt a FamNet with an Adam optimizer~\cite{kingma2014adam}. On FSC-147, the default learning rate is $0.02$ while the default number of gradient steps is 10. On FSCD-LVIS, FamNet does not converge as well, so the default learning rate and number of gradient steps are set to $0.01$ and $20$, respectively.

\subsubsection{Experiment results}

The proposed interactive framework can be used to improve the performance of various types of visual counters, including FamNet~\cite{ranjan2021learning}, SAFECount~\cite{Zhiyuan2023safecount}, and BMNet+~\cite{shi2022represent}. As can be seen from \Fref{fig:CONSECUTIVE_CLICKS}, the benefits are consistently observed in multiple experiments and metrics (three visual counters, two datasets, and two performance metrics). Significant error reduction is already achieved even after one feedback iteration, as also shown in \Fref{fig:QUAL}. After five iterations, the amount of error reduction is huge, with an average value of 30\%. 

\def\subFigSz{0.48\linewidth}
\begin{figure}[t] 
\centering
\makebox[\subFigSz]{\footnotesize{MAE on FSC-147 test set}} \hfill 
\makebox[\subFigSz]{\footnotesize{RMSE on FSC-147 test set}} 
\includegraphics[width=\subFigSz]{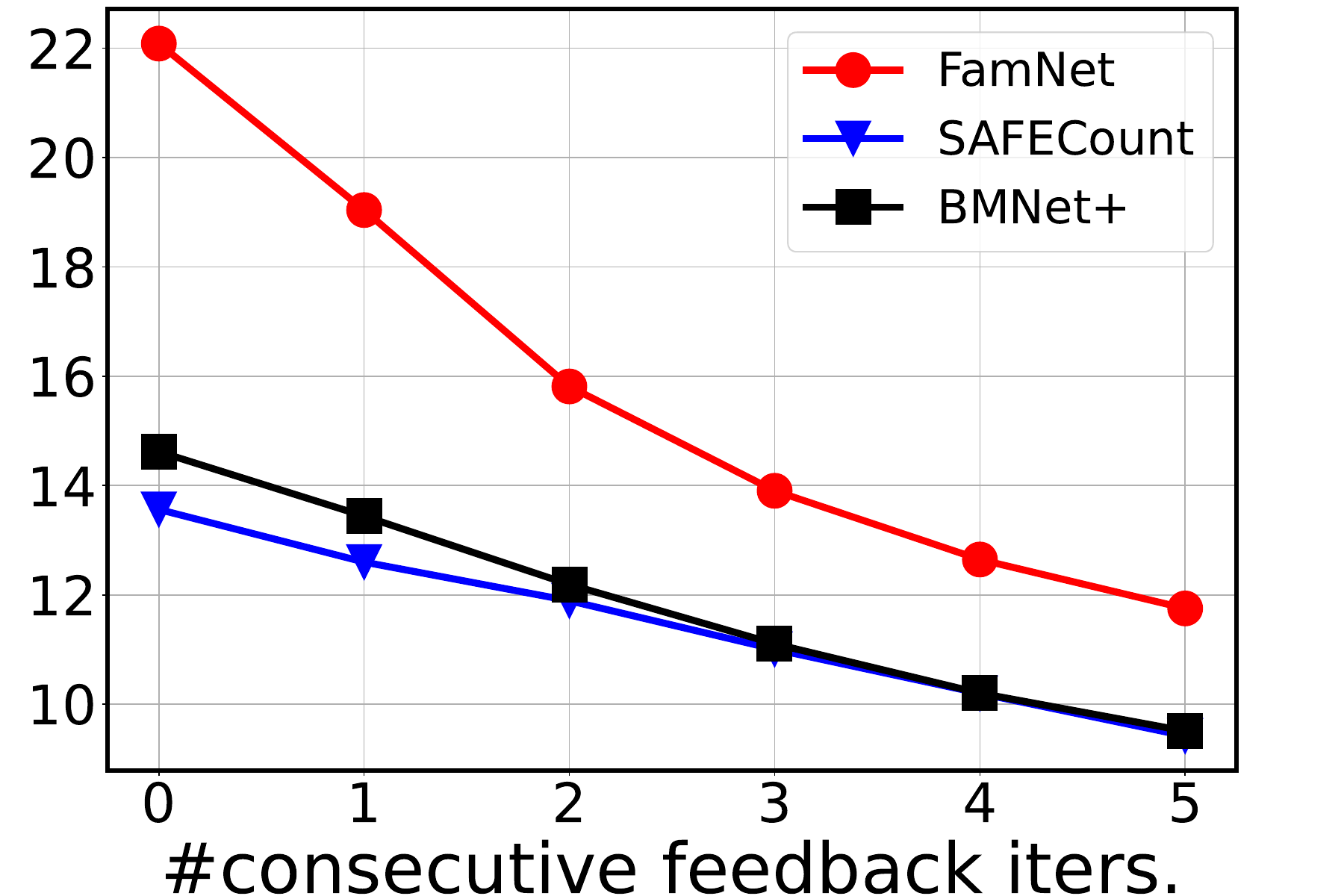} \hfill
\includegraphics[width=\subFigSz]{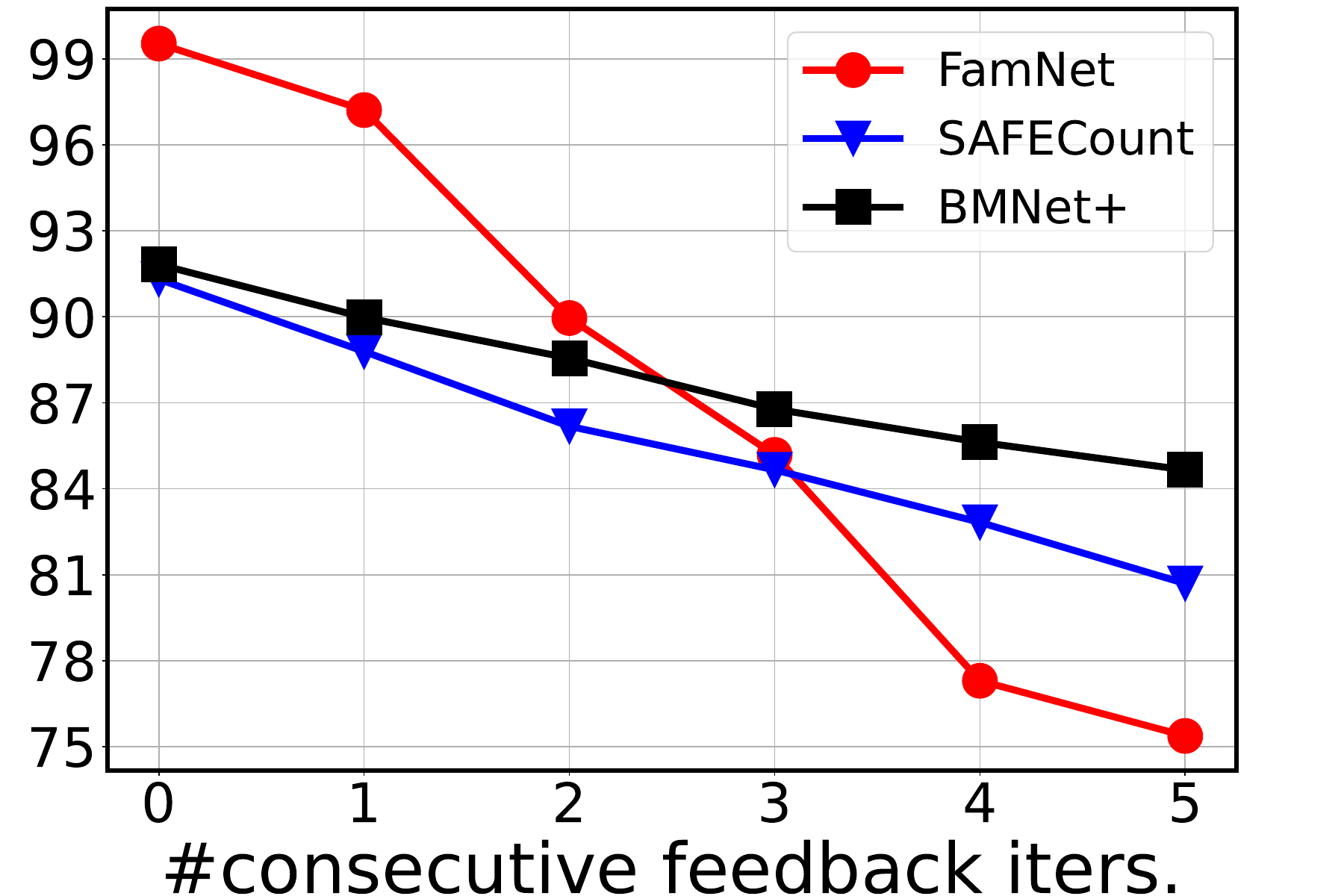} \\
\makebox[\subFigSz]{\footnotesize{MAE on FSCD-LVIS test set}} \hfill 
\makebox[\subFigSz]{\footnotesize{RMSE on FSCD-LVIS test set}} 
\includegraphics[width=\subFigSz]{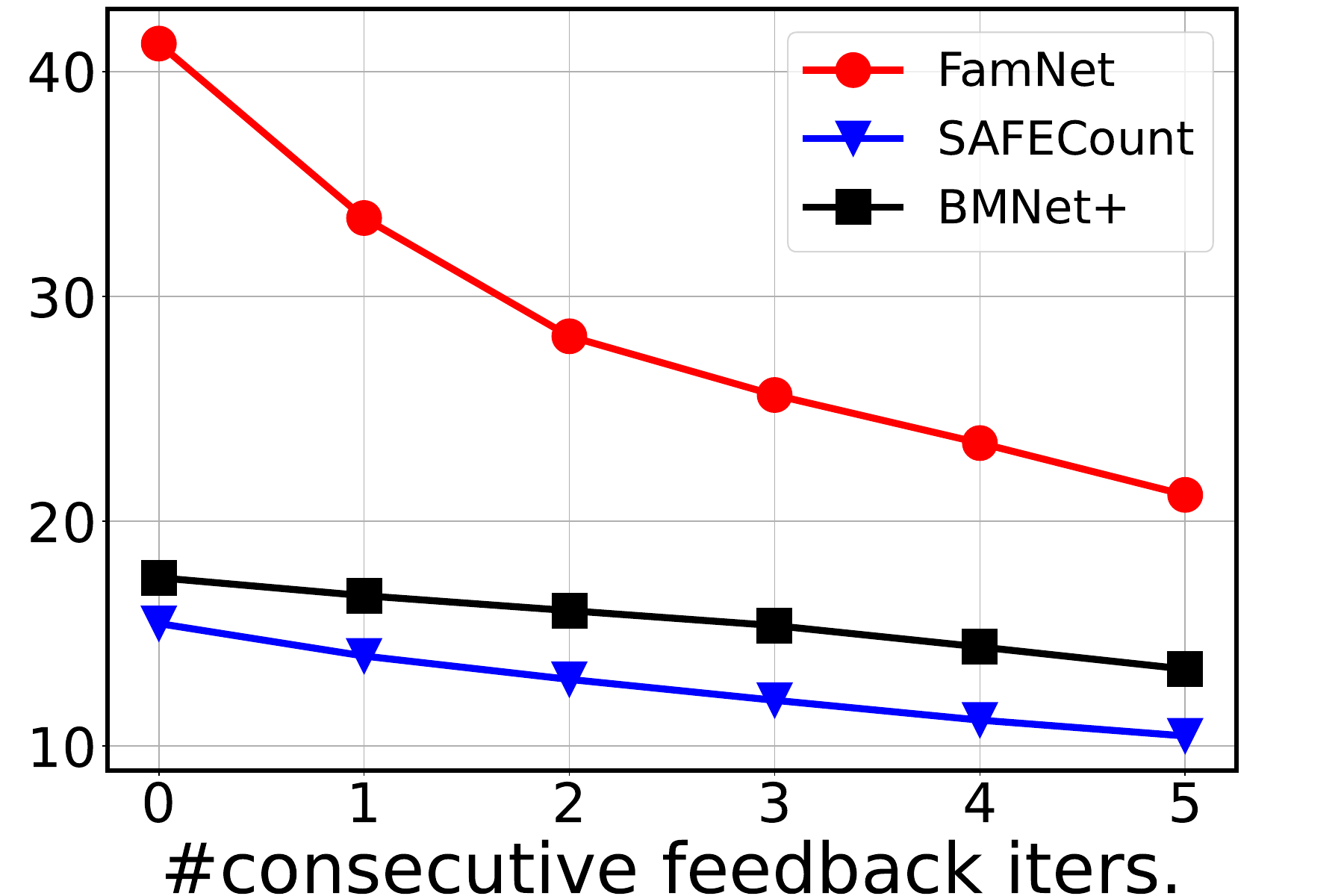} \hfill
\includegraphics[width=\subFigSz]
{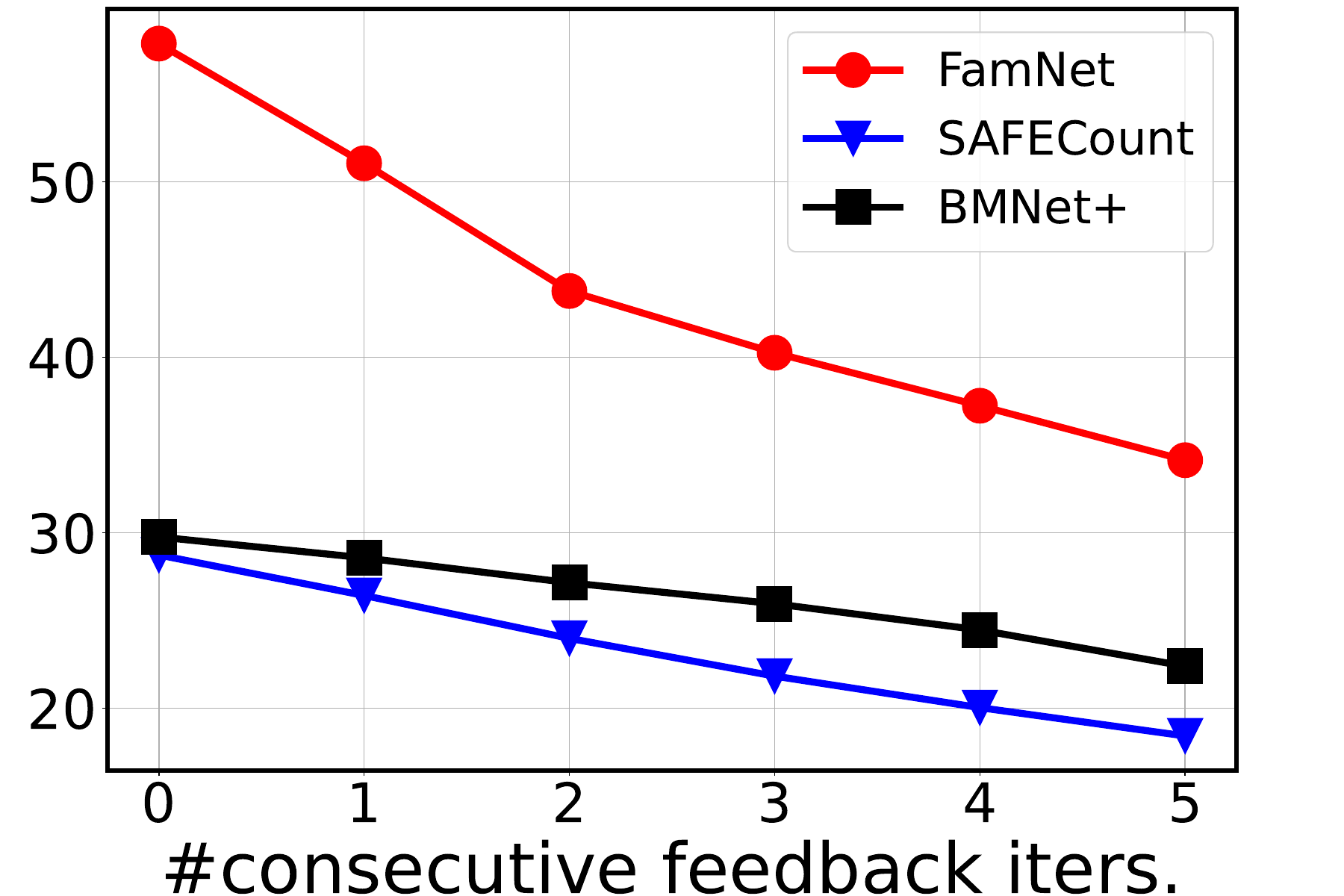} 
\caption{
The proposed framework can be used to improve the performance of various visual counters. This shows the MAE and RMSE values of FamNet, SAFECount, and BMNet+ on FSC-147 and FSCD-LVIS test data, as the number of feedback iterations is increased from zero (without any adaption) to five.  \label{fig:CONSECUTIVE_CLICKS}}
\end{figure}

\def\subFigSz{0.24\linewidth} 
\begin{figure*}[] 
\centering
\makebox[0.48\linewidth]{\small{Before interaction. Predict: \textbf{42.5}, GT: \textbf{17}}} \hfill 
\makebox[0.48\linewidth]{\small{After one interaction. Predict: \textbf{23.0}, GT: \textbf{17}}}\\
\includegraphics[width=\subFigSz]{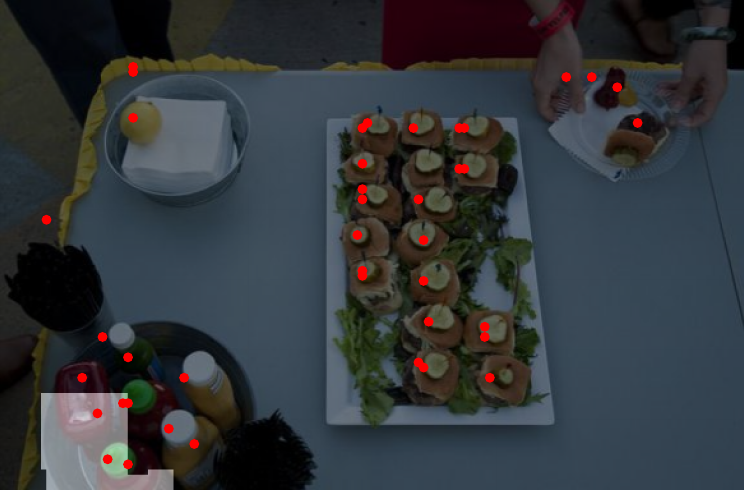} \hfill 
\includegraphics[width=\subFigSz]{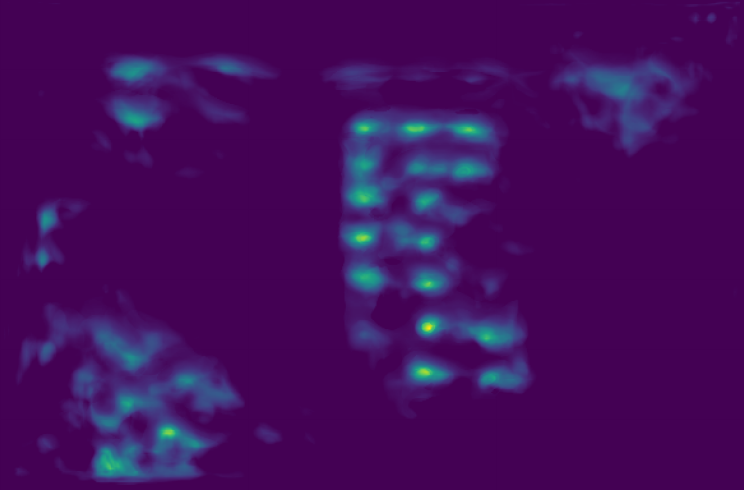} \hfill 
\includegraphics[width=\subFigSz]{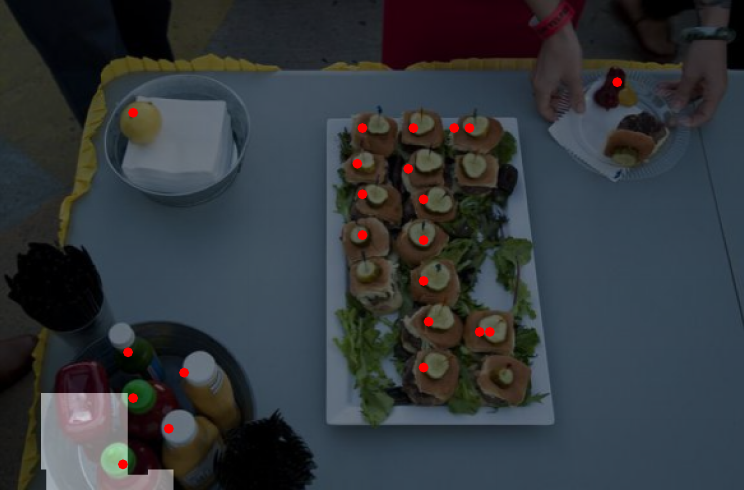} \hfill 
\includegraphics[width=\subFigSz]{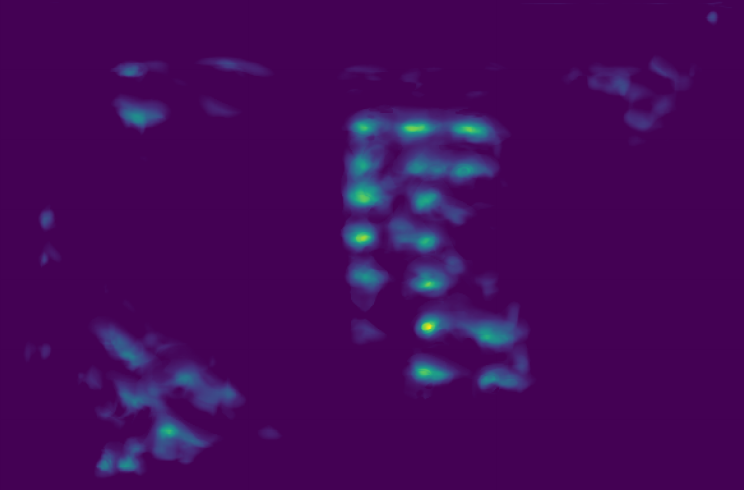}\\ 
\caption{Qualitative results of our approach for test images in FSC-147 with FamNet as the visual counter. The brighter region is the selected region, and the red dot is the approximate location of each region generated by peak selection and non-maximum suppression on each region. The selected region is highlighted in this example, and the input range is $(-\infty, 0]$, since the object of interest is cake. With one single interaction, our method can improve the counting result locally and globally. \textbf{More qualitative results and a demo video are in our supplementary.} \label{fig:QUAL}}
\end{figure*}

The proposed framework requires minimal user input, but it should not be viewed as a competitor to few-shot counting methods. Rather, it is an interactive framework that offers complementary benefits to few-shot methods. Notably, most class-agnostic visual counters, including FamNet, SAFECount, and BMNet+, are few-shot methods that can count with just a few exemplars. As shown earlier, the proposed framework enhances the performance of these counters. However, there is another approach to improve these counters, which is to provide more exemplars. Rather than using these visual counters in our interactive framework, we could offer them additional exemplars. As our framework requires two mouse clicks for each iteration and drawing a bounding box also requires the same effort, we compare the performance of our framework with five iterations to the performance of visual counters with five additional exemplars, and the results are shown in \Tref{tab:FSC_BMNET_SAFECOUNT}. As shown, our framework produces greater improvement with the same level of user effort. This may be due to several reasons: (1) supplying extra exemplars does not immediately highlight prediction errors, resulting in a weaker supervision signal; (2) an exemplar provides information for only one object, less informative than a region containing multiple objects; (3) most class-agnostic counting methods are trained with a predefined number of exemplars (e.g., three), so the model may not be able to fully utilize the additional exemplars to improve the performance.

\setlength{\tabcolsep}{2pt}
\begin{table}[]\footnotesize%
\centering
\begin{tabular}{lcccc}
\toprule 
&  \multicolumn{2}{c}{FSC-147 Test set} & \multicolumn{2}{c}{FSCD-LVIS Test set}\\ 
\cmidrule(lr){2-3} \cmidrule(lr){4-5} 
& MAE & RMSE & MAE & RMSE\\ 
\midrule 
FamNet~\cite{ranjan2021learning} & 22.08 & 99.54 & 41.26& 57.87\\
\hspace{2ex}+ 5 exemplars & 21.52 {\color{SpringGreen} $\downarrow$2\%}& 98.10 {\color{SpringGreen} $\downarrow$1\%} & 40.36 {\color{SpringGreen} $\downarrow$2\%}& 57.85 $\downarrow$0\%\\
\hspace{2ex}+ our framework & 11.75 {\color{ForestGreen} $\downarrow$47\%}& 75.37 {\color{YellowGreen} $\downarrow$24\%}
&  21.18 {\color{ForestGreen} $\downarrow$49\%}& 34.13 {\color{ForestGreen} $\downarrow$41\%}\\

SAFECount~\cite{Zhiyuan2023safecount}  & 13.56 & 91.31 & 15.45& 28.73\\
\hspace{2ex}+ 5 exemplars & 13.01 {\color{SpringGreen} 
 $\downarrow$4\%}& 94.22 {\color{RubineRed}  $\uparrow$3\%} & 14.83 {\color{SpringGreen} $\downarrow$4\%}& 28.01 {\color{SpringGreen} $\downarrow$2\%}\\
\hspace{2ex}+ our framework  & 9.42 {\color{ForestGreen} $\downarrow$31\%}& 80.69 {\color{YellowGreen} $\downarrow$12\%} & 10.45 {\color{ForestGreen} $\downarrow$32\%}& 18.42 {\color{ForestGreen} $\downarrow$36\%}\\

BMNet+~\cite{shi2022represent} & 14.62& 91.83& 17.49& 29.76\\
\hspace{2ex}+ 5 exemplars & 14.40 {\color{SpringGreen} $\downarrow$2\%}& 91.56 $\downarrow$0\% & 17.27 {\color{SpringGreen} $\downarrow$1\%}& 29.60 {\color{SpringGreen} $\downarrow$1\%}\\
\hspace{2ex}+ our framework  & 9.51 {\color{ForestGreen} $\downarrow$35\%}& 84.66 {\color{SpringGreen} $\downarrow$8\%} & 13.43 {\color{YellowGreen} $\downarrow$23\%}& 22.39 {\color{YellowGreen} $\downarrow$25\%}\\
\bottomrule 
\end{tabular}
\vskip 0.05in
\caption{Comparing the performance of the proposed interactive framework with \textbf{five feedback iterations} to a few-shot baseline approach that uses the base counters with \textbf{5 additional exemplars}.
\label{tab:FSC_BMNET_SAFECOUNT}}
\vspace{-0.4cm}
\end{table}

One technical contribution of our method is the innovative segmentation algorithm. \Tref{tab:FSC_FAMNET} compares this algorithm with four segmentation methods: MSER~\cite{arteta2014interactive}, K-means, {Watershed}, and {DBSCAN}. For K-means and DBSCAN, we use the spatial coordinate and the density value as the feature to perform clustering for segmentation. For K-means, $K$ is set to $\min(S_u, \max(S_l, \frac{Sum(D)}{C}))$, where $S_u$ and $S_l$ are the pre-defined upper bound and lower bound, $Sum(D)$ is the summed density, and $C$ is the count limit. For these density map segmentation baselines, we also use our feature refinement adaptation scheme to update the visual counter. As shown in \Tref{tab:FSC_FAMNET}, our proposed algorithm surpasses the other methods by a wide margin. \Fref{fig:QUAL} shows some qualitative results. 

\setlength{\tabcolsep}{3pt}
\begin{table}[]\footnotesize%
\centering
\begin{tabular}{lcccc}
\toprule
&  \multicolumn{2}{c}{FSC-147 Test set} & \multicolumn{2}{c}{FSCD-LVIS Test set}\\ 
\cmidrule(lr){2-3} \cmidrule(lr){4-5} 
& MAE & RMSE & MAE & RMSE\\ 
\midrule 
Initial error& 22.08 & 99.54 & 41.26& 57.87\\
\hline
MSER~\cite{arteta2014interactive}& 16.46$\pm$0.13 & 86.06$\pm$3.28
& 30.61$\pm$0.37 & 44.57$\pm$0.77\\
Watershed& 18.95$\pm$0.09 & \textbf{74.08$\pm$4.66}
& 27.35$\pm$0.25 & 43.21$\pm$0.79\\
K-means& 15.32$\pm$0.23 & 86.49$\pm$2.43
& 32.70$\pm$0.04 & 47.80$\pm$0.09\\
DBSCAN& 19.69$\pm$0.24 & 78.26$\pm$8.08
& 41.26$\pm$0.00 & 57.87$\pm$0.00\\
IPSE (proposed) & \textbf{11.75}$\pm$\textbf{0.12}& 75.37$\pm$5.21
&  \textbf{21.18$\pm$0.28}& \textbf{34.13$\pm$0.88}\\
\bottomrule 
\end{tabular}
\vskip 0.05in
\caption{Results of different segmentation methods under the same adaptation scheme
with {\bf five feedback iterations}, when FamNet is the base
visual counter. Each experiment is repeated five times, and the \textbf{mean} and \textbf{standard error} are reported. }
\label{tab:FSC_FAMNET}
\end{table}

\subsubsection{Ablation study}
\label{sec:aba}

We perform some experiments to evaluate the contribution of different components, including the refinement module, the adaptation loss, and the setting of user feedback simulation. All ablation studies are conducted on the FSC-147 validation set.

\myheading{Refinement module.} The results of the ablation study for the refinement module are shown in \Tref{tab:ABA_FRM}. Both the channel-wise and spatial-wise refinement steps are important. The channel-wise refinement contributes more to the improvement than the spatial-wise refinement step. This is perhaps because the spatial-wise refinement refines locally while the channel-wise refinement refines globally, as shown in~\Fref{fig:Refine}. Also, the order of these two refinement steps has little effect on the final result.

\def\subFigSz{0.30\linewidth} 
\begin{figure}[t] 
\centering
\begin{center}
\makebox[0.30\linewidth]{\scriptsize{Before interaction}} 
\makebox[0.30\linewidth]{\scriptsize{Channel-wise only}} 
\makebox[0.30\linewidth]{\scriptsize{Spatial-wise only}} \\
\includegraphics[width=\subFigSz]{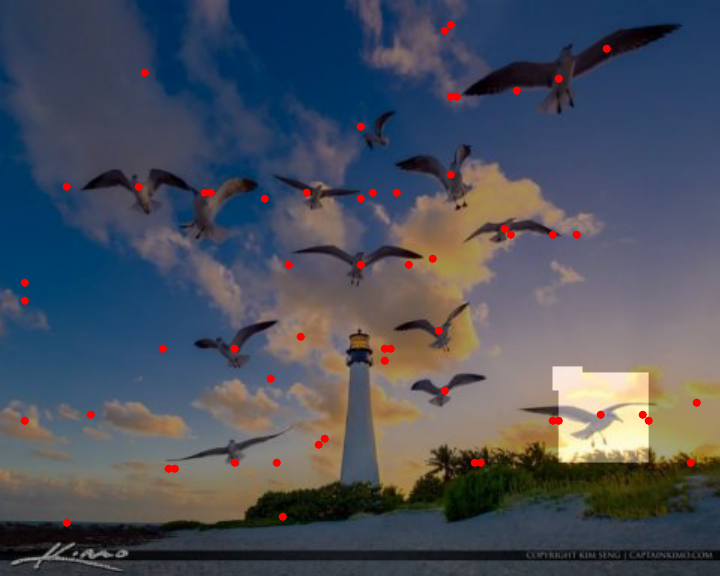} 
\includegraphics[width=\subFigSz]{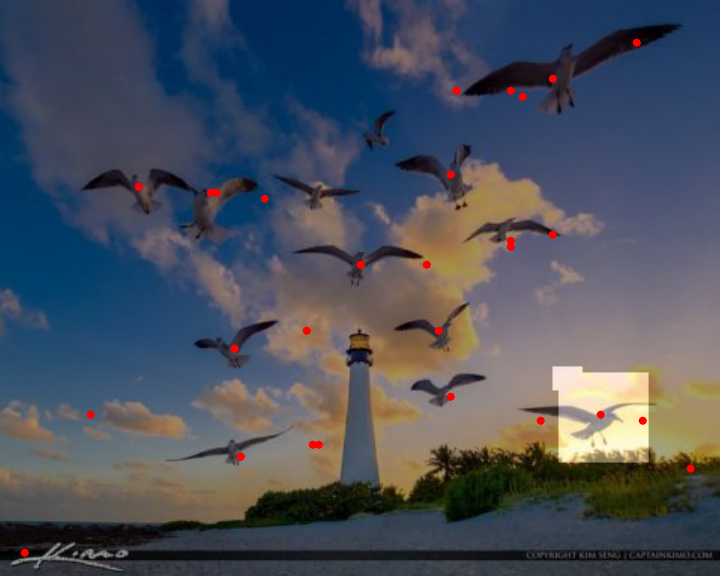} 
\includegraphics[width=\subFigSz]{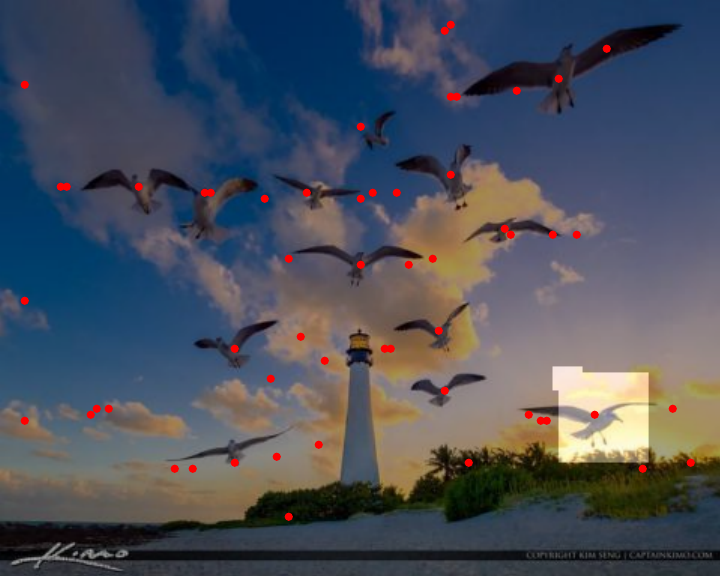}
\end{center}
\vskip -0.1in
\caption{Channel-wise contributes more to the improvement than spatial-wise refinement since channel-wise refinement corrects global errors, while spatial-wise focuses on local errors.}
\label{fig:Refine}
\end{figure}

\setlength{\tabcolsep}{5pt}
\begin{table}[t] 
\centering
\begin{tabular}{lcccc}
\toprule
& \multicolumn{2}{c}{FamNet} &  \multicolumn{2}{c}{SAFECount}\\ 
\cmidrule(lr){2-3} \cmidrule(lr){4-5} 
Refinement Component & MAE & RMSE & MAE & RMSE\\ 
\midrule
Spatial& 21.63& 67.84 & 13.90 & 10.67\\
Channel& 13.72& 52.99 & 9.46 & 13.01\\
Spatial + Channel& 12.84 & \textbf{45.85} & 9.98 & 36.15\\
Channel + Spatial& \textbf{12.79} & 47.21 & \textbf{9.29} & \textbf{33.83}\\
\bottomrule 
\end{tabular}
\vskip 0.05in
\caption{Analyzing the contribution of the spatial and refinement components of the refinement module. The result is on FSC-147 validation set with FamNet and SAFECount as the visual counter.}
\label{tab:ABA_FRM}
\end{table}

\begin{table}[t]\footnotesize%
\setlength\tabcolsep{2pt}
\centering
\begin{tabular}{lcccc}
\toprule 
Local Loss $\mL_L(\Omega)$ & \CheckmarkBold& \XSolidBrush& \CheckmarkBold& \CheckmarkBold\\
Global Loss $\mL_G(\Omega)$& \XSolidBrush& \CheckmarkBold& \CheckmarkBold& \CheckmarkBold\\
Confidence scaling& \XSolidBrush& \XSolidBrush& \XSolidBrush& \CheckmarkBold\\
\midrule 
MAE& 15.41$\pm$0.29 & 14.21$\pm$0.17 & 13.25$\pm$0.21& \textbf{12.79$\pm$0.16}\\
RMSE& 55.72$\pm$1.57 & 52.94$\pm$1.56 & 48.44$\pm$1.69& \textbf{47.21$\pm$2.05}\\ 
\bottomrule 
\end{tabular}
\vskip 0.05in
\caption{Ablation on the adaptation loss, the result is on FSC-147 validation set with FamNet as the visual counter.}
\label{tab:ABA_LOSS}
\end{table}

\myheading{Adaptation loss.} \Tref{tab:ABA_LOSS} shows the ablation study on the adaptation loss. Both the Local Loss $\mL_L(\Omega)$ and the Global Loss $\mL_G(\Omega)$ contribute to the reduction in MAE and RMSE. The confidence scaling is less important.

\setlength{\tabcolsep}{5pt}
\begin{table}[t]\small%
\centering
\begin{tabular}{lcccc}
\toprule
& \multicolumn{2}{c}{FamNet} &  \multicolumn{2}{c}{SAFECount}\\ 
\cmidrule(lr){2-3} \cmidrule(lr){4-5} 
Refinement Component & MAE & RMSE & MAE & RMSE\\ 
\midrule
LocalCorrection& 20.74& 64.58 & 13.01 & 49.15 \\
AllParamAdapt& 17.07& 61.61 & 9.98 & 36.15 \\
Proposed& \textbf{12.79} & \textbf{47.21} & \textbf{9.29} & \textbf{33.83} \\
\bottomrule 
\end{tabular}
\vskip 0.05in
\caption{Analyzing the effectiveness of our adaptation scheme.}
\label{tab:ABA_ADPT_SCHEME}
\end{table}

\myheading{Adaptation scheme.} \Tref{tab:ABA_ADPT_SCHEME} shows the ablation study on the adaptation scheme. 
LocalCorrection is the method that only corrects the prediction of the selected region, and will not adapt the visual counter. AllParamAdapt is the method that updates all the parameters in the regression head. 

\myheading{User feedback simulation.} To further assess the efficiency and efficacy of various region selection strategies, we consider two methods that may offer advantages over random selection used in previous experiments. These two strategies are: 1) prioritizing background regions containing no objects, and 2) selecting regions with largest errors. 
The comparison of these region selection strategies is presented in \Tref{tab:R3_Q1}. The error-based approach emerges as the most successful, whereas prioritizing background regions has a detrimental impact on performance. Concerning time efficiency, random selection is the quickest, followed by background prioritization, with error-based selection being the slowest due to the need for error estimation for each region.


\begin{table}[t] 
\centering
\begin{tabular}{lcc}
\toprule
Region Selection Strategy & MAE & RMSE\\ 
\midrule
Background Prior& 14.02& 55.31\\
Error Based& \textbf{10.61}& \textbf{46.85}\\
Random Selection& 12.79 & 47.21\\
\bottomrule 
\end{tabular}
\vskip 0.05in
\caption{Result on different region selection strategies with FamNet on FSC-147 validation set.}
\label{tab:R3_Q1}
\end{table}

\begin{figure}[t]
    \centering
    \includegraphics[width=0.9\linewidth]{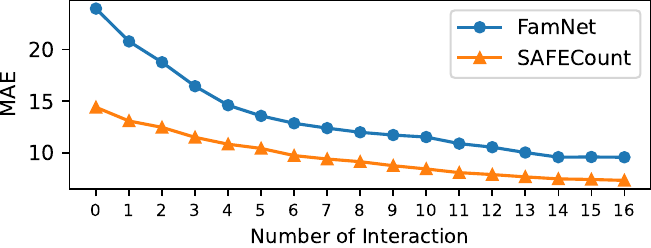}
    \caption{MAE with varying numbers of interactions on the \mbox{FSC-147} validation set.}
    \label{fig:R2_Q4}
\end{figure}

In the primary experiment, we set interactions at five, given consecutively. To validate this five-interaction approach, we conducted extra trials on the FSC-147 dataset with varying interaction counts. Results are shown in~\Fref{fig:R2_Q4}, revealing continued performance improvement beyond five interactions, though at a slower rate.

We also compared consecutive and non-consecutive interactions to confirm the importance of sequential engagement. \Tref{tab:R3_Q5} outlines these results, indicating better performance with sequential interactions. However, this approach increases time consumption due to added segmentation and adaptation time.

\begin{table}[t]
\centering
\begin{tabular}{lcccc}
\toprule
& \multicolumn{2}{c}{FamNet} &  \multicolumn{2}{c}{SAFECount}\\ 
\cmidrule(lr){2-3} \cmidrule(lr){4-5} 
 & MAE & RMSE & MAE & RMSE\\ 
\midrule
Consecutive& 12.79 & 47.21  & 9.29 & 33.83 \\
Non Consecutive& 16.87 & 54.42  & 12.36 & 46.97 \\
\bottomrule 
\end{tabular}
\vskip 0.05in
\caption{Comparison of consecutive and non-consecutive interactions, the result is on the FSC-147 validation set.}
\label{tab:R3_Q5}
\end{table}

\subsection{Class-specific counting}

Our interactive counting framework has primarily focused on class-agnostic counters, with its foundation being the correlation between the exemplar objects and the image. However, it is natural to wonder if this framework can be extended to class-specific visual counters. The main concern is that the user feedback collected in this manner may be too weak for class-specific counters that are trained on hundreds of thousands or even millions of annotated objects. In this section, we report our experiments on the crowd-counting and car-counting task, where we have found that our method can be used to reduce the errors of a crowd counter if the counting ranges for the user feedback are adjusted. 

Specifically, we apply our framework to DM-Count~\cite{wang2020DMCount} for crowd counting on two crowd-counting benchmarks: ShanghaiTech~\cite{zhang2016single} and UCF-QNRF~\cite{idrees2018composition}. We set the counting limit to 50, and the range interval to 10, so the counting ranges are $\{[-\infty, 0], (0,10], \ldots, (40~50], (50~\infty)\}$. We insert the refinement module after the first convolution layer. We adapt the DM-Count with 20 gradient steps with a learning rate of $5\times10^{-4}$.  We compare with the other baseline methods, and the feedback simulation is identical to those reported in \Sref{sec:cac}. The quantitative results are shown in~\Tref{tab:CSC_DMCOUNT}, and the qualitative results are shown in~\Fref{fig:R3_Q3}. Our approach reduces the MAE and RMSE by approximately 40\% and 30\% on ShanghaiTech A, and approximately 30\% and 25\% on UCF-QNRF. For car counting, we apply our framework to FamNet and SAFECount on CARPK~\cite{hsieh2017drone}. The results are shown in~\Tref{tab:CARPK}. The MAE decreased more than 15\% on CARPK.

\setlength{\tabcolsep}{2pt}
\begin{table}[t]\footnotesize%
\centering
\begin{tabular}{lcccccc}
\toprule 
& \multicolumn{2}{c}{ShanghaiTech A} &  \multicolumn{2}{c}{UCF-QNRF}\\ 
\cmidrule(lr){2-3} \cmidrule(lr){4-5} 
& MAE & RMSE & MAE & RMSE\\ 
\midrule 
Initial error & 59.60 & 95.56 & 85.65  & 148.35 \\
\midrule 
MSER~\cite{arteta2014interactive}& 57.90$\pm$0.32& 94.84$\pm$0.50& 81.38$\pm$0.58& 142.73$\pm$0.93\\
Watershed& 58.84$\pm$0.71& 90.17$\pm$2.06& 82.78$\pm$0.46& 144.70$\pm$0.39\\
K-means& 48.20$\pm$0.75& 79.23$\pm$2.46& 73.29$\pm$1.77& 131.37$\pm$3.98\\
DBSCAN& 56.99$\pm$0.21& 92.73$\pm$0.28& 80.62$\pm$0.93& 142.29$\pm$1.66\\
LocalCorrection& 43.86$\pm$0.26& 74.76$\pm$0.84& 78.27$\pm$0.23& 135.07$\pm$0.53\\
AllParamAdapt& \textbf{31.03$\pm$0.33}& 59.47$\pm$0.96& 62.30$\pm$2.81& 123.03$\pm$7.52\\ 
Proposed& 33.85$\pm$0.78& \textbf{57.50$\pm$1.94}& \textbf{58.13$\pm$1.04}& \textbf{102.32$\pm$2.63}\\
\bottomrule 
\end{tabular}
\vskip 0.05in
\caption{Results of interactive adaptation methods for a crowd-counting network (DM-Count) using five feedback iterations.}
\label{tab:CSC_DMCOUNT}
\end{table}

\setlength{\tabcolsep}{5pt}
\begin{table}[t]
\centering
\begin{tabular}{lcccc}
\toprule 
& \multicolumn{2}{c}{Initial Error} & \multicolumn{2}{c}{Five Interactions}\\
\cmidrule(lr){2-3} \cmidrule(lr){4-5}
&  MAE & RMSE & MAE & RMSE\\ 
\midrule 
FamNet& 18.34& 35.77 & 13.91 & 20.14\\
SAFECount& 4.91 & 6.32 & 4.16 & 5.91\\
\bottomrule 
\end{tabular}
\vskip 0.05in
\caption{Result on CARPK with FamNet and SAFECount using five feedback iterations.}
\label{tab:CARPK}
\end{table}

\def\subFigSz{0.49\linewidth} 
\begin{figure}[t] 
\centering
\makebox[0.49\linewidth]{\scriptsize{GT:553.0, Pred:615.3}} \hfill 
\makebox[0.49\linewidth]{\scriptsize{GT:553.0, Pred:534.5}} \\
\includegraphics[width=\subFigSz]{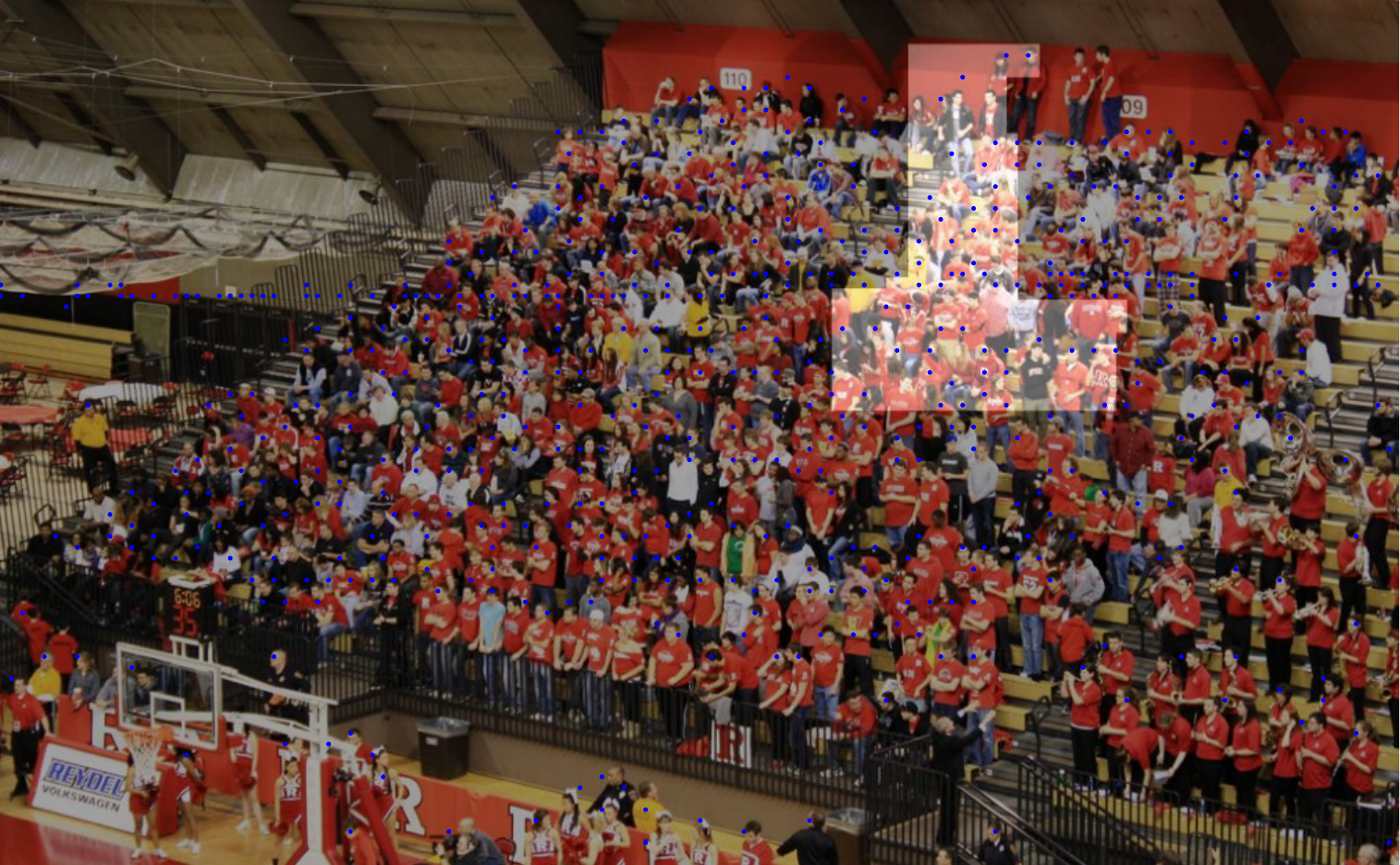} \hfill 
\includegraphics[width=\subFigSz]{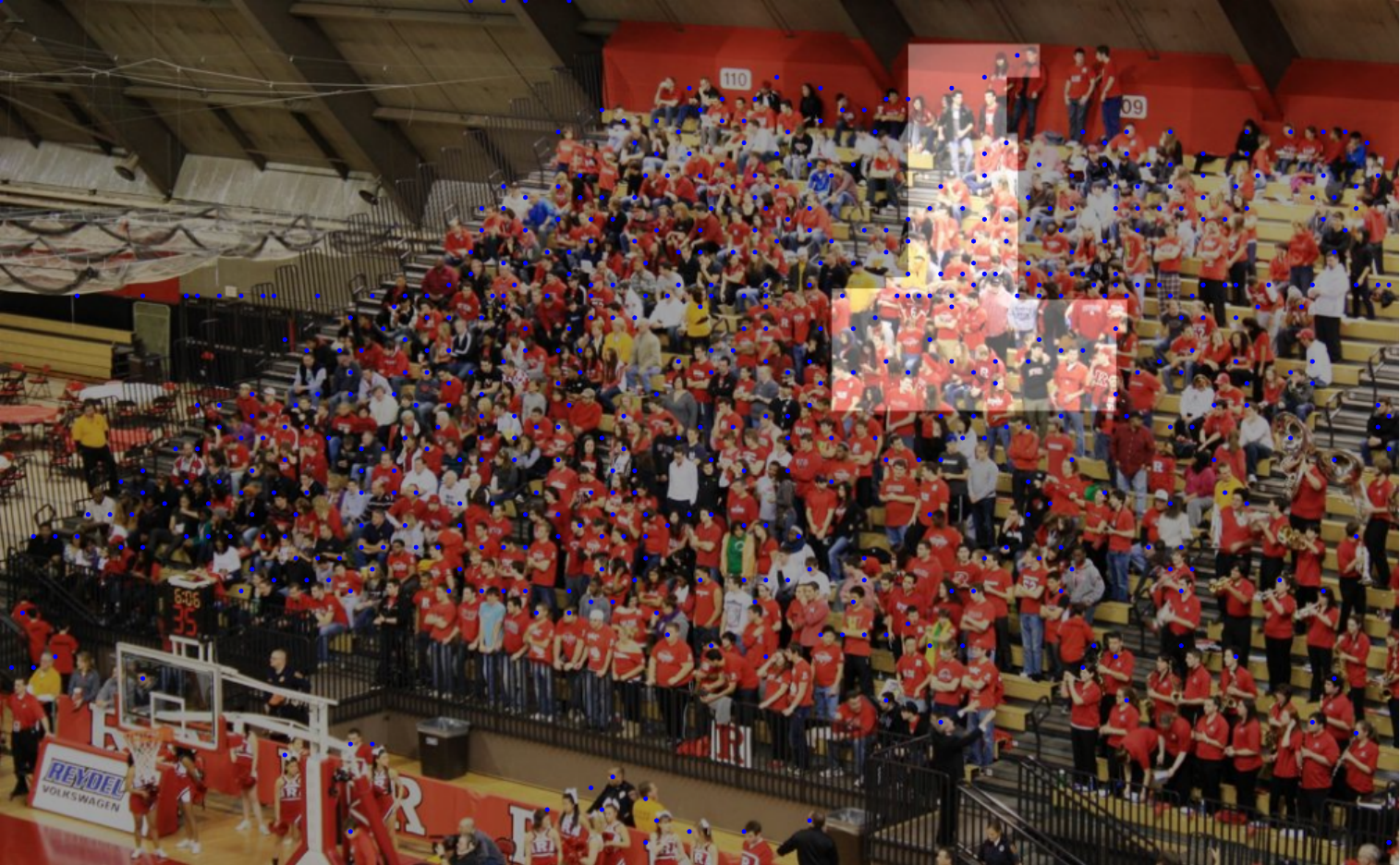}
\caption{Qualitative results of our approach for test images in ShanghaiTech A with DM-Count as the visual counter. The left figure is before interaction, and the right is after one interaction. The selected region is highlighted, and the input range is $(20, 30]$.}
\label{fig:R3_Q3}
\end{figure}

\begin{figure}[t]
  \centering
  \includegraphics[width=0.9\linewidth]{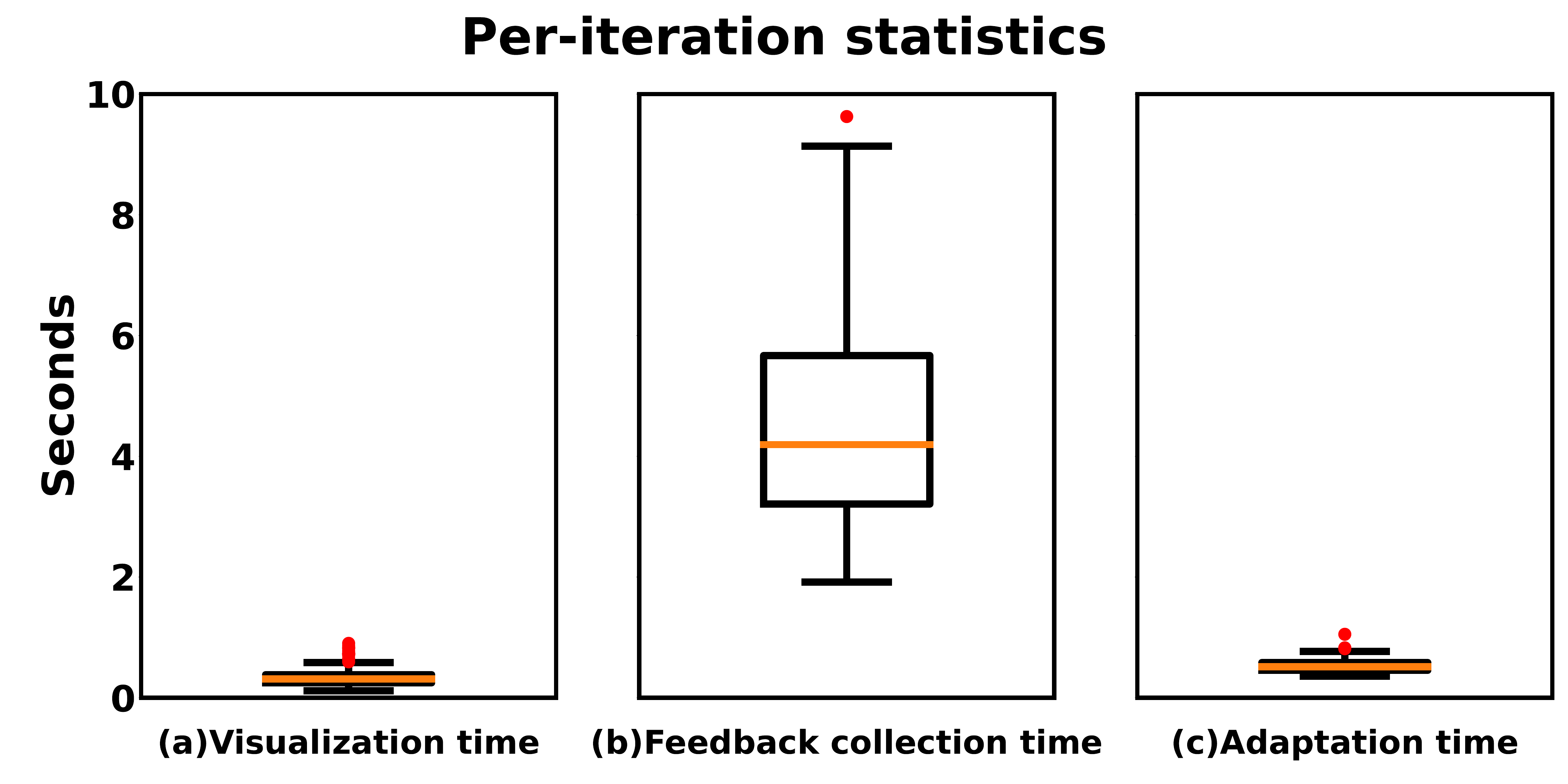}
  \caption{Per-iteration statistics. The visualization time (including segmentation time) and the adaptation time are both less than one second, sufficiently fast for interactive systems.}
  \label{fig:EXP_USER}
  \vspace{-0.2cm}
\end{figure}

\subsection{User study}
\label{sec:user_study}

To assess the feasibility of the proposed interactive counting framework, we conducted a user study with eight participants. We selected 20 images with high counting errors from the FSC-147 dataset and used FamNet as the visual counter. Each participant was allowed a maximum of five iterations for each image, but they could terminate the process if they felt the prediction was accurate enough. The average number of iterations for one image is $3.08$, and the variance is 0.41. Additionally, we carried out an experiment involving simulated user feedback on the same set of images, to compare the results with those obtained from real user feedback. As shown in \Tref{tab:USER_FEEDBACK_COMP}, the users were able to improve the performance of the counter using our framework, demonstrating its potential for practical usage. Moreover, the benefits achieved from real user feedback were comparable to those obtained from simulated feedback, indicating that many of our analyses using simulated feedback can be extrapolated to real-world scenarios.

\setlength{\tabcolsep}{2pt}
\begin{table}[t]\footnotesize%
\centering
\begin{tabular}{lcc}
\toprule
\#Feedback Iterations & MAE & RMSE\\ 
\midrule
Initial error& 93.41 & 125.57\\
Real User (avg. 3.08 iters)& $45.11\pm2.63$ $\downarrow$52\%& $90.64\pm2.68$ $\downarrow$28\%\\
Simulated feedback (3 iters) & $59.97\pm3.34$ $\downarrow$ 34\%& $110.67\pm5.96$ $\downarrow$12\%\\
Simulated feedback (5 iters) & $43.86\pm 3.02$ $\downarrow$ 52\% & $92.93\pm2.90$ $\downarrow$26\%\\
\bottomrule 
\end{tabular}
\vskip 0.05in
\caption{Comparison between real user's feedback and the simulated feedback used in our quantitative experiments.}
\label{tab:USER_FEEDBACK_COMP}
\vspace{-0.4cm}
\end{table}

The user study was conducted on an RTX3080 machine, and several time statistics of a single iteration are presented in \Fref{fig:EXP_USER}. It took less than a second to segment any density image and display it. The average time for a user to select a region and specify a range was four seconds, and the adaptation time for a single iteration was less than one second. All operations are sufficiently fast for interactive systems.

\section{Conclusions} 
We have proposed an interactive framework primarily for class-agnostic counting, but can also be extended to class-specific counting. It uses a novel method for density map segmentation to generate an intuitive display for the user, enabling them to visualize the results and provide feedback. Additionally, we have developed an adaptation loss and a refinement module to efficiently and effectively update the visual counter with the user's feedback. Experiments on two class-agnostic counting datasets and two crowd-counting benchmarks with four different visual counters demonstrate the effectiveness and general applicability of our framework. 

\myheading{Acknowledgement}: This project was supported by US National Science Foundation Award NSDF DUE-2055406.

\newpage

\section{Supplementary Overview}
In the supplementary, we first provide more details about our approach in section~\ref{sec:ApproachDetails}. Then provide more implementation details in section~\ref{sec:ImplementationDetails}, analyze the time efficiency and conduct the ablation of the location of the feature refinement module in section~\ref{sec:AdditionalAblation}, analyze our method's robustness in~\ref{sec:Robustness} and analyze the effectiveness of confidence scaling in the other two class-agnostic visual counters in section~\ref{sec:ConfidenceScaling}. After that, we introduce the interface of our interactive system in section~\ref{sec:Interface}, give more qualitative results 
 in section~\ref{sec:Qual1} and section~\ref{sec:Qual2}. Finally, we briefly discuss the limitation and future work in section~\ref{sec:Limitation}. In addition to the supplementary material, we also provide a demo video of our interactive counting system.

\section{Supplementary Material}

\subsection{Additional details for our approach}
\label{sec:ApproachDetails}
In this section, we provide additional details for the interaction loop and the IPSE density map segmentation.

\subsubsection{Interaction loop}
A detailed algorithm for the interaction loop is illustrated in Algorithm~\ref{IntLoop}. The input visual counter contains the following components, a feature extractor $f$, a spatial-similarity learning module $g$, layers before the refinement module in regression head $h_b$, refinement module $\mR_{\theta_r}$, and layers after the refinement module in regression head $h_a$. We need to update $\Omega$ with $\bm{D}$ in each gradient step because the summation over each region depends on the estimated density map.
\begin{algorithm}[t]
\caption{Interaction loop}
\label{IntLoop}
{\textbf{Input:} Input image: $\I$, Exemplars: $\E$, Gradient steps: $N$, Adaptation learning rate $\gamma$, Interaction times: $T$.}\\
{\textbf{Initalization:} User feedback list: $\Omega = [\ ]$.}
\begin{algorithmic}[1]
\State $\S = g(f(\I), f(\E))$
\State $\bm{F} = h_b(\S)$
\State {\small Initialize $\mR_{\theta_r}$ correspond to $\bm{F}$'s size}
\For{$T$ interactions}
\State $\bm{D} = h_a(\mR_{\theta_r}(\bm{F}))$
\State {\small Visualize $\D$ with IPSE}
\State {\small Collect user feedback $(R, c)$}, $\Omega$.append($(R, c)$)
\State $\gamma^\prime = \gamma F_C(\Omega)$, $N^\prime = \frac{N}{F_C(\Omega)}$
\For{$N^\prime$ gradient steps}
\State $\bm{F^\prime} = \mR_{\theta_r}(\bm{F})$
\State $\bm{D} = h_b(\bm{F^\prime})$
\State {\small Update} $\Omega$ {\small with} $\bm{D}$
\State $\theta_r \leftarrow \theta_r - \gamma^\prime\nabla \mathcal{L}(\Omega)$
\EndFor
\EndFor
\end{algorithmic}
\end{algorithm}

\subsubsection{IPSE density map segmentation}
A detailed algorithm for IPSE is illustrated in Algorithm~\ref{VisualizationAlgorithm}. The peak expansion algorithm is shown in Algorithm~\ref{PeakExpandingAlgorithm}. In Algorithm~\ref{VisualizationAlgorithm}, background splitting is simply expanding at a random background peak with iteratively including the neighbor pixels with the same upper bound, and the small region merging is merging some small region to its neighbor region. More specifically, the region size upper bound $T_u$ is set to 1250, and the region size lower bound $T_l$ in the objective function is set to 250.
\begin{algorithm}[t]
\caption{\textbf{IPSE} Density Map Segmentation Algorithm}
\label{VisualizationAlgorithm}
\hspace*{0.02in} 
{\textbf{Input:} Density map: $\bm{D}$, Smooth kernel: $\bm{G}$, Objective function:$h(R)$, Region size upper bound: $T_u$.}\\
  \hspace*{0.02in}
  {\textbf{Initalization:} Foreground region set: $\mathbb{V}_f = \{\ \}$, Background region set: $\mathbb{V}_b = \{\ \}$}
   \begin{algorithmic}[1]
  \State $\bm{\widetilde{D}} \leftarrow \bm{D}*\bm{G}$
  \State $S \leftarrow sum(\bm{D})$
  \While{$S \geq 1$}
  \State $p = argmax(\bm{\widetilde{D}})$
  \State $\bm{\widetilde{D}}[p]  \leftarrow -\infty$
  \State $R$ = {\small Peak Expansion}($\bm{D}$, $\bm{\widetilde{D}}$, $p$, $h(R)$, $T_u$)
  \State $\mathbb{V}_f.append(R)$
  \State $S \leftarrow S - R_s$
  \EndWhile
  \State $\mathbb{V}_b$ = {\small Background\ Spliting} $(\bm{D}, \bm{\widetilde{D}})$
  \State $\mathbb{V} = \mathbb{V}_f \cup \mathbb{V}_b$
  \State $\mathbb{V} \leftarrow$ {\small Small\ Region\ Merging}$(\mathbb{V})$
  \State \Return $\mathbb{V}$
  \end{algorithmic}
\end{algorithm}

\begin{algorithm}[t]
  \caption{Peak Expansion Algorithm}
  \label{PeakExpandingAlgorithm}
{\textbf{Input:}  Density map: $\bm{D}$, 
  Smooth density map: $\bm{\widetilde{D}}$, 
  Peak: $p$, 
  Objective function:$h(R)$, 
  Region size upper bound: $T_u$.}\\
  {\textbf{Initalization:}  $R_s=0$, $R_i=0$, 
  Region pixel list $R_L=[\ ]$, 
  Optimal objective value:$P^* = \infty$, 
  Foreground size:$F_i=0$, 
  Background size:$B_i=0$}.
  \begin{algorithmic}[1]
    \normalsize
    \State{$\hat{L} = [p]$, $L = [\ ]$}
    \While{$\hat{L}$ is not empty \textbf{and} $R_i < T_u$}
    \State $\hat{p} = \hat{L}.pop()$, $L.append(\hat{p})$
    
    \For{$\hat{p}_n  \in \hat{p}$\ 's neighbour}
    \If{$\hat{p}_n$ not in any regions}
    
    \If{$\bm{D}[\hat{p}_n] > 0$}
    \State $\hat{L}.append(\hat{p})$
    \State $R_s \leftarrow R_s + D[\hat{p}_n],\ R_i \leftarrow R_i + 1$
    \State $R_L \leftarrow L + \hat{L},\ F_i \leftarrow F_i + 1$
    \State $\bm{\widetilde{D}}[\hat{p}_n] \leftarrow -\infty$
    \State $P \leftarrow h(R)$
    \If{$P < P^*$}
    \State $R^* \leftarrow R,\ P^* \leftarrow P$
    \EndIf
    \Else
    \If{$F_n > B_n$}
    \State $\hat{L}.append(\hat{p}_n)$
    \State $R_s \leftarrow R_s + D[\hat{p}_n],\ R_i \leftarrow R_i + 1$
    \State $R_L \leftarrow L + \hat{L},\ B_n \leftarrow B_n + 1$
    \State $\bm{\widetilde{D}}[\hat{p}_n] \leftarrow -\infty$
    \EndIf
    \EndIf
    \EndIf
    \EndFor
    \EndWhile
    \State \Return $R^*$

\end{algorithmic}
\end{algorithm}

\subsection{Additional implementation details}
\label{sec:ImplementationDetails}
\myheading{FamNet~\cite{ranjan2021learning}.} For FSC-147~\cite{ranjan2021learning} we used the released pre-trained model. For FSCD-LVIS~\cite{nguyen2022few}, we train it on one RTX A5000 machine for 150 epochs, and the learning rate is $1\times10^{-6}$. On FSC-147, following~\cite{ranjan2021learning}, we do the test-time adaptation, on FSCD-LVIS we do not do the test-time adaptation for time efficiency. 

\myheading{SAFECount~\cite{Zhiyuan2023safecount}.} For FSC-147 we used the released pre-trained model. For FSCD-LVIS, we train it on one RTX A5000 machine for 100 epochs, and the learning rate is $2\times10^{-5}$. The interactive adaptation gradient steps are set to 30, and the interactive adaptation learning rate is 0.001.

\myheading{BMNet+~\cite{shi2022represent}.} For FSC-147 we used the released pre-trained model. For FSCD-LVIS, we train it on one RTX A5000 machine for 100 epochs, and the learning rate is $1\times10^{-5}$. The interactive adaptation gradient steps are set to 30, and the interactive adaptation learning rate is 0.001.

\myheading{DM-Count~\cite{wang2020DMCount}.} For ShanghaiTech and UCF-QNRF, we used the released pre-trained model.

\subsection{Robustness experiment on crowd counting}
\label{sec:Robustness}
\begin{table}[]\footnotesize%
\setlength\tabcolsep{2pt}
\centering
\begin{tabular}{lcrrr}
\toprule 
& & \multicolumn{3}{c}{Level of feedback noise} \\
\cmidrule(lr){3-5}
Dataset & Initial error & None & Moderate & Large \\ 
\midrule
ShanghaiTech A& 59.60 & 33.85$\pm$0.78 & 36.15$\pm$0.99 & 40.93$\pm$0.65\\
UCF-QNRF& 85.65 & 58.13$\pm$1.04 & 78.72$\pm$2.36 & 78.14$\pm$1.34\\
\bottomrule
\end{tabular}
\vskip 0.05in
\caption{MAE of the proposed interactive counting method for different levels of feedback noise.}
\label{tab:ROBUSTNESS_EXP}
\vspace{-0.3cm}
\end{table}

Our experiment on crowd counting has demonstrated the  effectiveness of our adaptation method from the computational perspective. But from the human perspective, it may be possible that the human user cannot easily provide feedback for the count ranges needed for crowd counting. This is not a concern for the small count limit and the count ranges  used in the class-agnostic counting setting given the subitizing ability of humans. But for the count ranges $\{[-\infty, 0], (0,10], \ldots, (40~50], (50~\infty)\}$ used in this crowd counting experiment, a human user might make estimation mistakes leading to noisy feedback. We therefore perform an experiment to study the robustness of our method to noisy feedback.  Specifically, we introduce random biases to the ground truth estimation to simulate  mistakes. We consider two estimation biases. For moderate bias, a random noise of 30\% of the count limit is added ([-15, 15]). For large bias, a random noise of 50\% is added([-25, 25]). With a biased estimation, our approach still can reduce the MAE by approximately 30\% on ShanghaiTech A and 10\% on the other two datasets, as shown in  \Tref{tab:ROBUSTNESS_EXP}.

\subsection{Additional ablation study}
\label{sec:AdditionalAblation}
All additional ablation study is conduct on FSC-147 validation set or FSCD-LVIS validation set with FamNet as the visual counter.
\subsubsection{Time efficiency analysis}
\Tref{tab:ABA_TIME} shows the time efficiency comparison with vanilla adaptation(Adapt the whole regression head). In \Tref{tab:ABA_TIME} the average adaptation time(second) for one single click is reported. This experiment is run on RTX A5000, for FSC-147 both of them use 10 gradient steps for one adaptation, and for FSCD-LVIS is 20. We find that our approach is 11.88\% faster than vanilla adaptation on FSC-147, and 8.92\% faster on FSCD-LVIS. Our method is faster because our method requires less computation in feedforward, backpropagation, and parameter updating, as illustrated in Algorithm~\ref{IntLoop}. In feedforward, we only need to compute the layer before the refinement module one time, and in backpropagation, we only need to compute the gradient for the layers after the refinement module. Also, we only need to update the parameters in the refinement module.

\subsubsection{Location of the refinement module.}
The ablation of the location of the refinement module is shown in \Tref{tab:ABA_POS}. This experiment is conduct on FSC-147 validation set. Correlation map means directly refine the spatial correlation map between the exemplar and the input image. We can find that inserting at the shallow position has better performance, and directly refining the correlation map doesn't work well.

\begin{table}[]\footnotesize%
\centering
\begin{tabular}{lcc}
\hline
Benchmarks & FSC-147 Val & FSCD-LVIS Val\\ 
\midrule
Vanilla Adaptation& 0.179$\pm$0.00& 0.56$\pm$0.04\\
Refinement Module& 0.160$\pm$0.00& 0.51$\pm$0.00\\
\hline
\end{tabular}
\caption{Average adaptation time(second) for one single interaction. The mean and the standard error of five experiments with different seeds are reported.}
\label{tab:ABA_TIME}
\end{table}

\begin{table}[]%
\centering
\begin{tabular}{lccc}
\hline
Component & MAE & RMSE\\ 
\midrule
Correlation map& 18.71$\pm$0.78 & 64.15$\pm$9.69\\
After first conv& \textbf{12.79$\pm$0.16} & \textbf{47.21$\pm$2.05}\\
After second conv& 13.63$\pm$0.34 & 48.11$\pm$3.95\\
After third conv& 13.87$\pm$0.13 & 51.56$\pm$1.62\\
\hline
\end{tabular}
\caption{Results of different locations of refinement module on the regression head of FamNet. The mean and the standard error of five experiments with different seeds are reported.}
\label{tab:ABA_POS}
\end{table}

\subsection{Additional analysis on confidence scaling}
\label{sec:ConfidenceScaling}
In the ablation of the adaptation loss in our main paper, confidence scaling seems less important on FamNet. To further analyze its effectiveness, we conduct additional analysis of confidence scaling on SAFECount and BMNet+. As shown in \Fref{fig:SAFECOUNT_CONFIDENCE} and \Fref{fig:BMNET_CONFIDENCE} We can find that confidence scaling can make the adaptation smoother and improve the final result significantly.

\def\subFigSz{0.48\linewidth} 
\begin{figure}[t] 
\centering
\makebox[\subFigSz]{\small{(a) FSC-147}} \hfill 
\makebox[\subFigSz]{\small{(b) FSCD-LVIS}}
\includegraphics[width=\subFigSz]{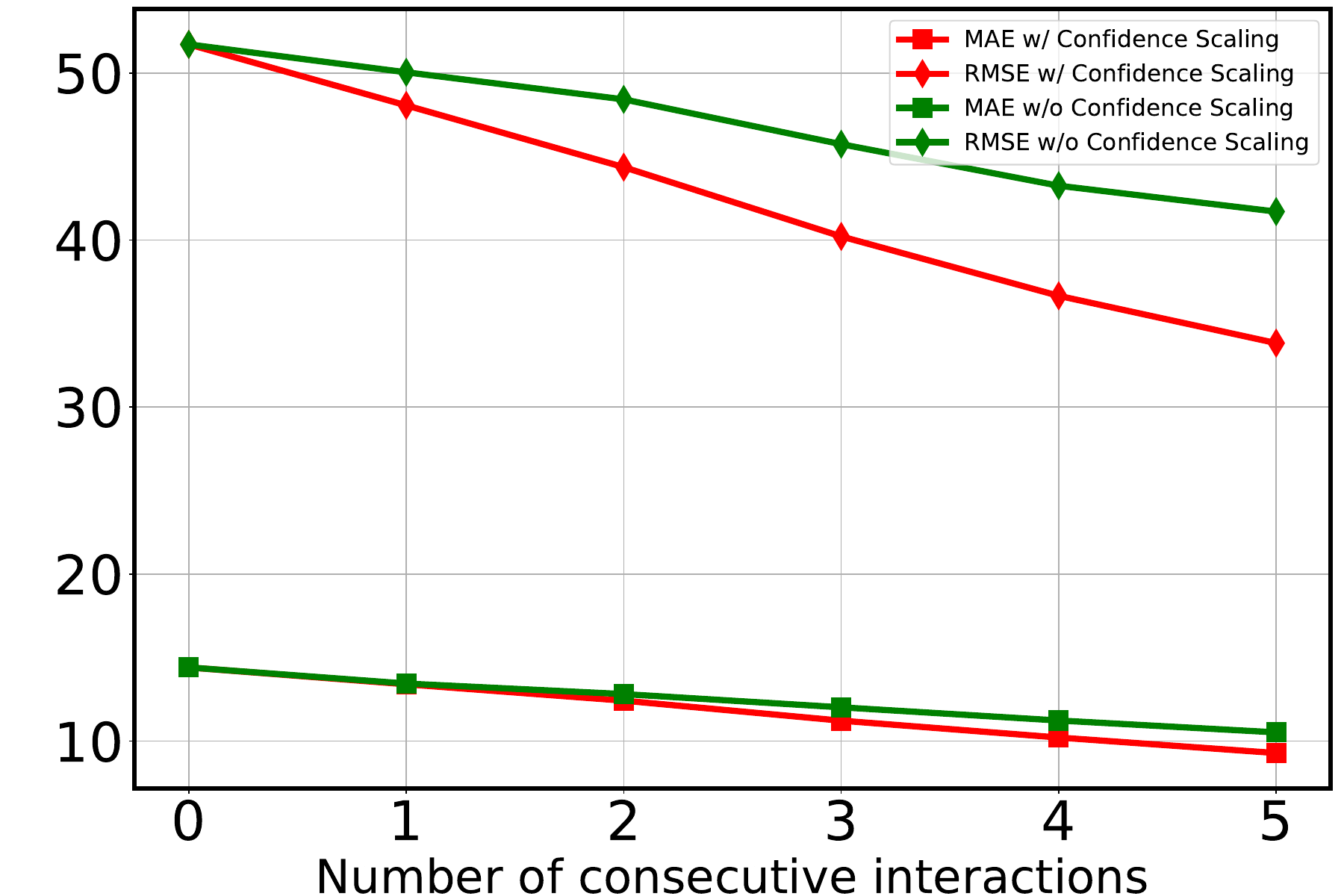} \hfill 
\includegraphics[width=\subFigSz]{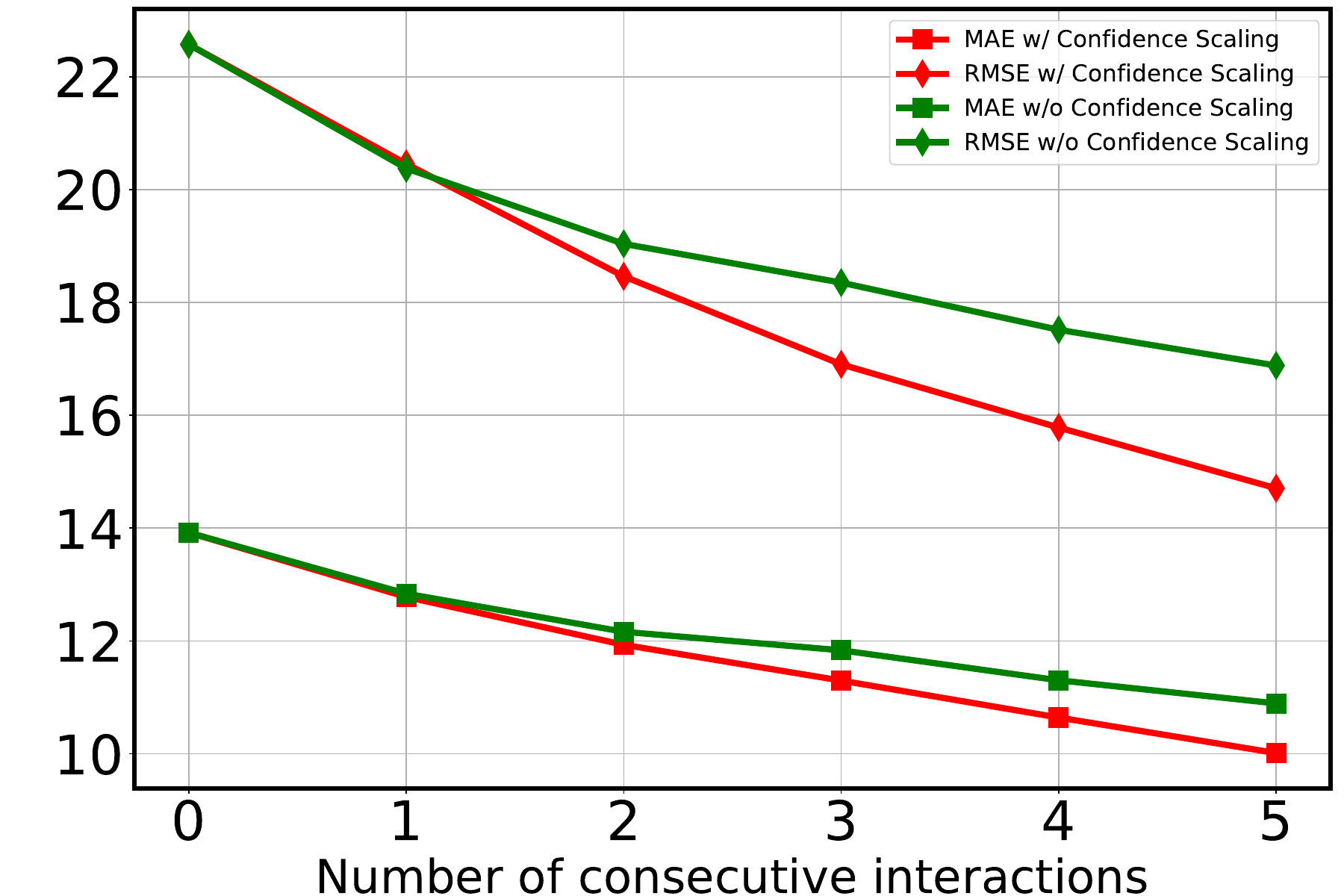} \hfill
\caption{MAE and RMSE with respect to the number of feedback iterations on \textbf{SAFECount}. We find that confidence scaling can make the adaptation smoother and improve the final result significantly.}
\label{fig:SAFECOUNT_CONFIDENCE}
\end{figure}

\def\subFigSz{0.48\linewidth} 
\begin{figure}[t] 
\centering
\makebox[\subFigSz]{\small{(a) FSC-147}} \hfill 
\makebox[\subFigSz]{\small{(b) FSCD-LVIS}}
\includegraphics[width=\subFigSz]{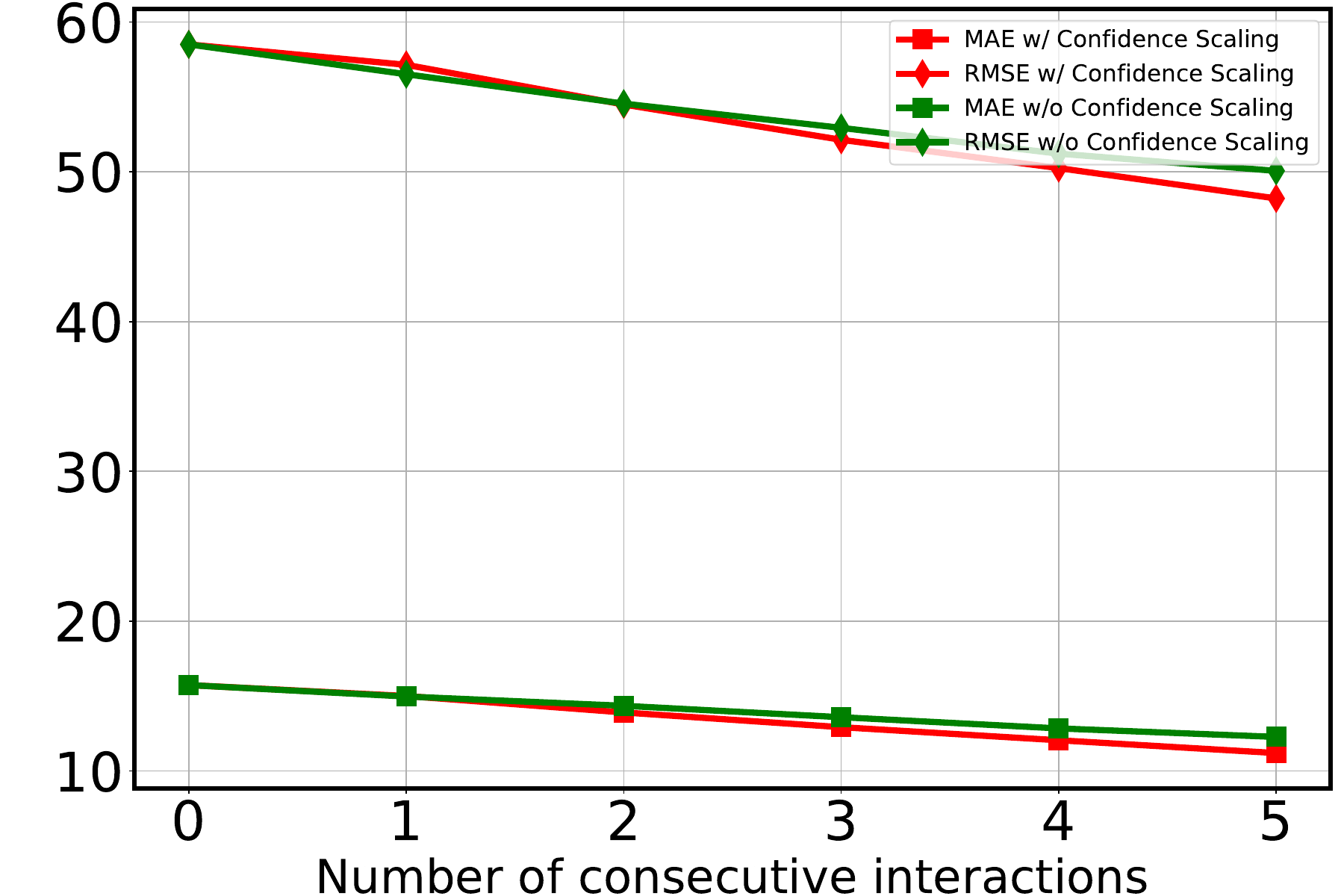} \hfill 
\includegraphics[width=\subFigSz]{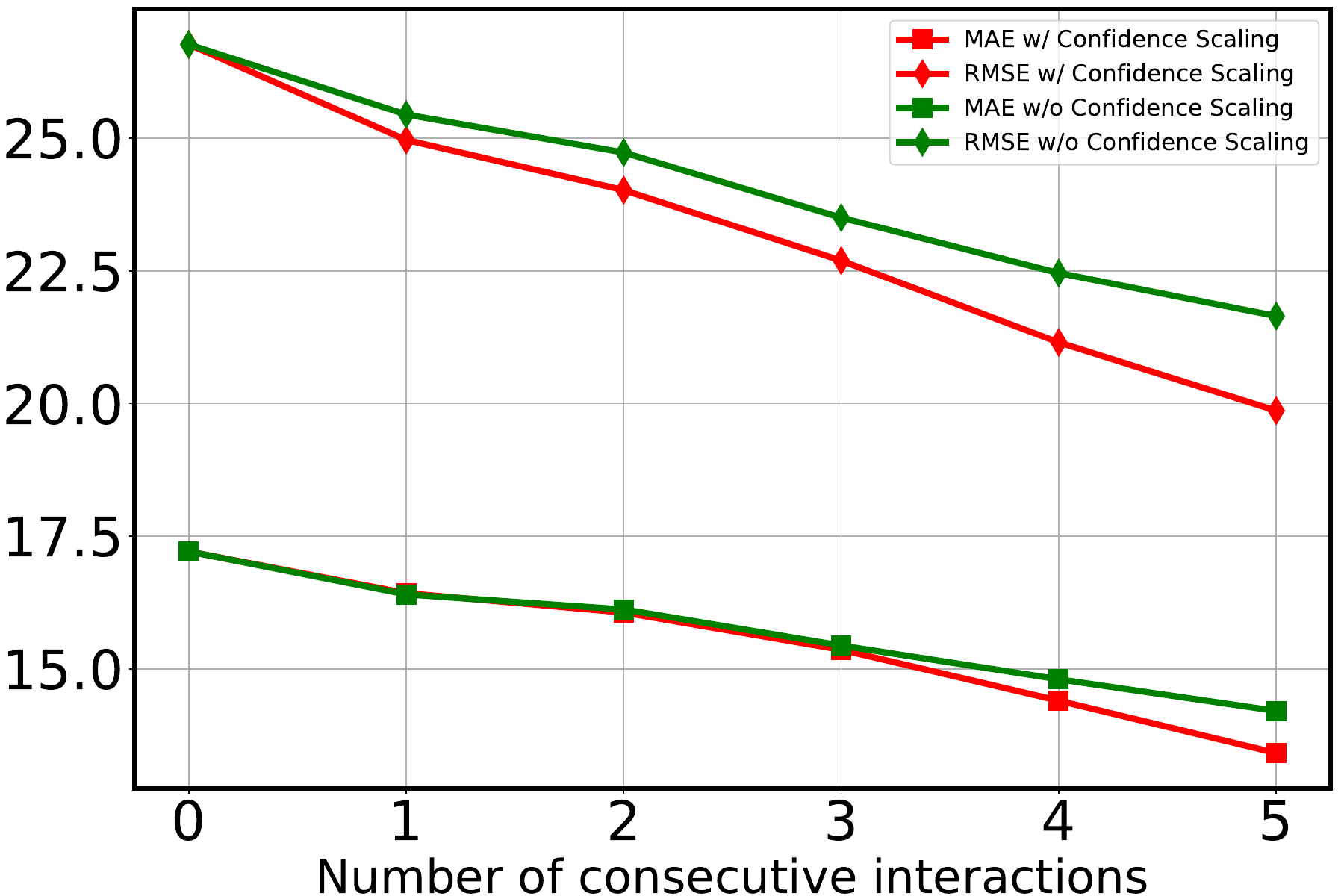} \hfill
\caption{MAE and RMSE with respect to the number of feedback iterations on \textbf{BMNet+}. We find that confidence scaling can make the adaptation smoother and improve the final result significantly.}
\label{fig:BMNET_CONFIDENCE}
\end{figure}

\subsection{Interactive Interface}
\label{sec:Interface}
\begin{figure*}[htbp]
  \centering
  \includegraphics[width=1\linewidth]{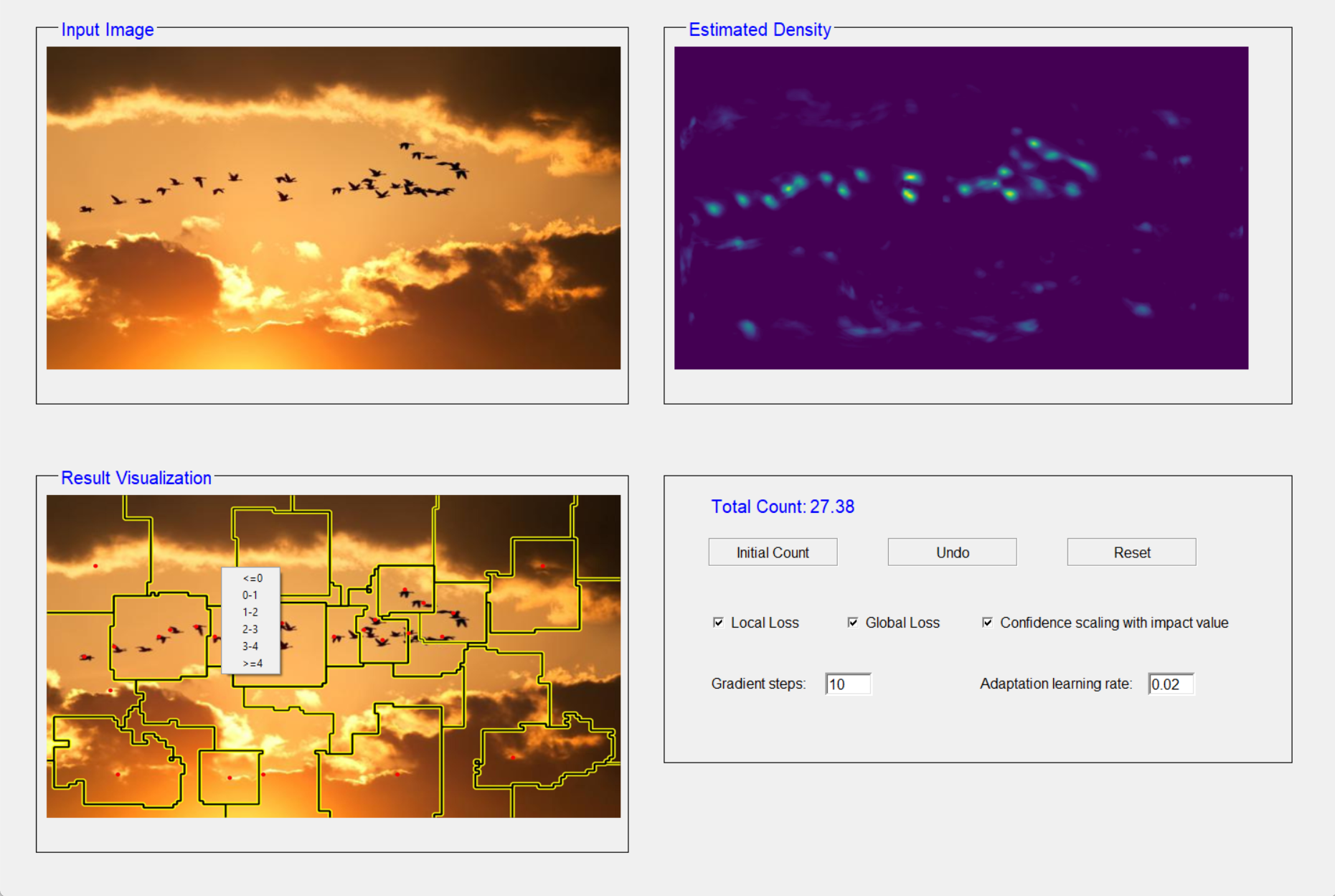}
  \caption{An illustrative graphical user interface for displaying results and collecting the user's feedback. The image is segmented into smaller regions, each region has a moderate size and small density sum. The user can provide feedback by clicking a region and selecting the count range for that region; a total of two clicks per iteration.}
  \label{fig:Interface}
\end{figure*}

The frontend interface of our interactive software is shown \Fref{fig:Interface}. In the visualization, We also provide approximate locations of the detected objects by putting some dots in the regions. The locations of these dots are found automatically, by iteratively selecting a peak of the density map and performing non-maximum suppression for the neighboring pixels. We also provide a demo video in the supplementary. In the demo video the running time for each interaction is around two seconds. This is because in the demo video one interaction includes four stages: adaptation, density map display, segmentation, and visualizing the final result(overlay the image with region boundary and approximate location for each counted object). Analysis of these stages, using images from our user study with three interactions each, shows a mean interaction time of 2.07 seconds. Breakdown: adaptation 0.52s, map display 0.50s, segmentation 0.40s, visualizing the final result 0.64s. Although segmentation takes less than a second, the full process lasts over two seconds due to the image save-load-visualize process. We aim to optimize our software for increased speed in the fut. 

\subsection{Qualitative results of refinement module.}
\label{sec:Qual1}
The qualitative results of the feature refinement are shown in \Fref{fig:FR_QUAL}. In this figure, for each example, the first row shows the initial result, and the second row shows the result after one interaction. In each row, we show the prediction, the estimated density map, the refined feature map, and the scale parameters in the refinement module. From the last three columns, we can find that the spatial-wise refinement focuses on the local error that only the parameters close to the region are updated. Thus the spatial-wise refinement contributes more to the refinement of local error. We also find that channel-wise refinement can refine the feature map globally and can correct the global error. This also explains why the channel-wise refinement contributes more to the final result, as illustrated in the refinement module's ablation study in the main paper.

\def\subFigSz{0.19\linewidth} 
\begin{figure*}[t] 
\centering
\makebox[\subFigSz]{\footnotesize{Prediction}} \hfill 
\makebox[\subFigSz]{\footnotesize{Density map}} \hfill 
\makebox[\subFigSz]{\footnotesize{Feature map}} \hfill 
\makebox[\subFigSz]{\footnotesize{Scale parameters of SP}} 
\makebox[\subFigSz]{\footnotesize{Scale parameters of CH}} 
\includegraphics[width=\subFigSz]{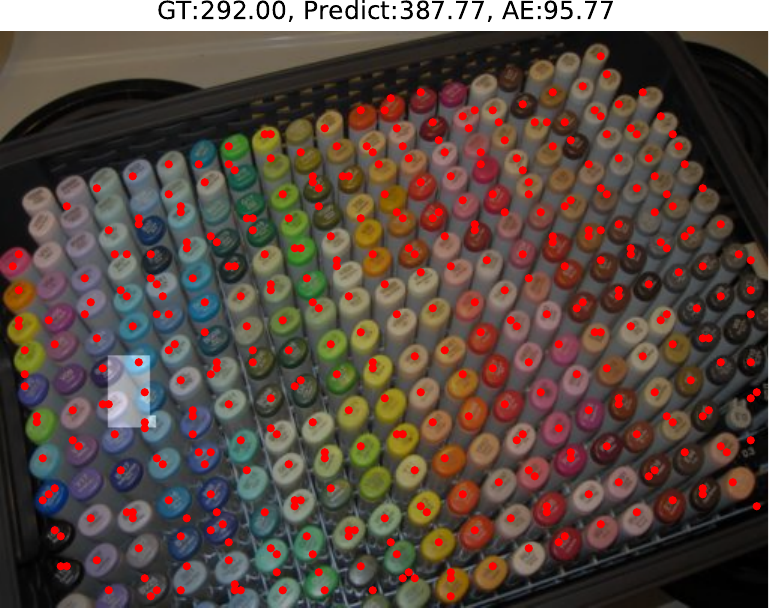} \hfill 
\includegraphics[width=\subFigSz]{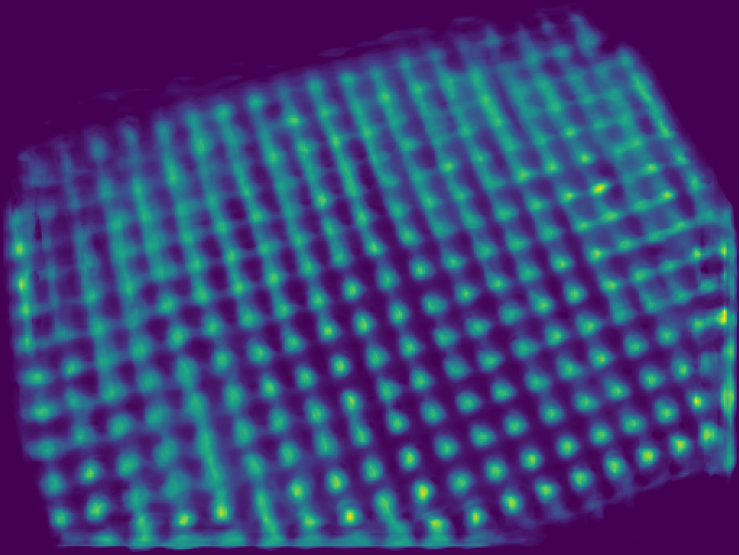} \hfill 
\includegraphics[width=\subFigSz]{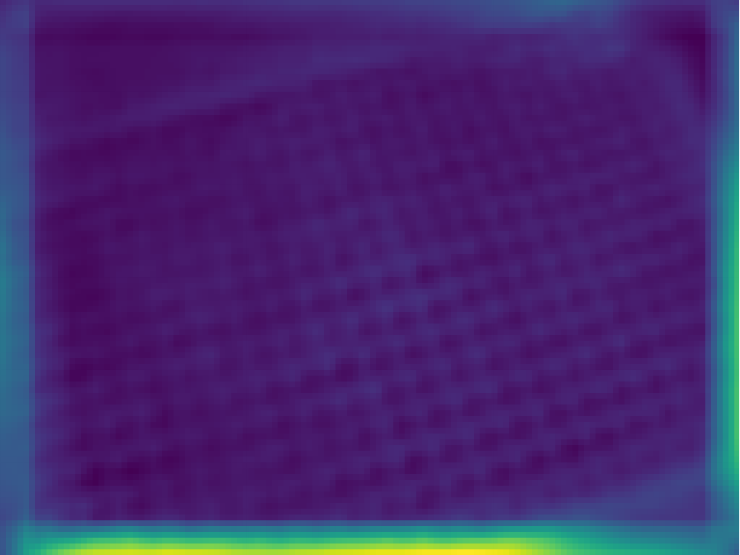} \hfill
\includegraphics[width=\subFigSz]{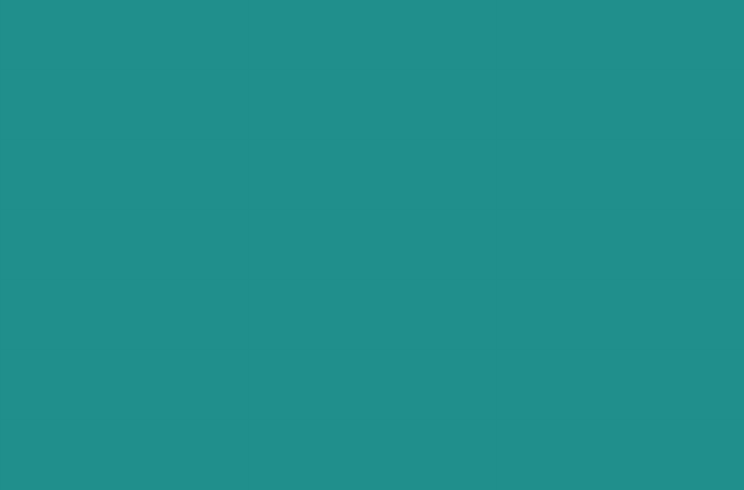} \hfill
\includegraphics[width=\subFigSz]{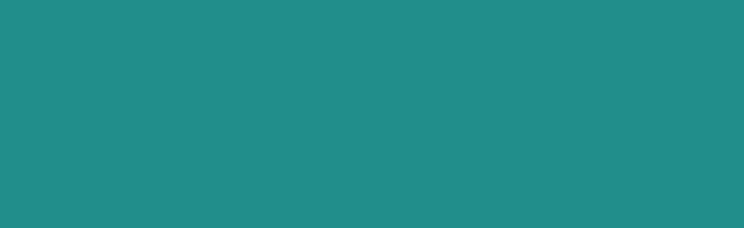} \\
\includegraphics[width=\subFigSz]{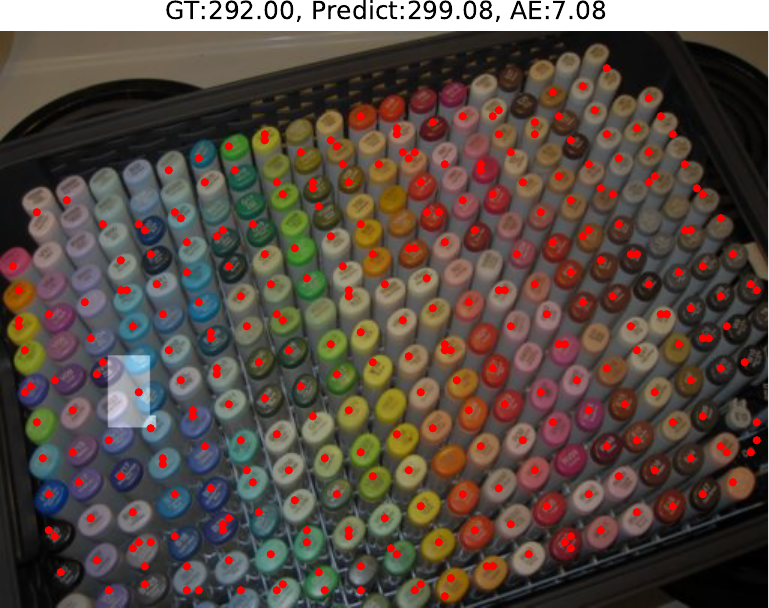} \hfill 
\includegraphics[width=\subFigSz]{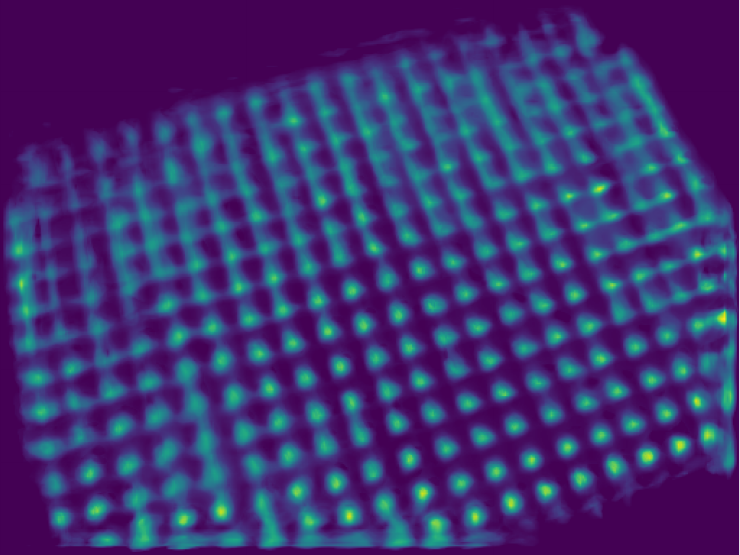} \hfill 
\includegraphics[width=\subFigSz]{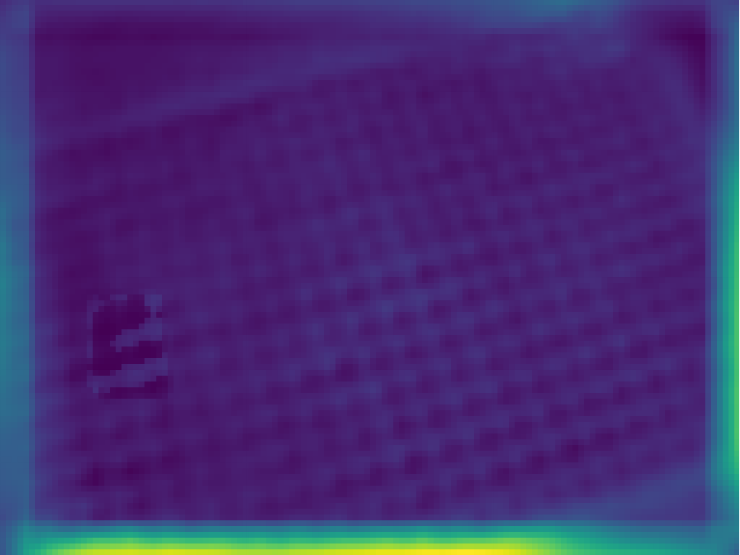} \hfill 
\includegraphics[width=\subFigSz]{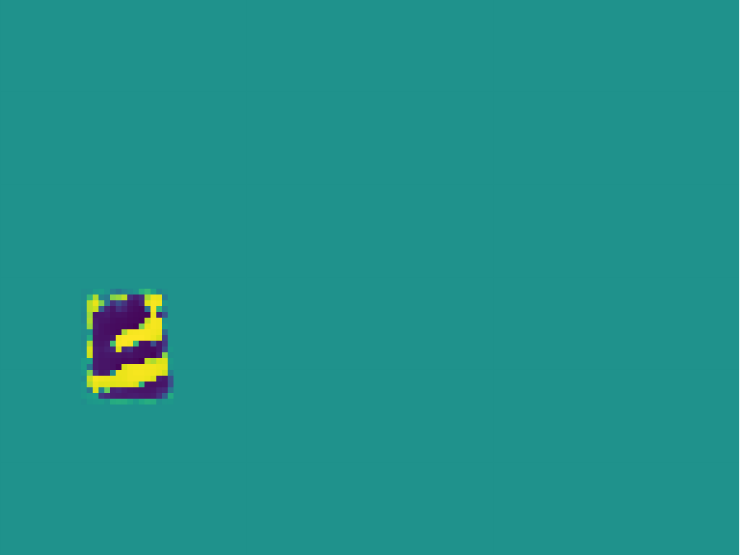} \hfill
\includegraphics[width=\subFigSz]{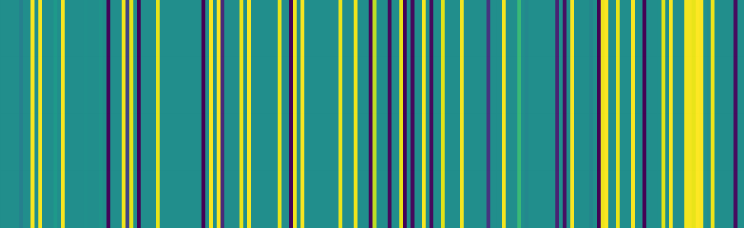} 
\makebox[\subFigSz]{\small{(a) Example 1}}
\vskip 0.05in
\includegraphics[width=\subFigSz]{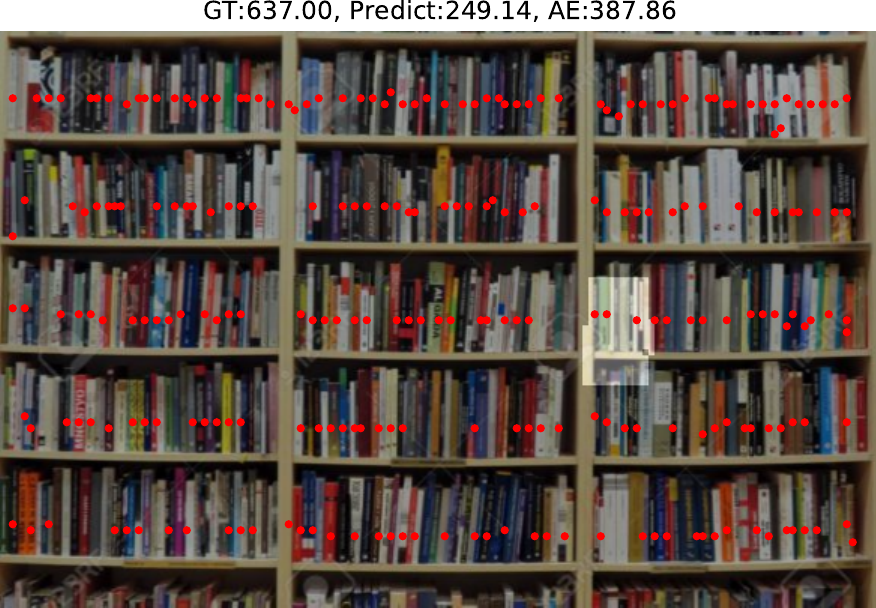} \hfill 
\includegraphics[width=\subFigSz]{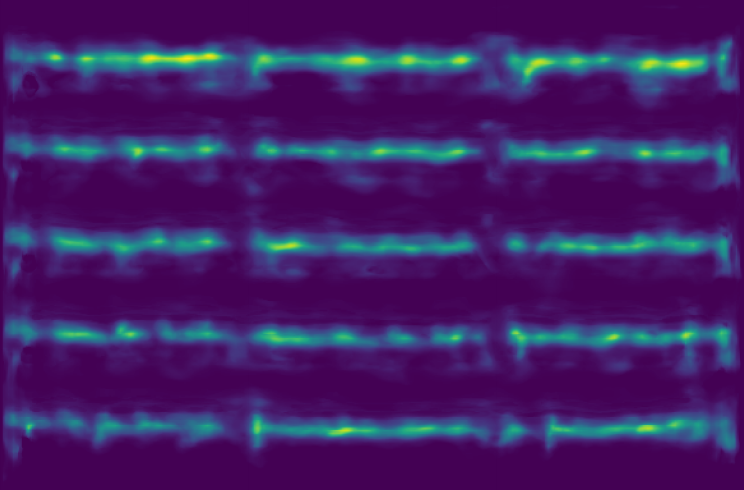} \hfill 
\includegraphics[width=\subFigSz]{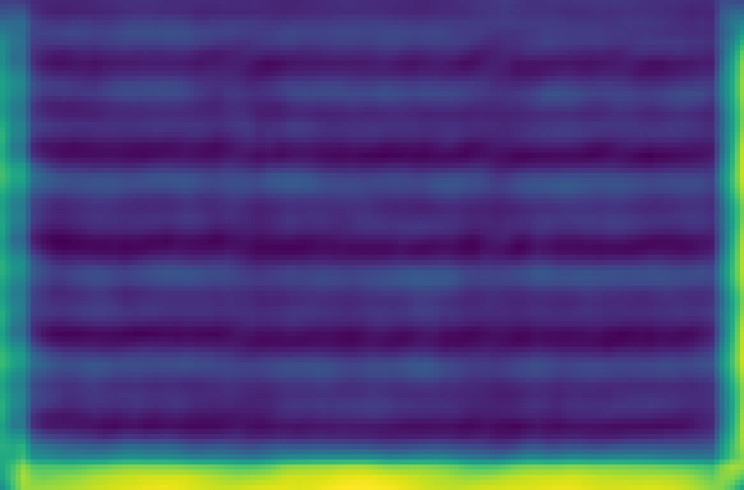} \hfill
\includegraphics[width=\subFigSz]{SupFigures/Qual/FRM_BEFORE_SP_SCALE.pdf} \hfill
\includegraphics[width=\subFigSz]{SupFigures/Qual/FRM_BEFORE_CH_SCALE.pdf} \\
\includegraphics[width=\subFigSz]{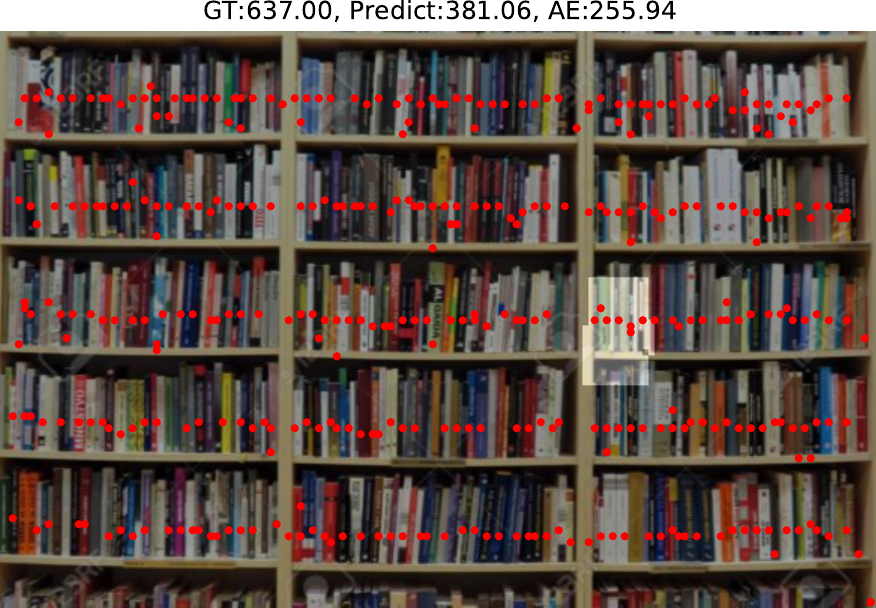} \hfill 
\includegraphics[width=\subFigSz]{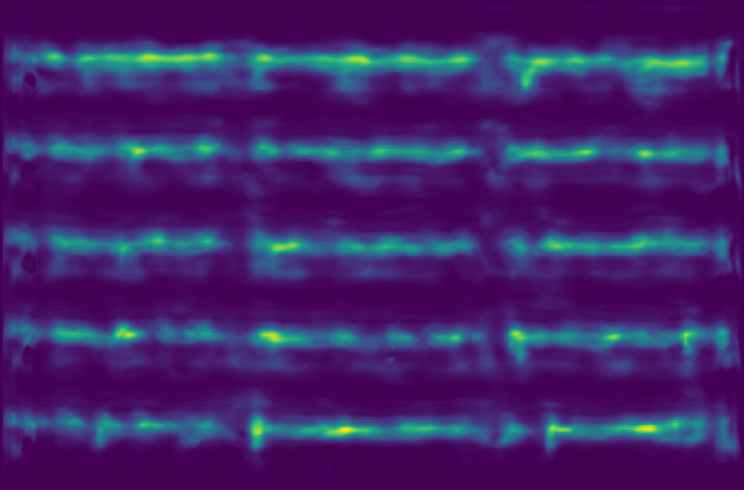} \hfill 
\includegraphics[width=\subFigSz]{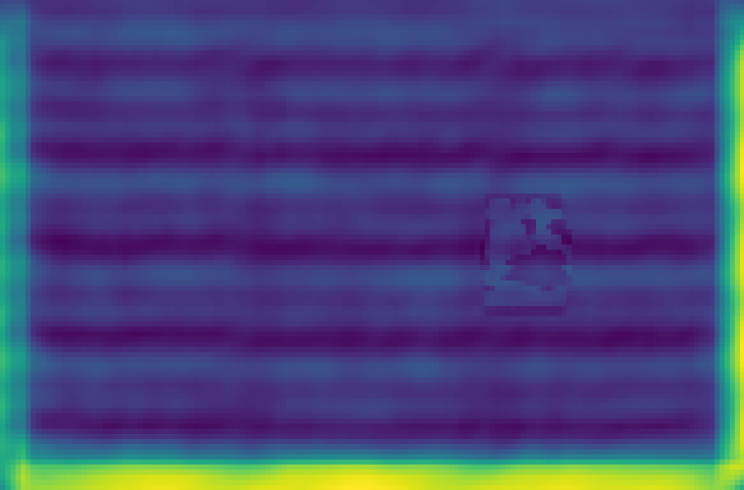} \hfill 
\includegraphics[width=\subFigSz]{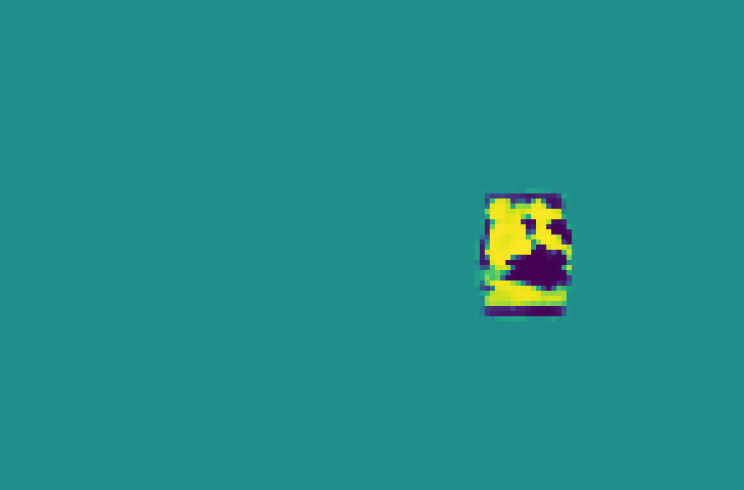} \hfill
\includegraphics[width=\subFigSz]{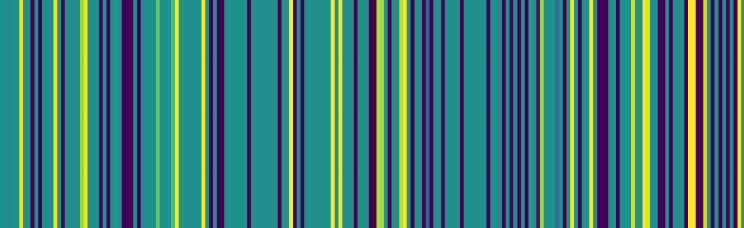} 
\makebox[\subFigSz]{\small{(b) Example 2}}
\vskip 0.05in
\includegraphics[width=\subFigSz]{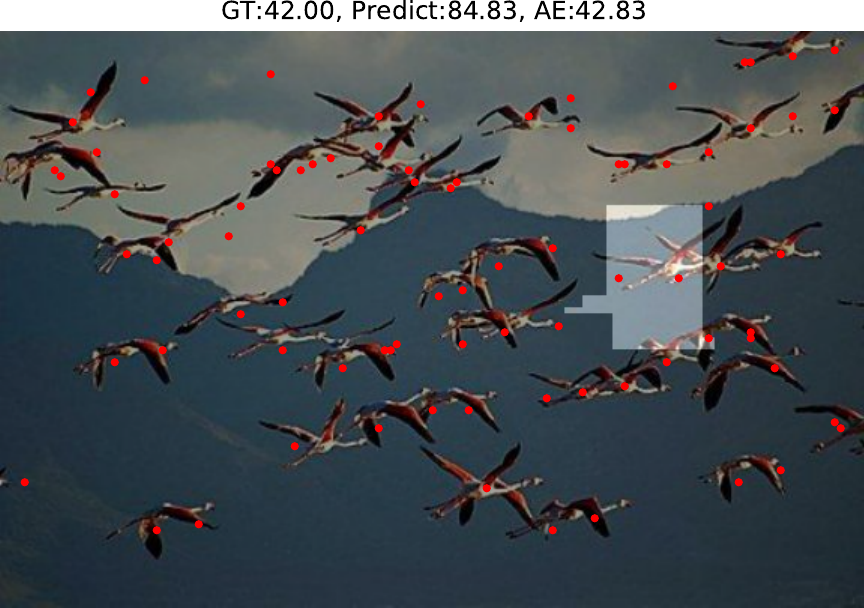} \hfill 
\includegraphics[width=\subFigSz]{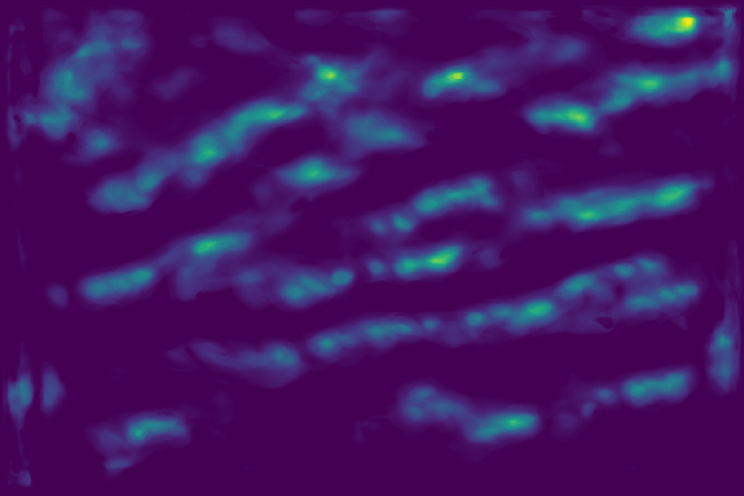} \hfill 
\includegraphics[width=\subFigSz]{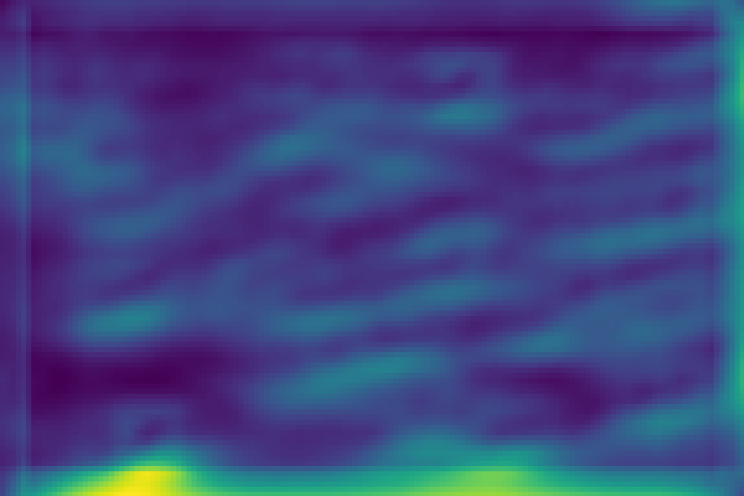} \hfill
\includegraphics[width=\subFigSz]{SupFigures/Qual/FRM_BEFORE_SP_SCALE.pdf} \hfill
\includegraphics[width=\subFigSz]{SupFigures/Qual/FRM_BEFORE_CH_SCALE.pdf} \\
\includegraphics[width=\subFigSz]{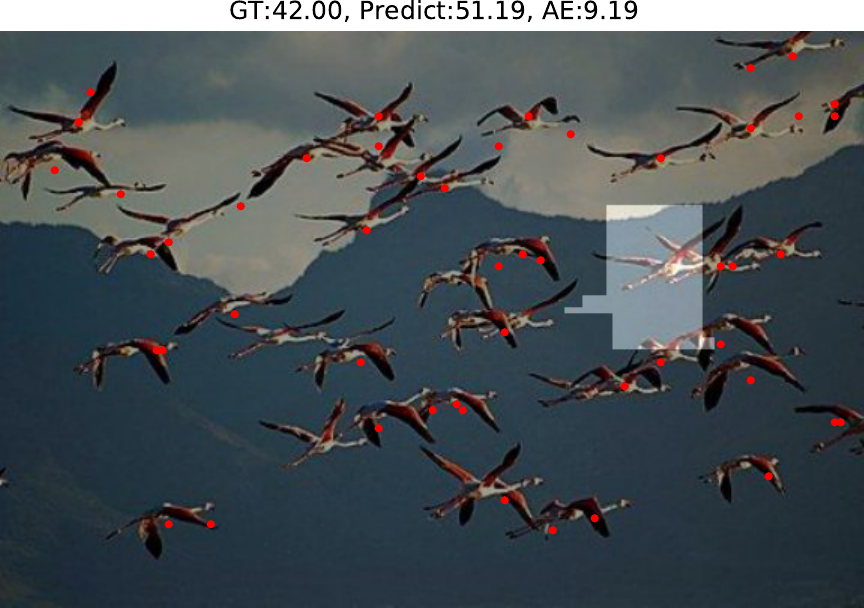} \hfill 
\includegraphics[width=\subFigSz]{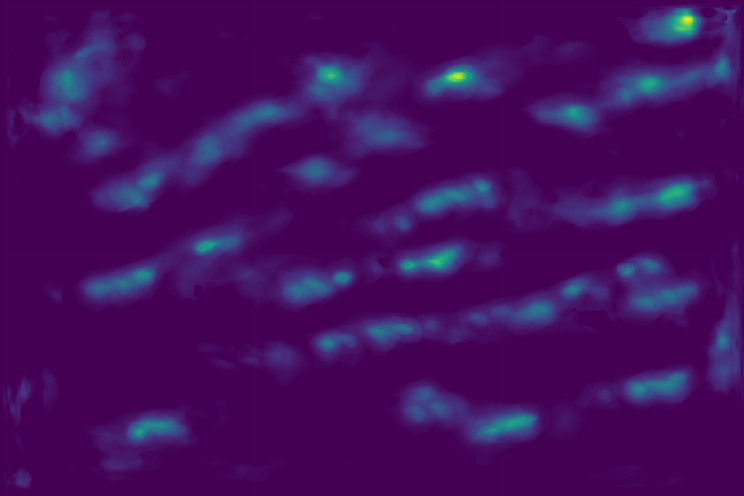} \hfill 
\includegraphics[width=\subFigSz]{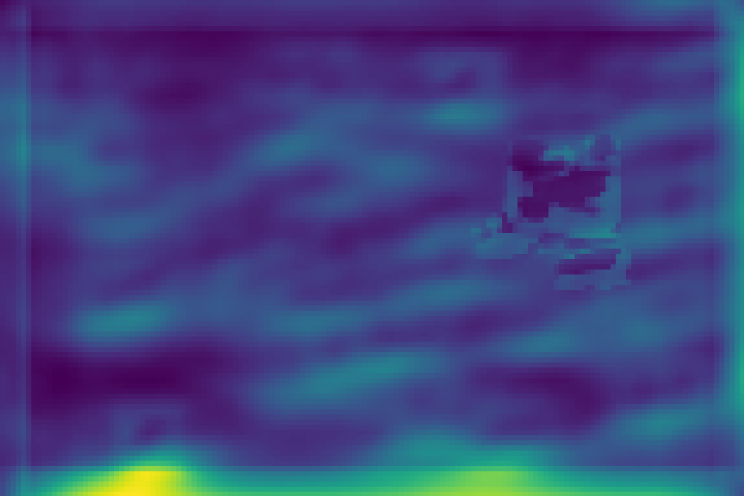} \hfill 
\includegraphics[width=\subFigSz]{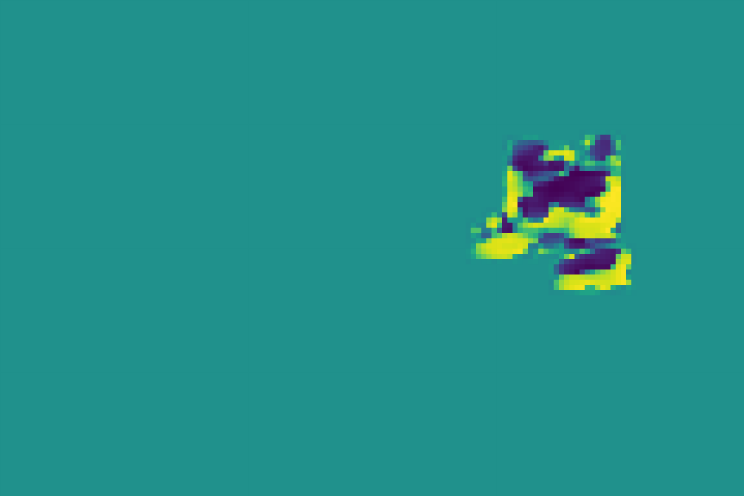} \hfill
\includegraphics[width=\subFigSz]{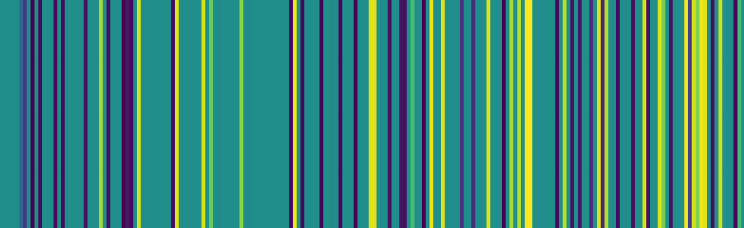} 
\makebox[\subFigSz]{\small{(c) Example 3}}

\caption{Qualitative results of the feature refinement. These three examples are from FSC-147 with FamNet as the visual counter. For each example, the first row is before having any interactive feedback, and the second row is after utilizing one interactive feedback. The first column shows the ground truth, prediction, and absolute error. The second column shows the estimated density map. The third column shows the feature map that the refinement module refines. We can know how the feature refinement refines the feature map from this column. The last two columns show the scale parameters of spatial-wise refinement and channel-wise refinement in the refinement module.}
\label{fig:FR_QUAL}
\end{figure*}

\subsection{Additional Qualitative results.}
\label{sec:Qual2}
Additional qualitative results on FSC-147 with FamNet is shown in \Fref{fig:ADDITIONAL_QUAL1} and \Fref{fig:ADDITIONAL_QUAL2}.

\subsection{Limitation and future work}
\label{sec:Limitation}
Our approach has several limitations. First, the user's feedback is for the entire region, not individual objects. Second, the specified count is a range, not a precise number. Third, local adaptation may improve global error, due to the inconsistency between local and global errors. Despite these limitations, the proposed method provides a practical way for the user to provide feedback and reduce counting errors in most cases. Also important is the availability of an intuitive graphical user interface for the user to decide whether to trust the automated counting results before and after the adaptation. 

In this work, we aim for a system that reduces the user's burden so that the user is not asked to delineate or localize objects. But we envision that localizing an object and delineating its spatial extent would be a stronger form of supervision, and it would be necessary for certain situations. This will be explored in our future work.
\def\subFigSz{0.24\linewidth} 
\begin{figure*}[] 
\centering
\makebox[0.24\linewidth]{\small{Before Click 1}} \hfill 
\makebox[0.24\linewidth]{\small{After Click 1}} \hfill 
\makebox[0.24\linewidth]{\small{Before Click 2}} \hfill 
\makebox[0.24\linewidth]{\small{After Click 2}} \hfill 
\includegraphics[width=\subFigSz]{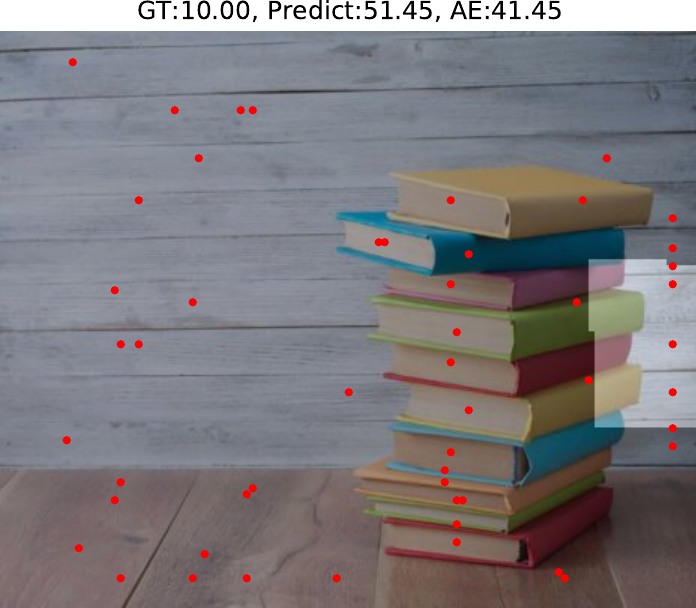} \hfill 
\includegraphics[width=\subFigSz]{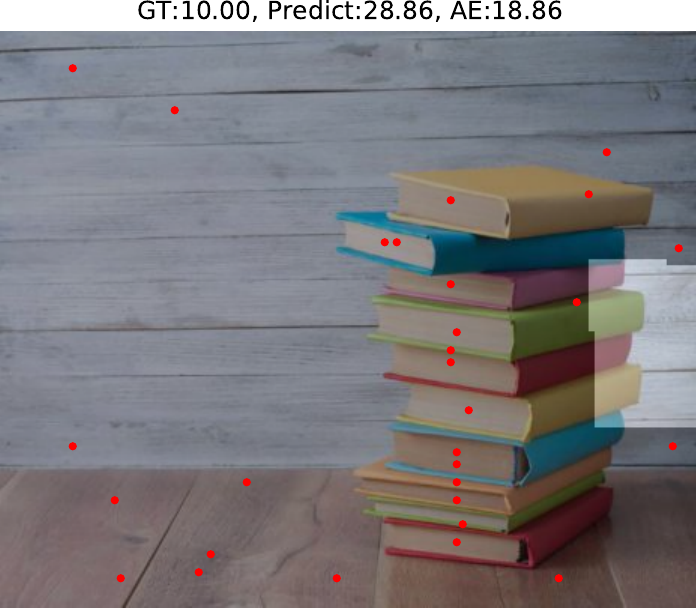} \hfill 
\includegraphics[width=\subFigSz]{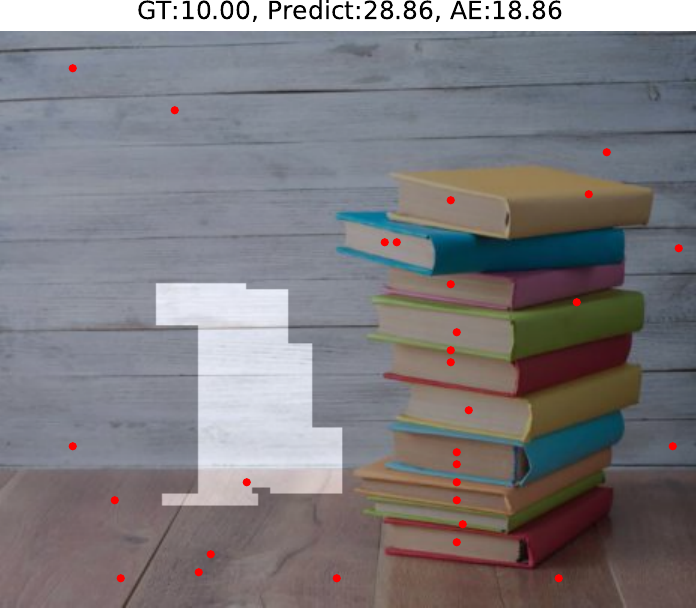} \hfill 
\includegraphics[width=\subFigSz]{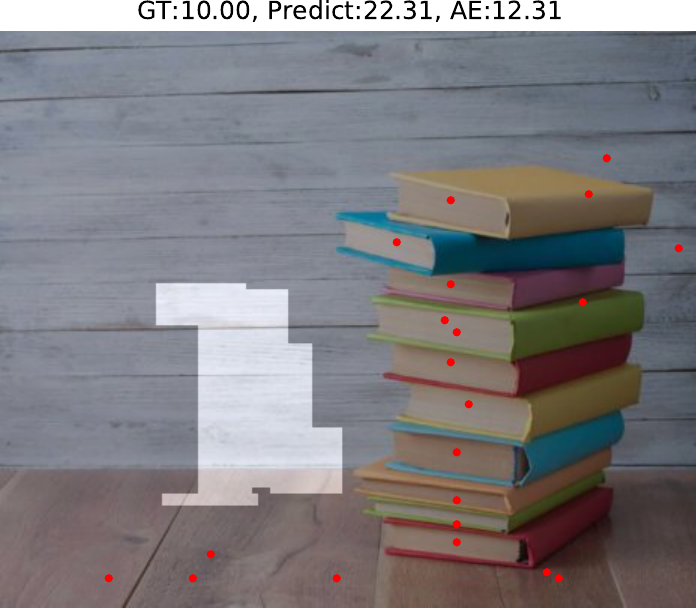}\\ 
\vskip 0.05in
\includegraphics[width=\subFigSz]{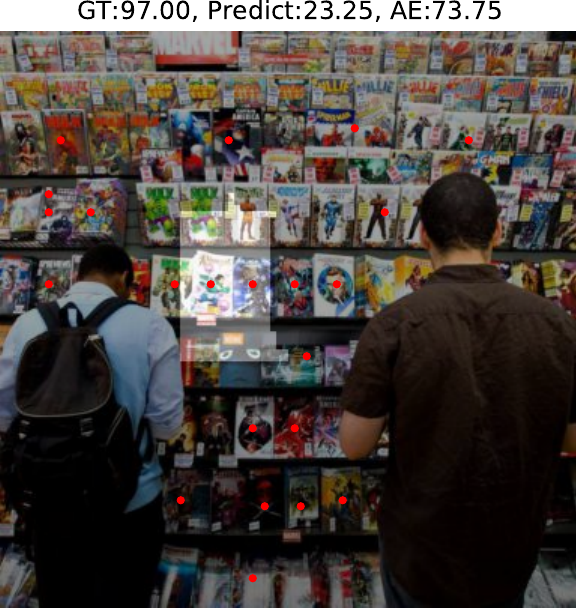} \hfill 
\includegraphics[width=\subFigSz]{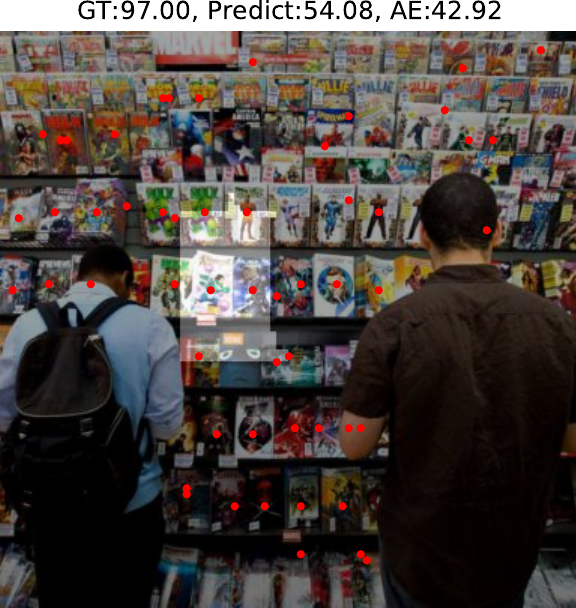} \hfill 
\includegraphics[width=\subFigSz]{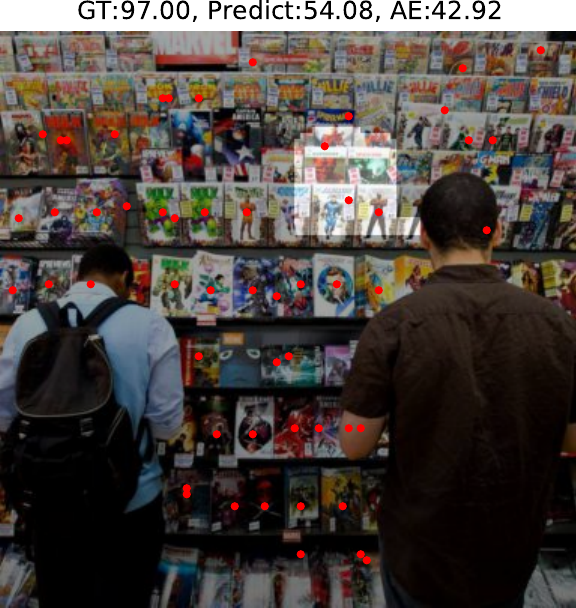} \hfill 
\includegraphics[width=\subFigSz]{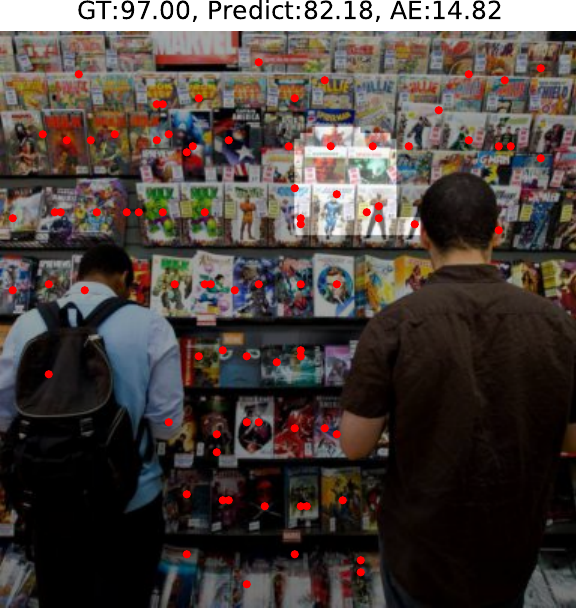}\\ 
\vskip 0.05in
\includegraphics[width=\subFigSz]{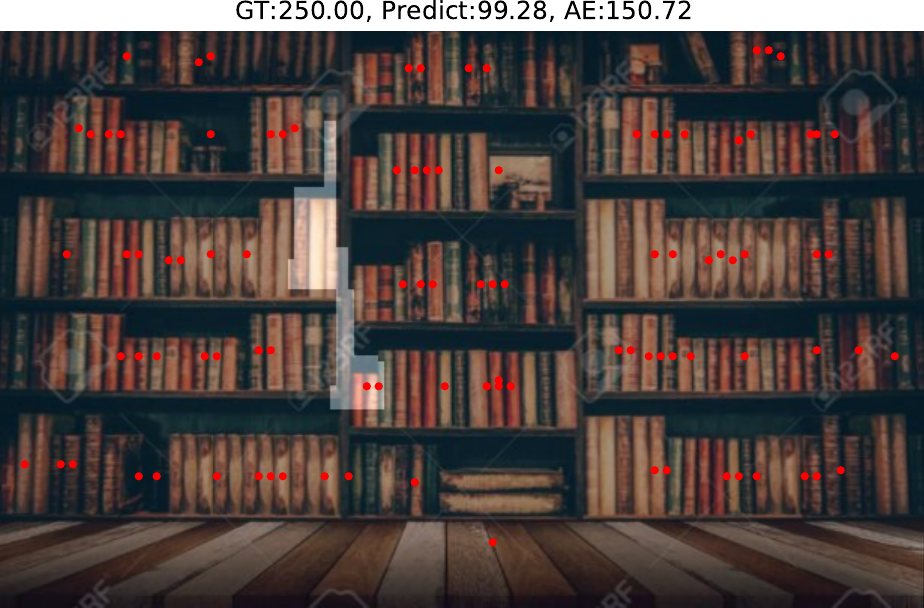} \hfill 
\includegraphics[width=\subFigSz]{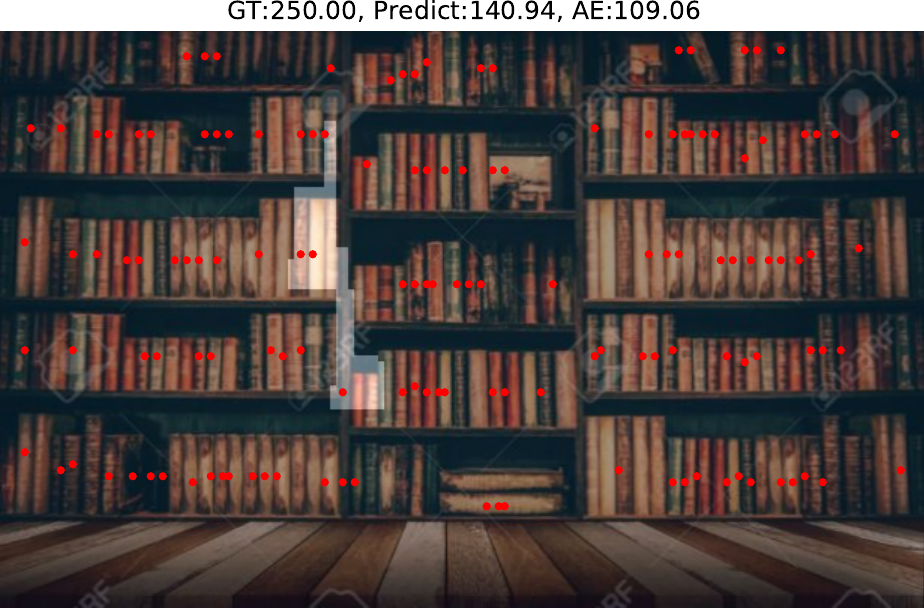} \hfill 
\includegraphics[width=\subFigSz]{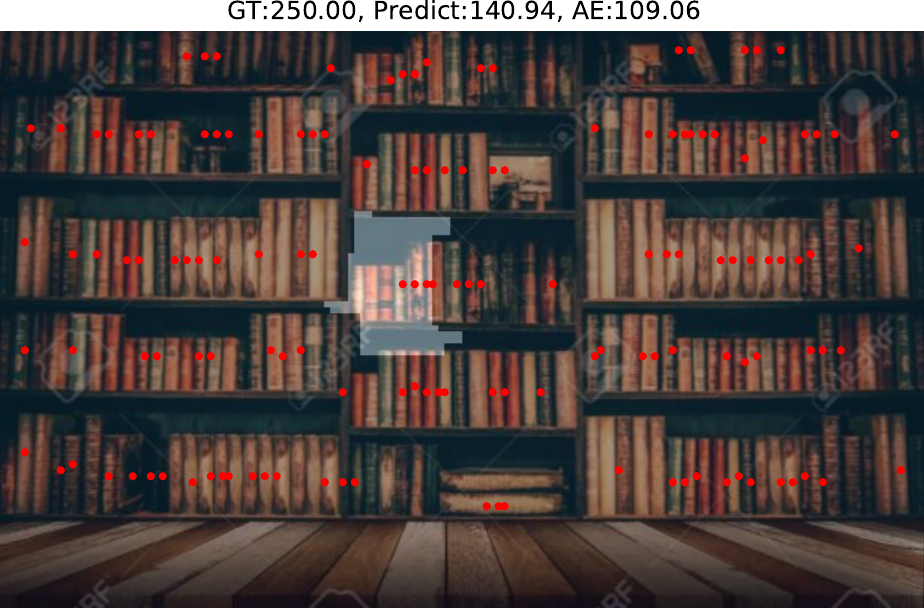} \hfill 
\includegraphics[width=\subFigSz]{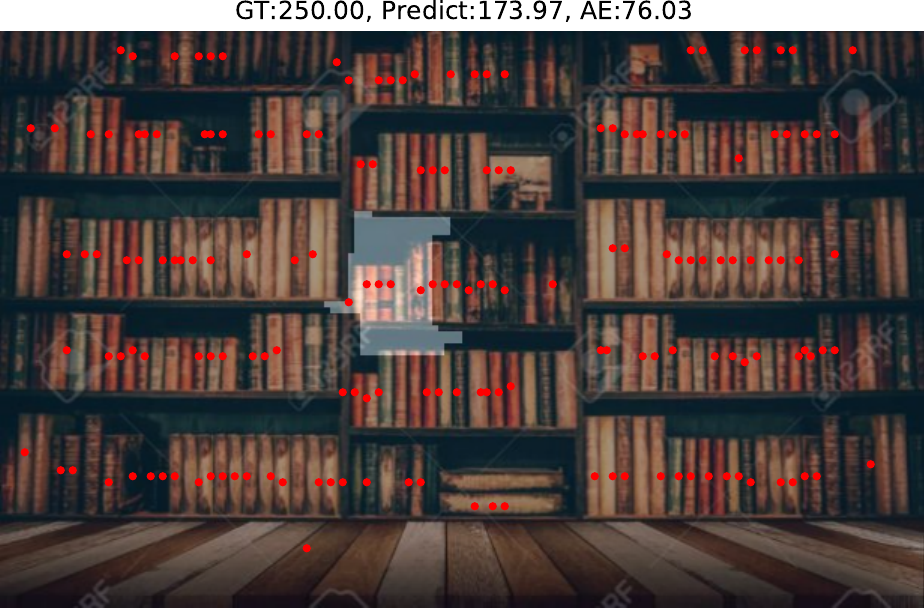}\\ 
\vskip 0.05in
\includegraphics[width=\subFigSz]{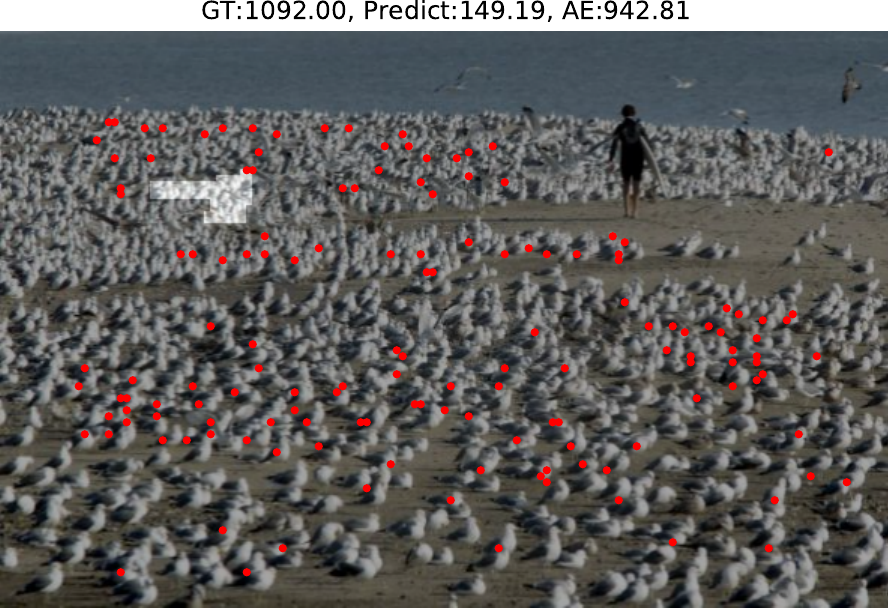} \hfill 
\includegraphics[width=\subFigSz]{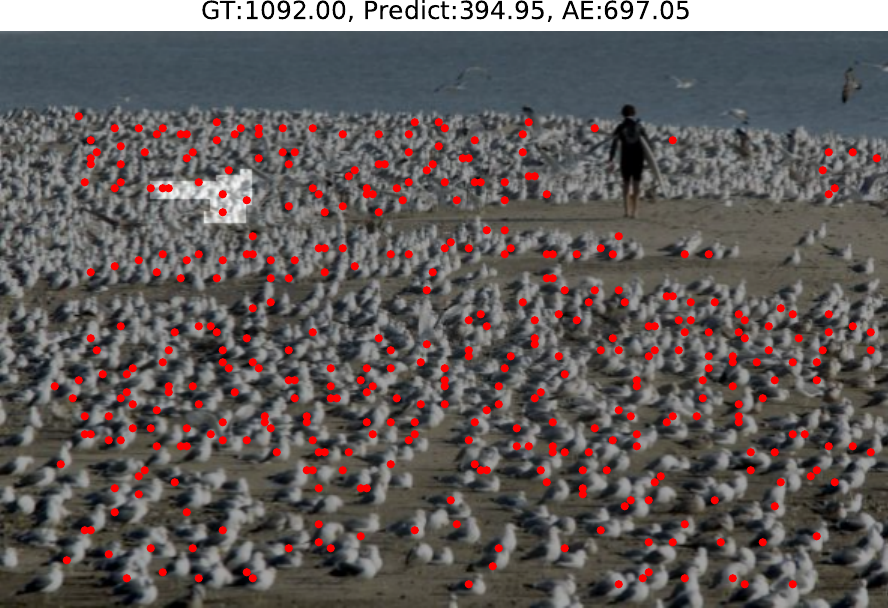} \hfill 
\includegraphics[width=\subFigSz]{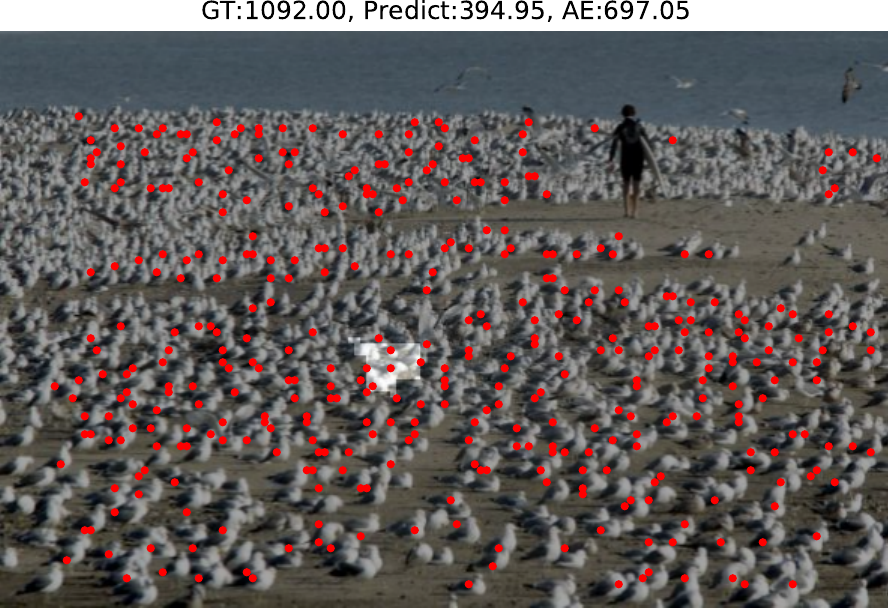} \hfill 
\includegraphics[width=\subFigSz]{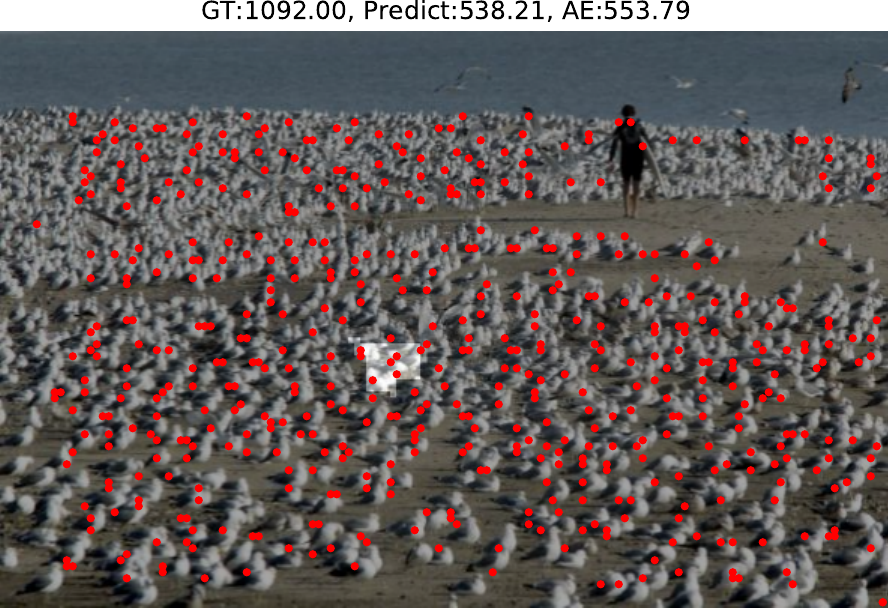}\\ 
\vskip 0.05in
\includegraphics[width=\subFigSz]{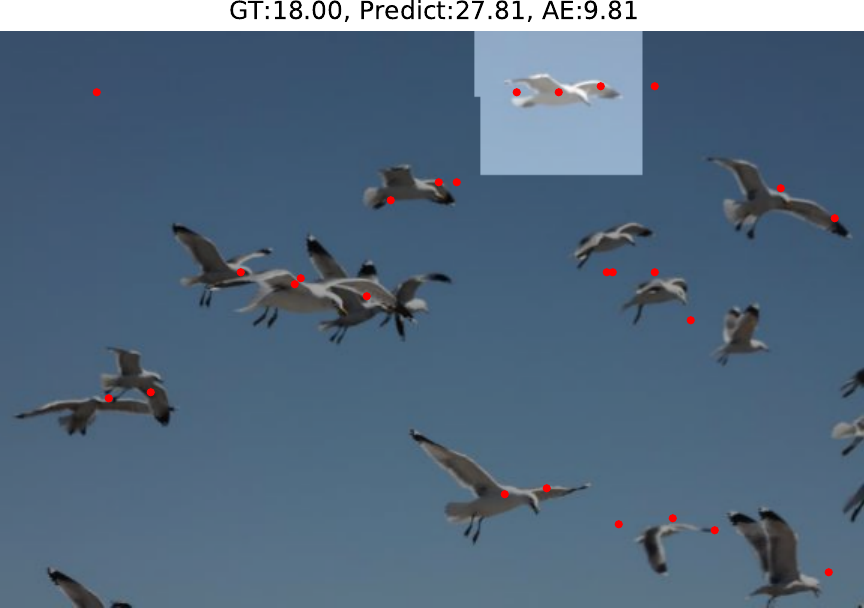} \hfill 
\includegraphics[width=\subFigSz]{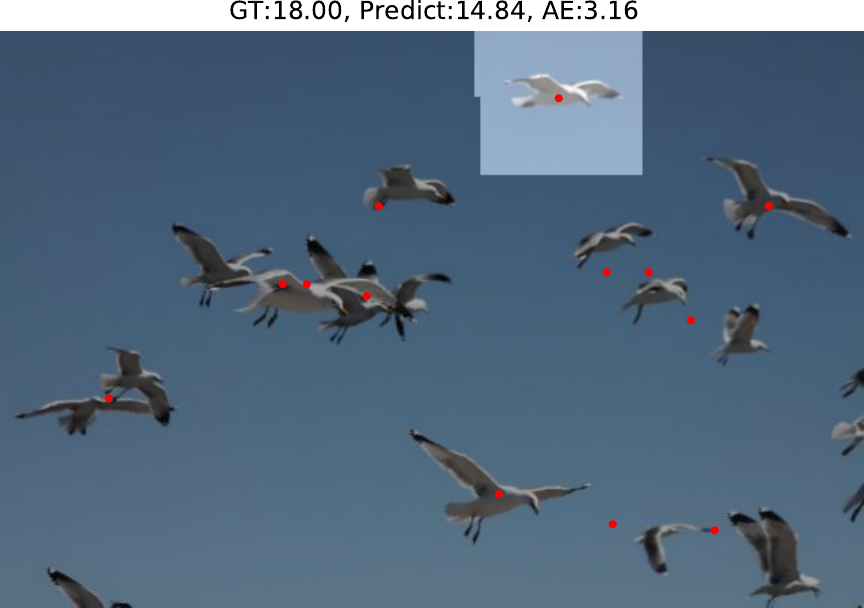} \hfill 
\includegraphics[width=\subFigSz]{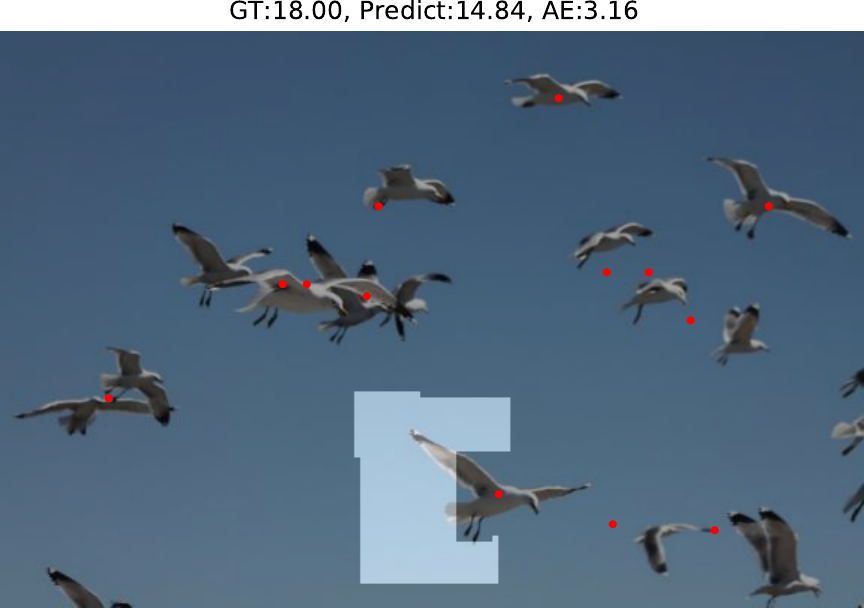} \hfill 
\includegraphics[width=\subFigSz]{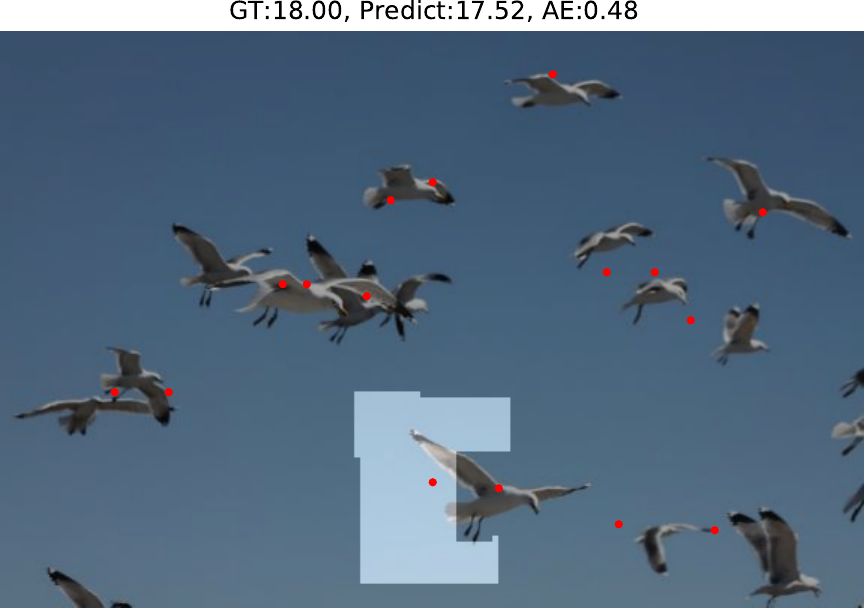}\\ 
\vskip 0.05in
\includegraphics[width=\subFigSz]{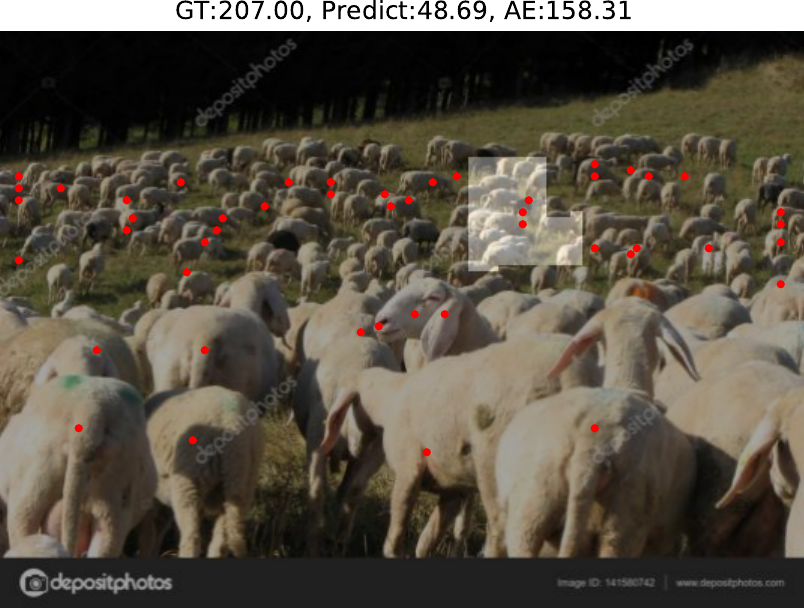} \hfill 
\includegraphics[width=\subFigSz]{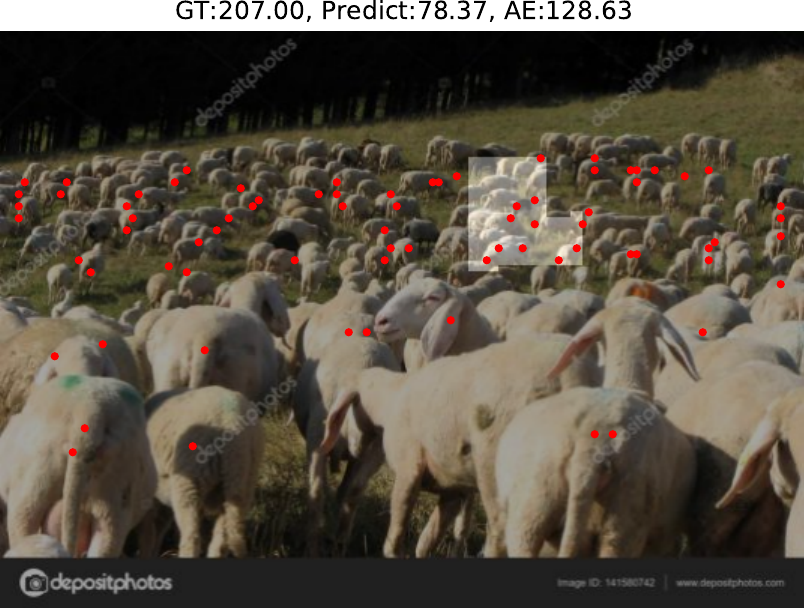} \hfill 
\includegraphics[width=\subFigSz]{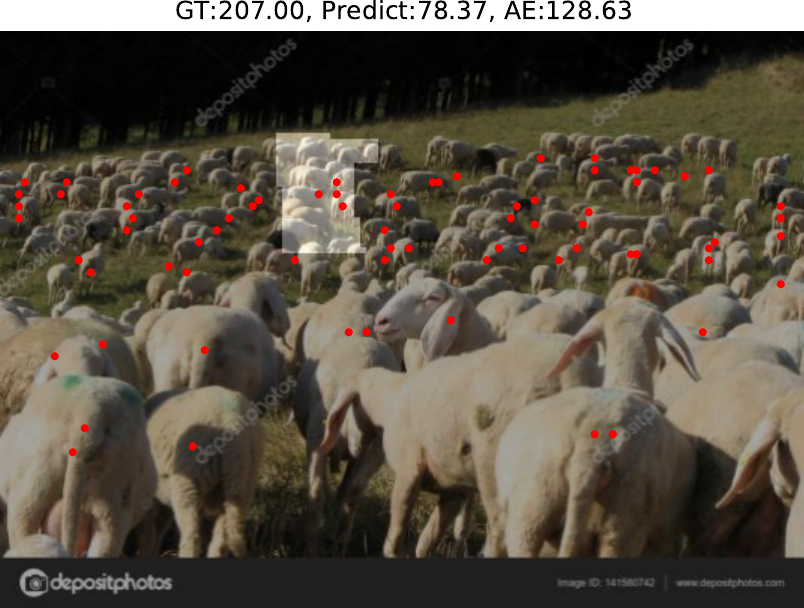} \hfill 
\includegraphics[width=\subFigSz]{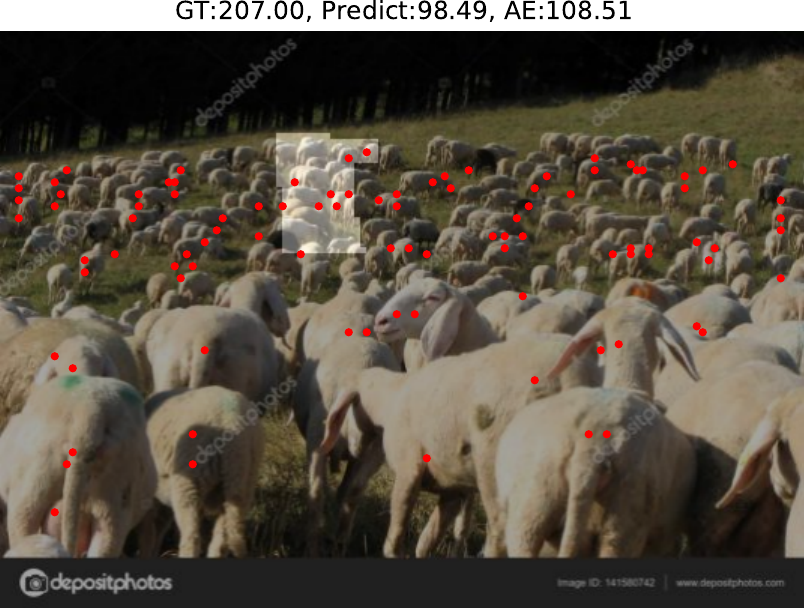}\\ 
\caption{Additional qualitative results. The examples are from FSC-147 with FamNet as the visual counter. The brighter region is the selected region, and the red dot is the approximate location of each region generated by peak selection and non-maximum suppression on each region. Our approach can improve the counting result locally(the selected region) and globally(the whole image).}
\label{fig:ADDITIONAL_QUAL1}
\end{figure*}

\def\subFigSz{0.24\linewidth} 
\begin{figure*}[] 
\centering
\makebox[0.24\linewidth]{\small{Before Click 1}} \hfill 
\makebox[0.24\linewidth]{\small{After Click 1}} \hfill 
\makebox[0.24\linewidth]{\small{Before Click 2}} \hfill 
\makebox[0.24\linewidth]{\small{After Click 2}} \hfill 
\includegraphics[width=\subFigSz]{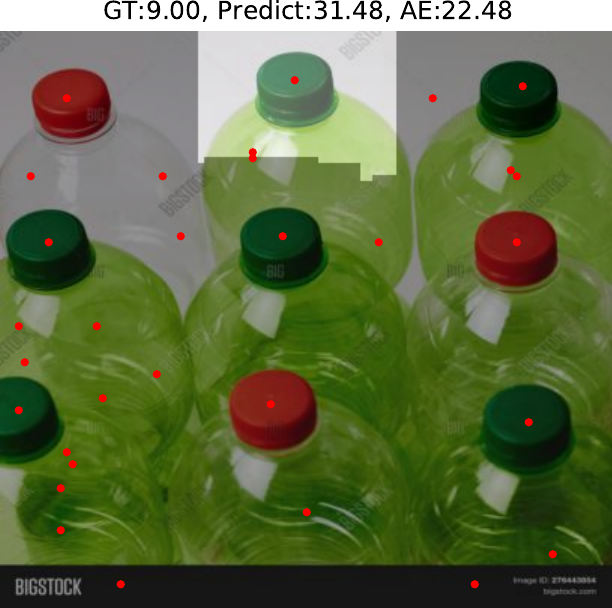} \hfill 
\includegraphics[width=\subFigSz]{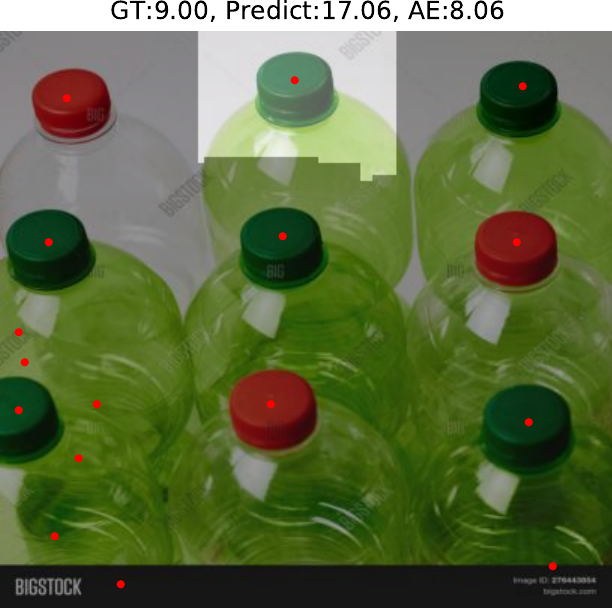} \hfill 
\includegraphics[width=\subFigSz]{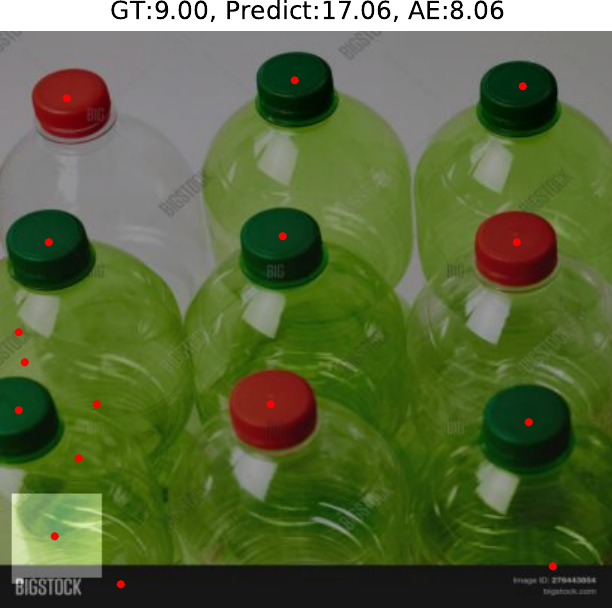} \hfill 
\includegraphics[width=\subFigSz]{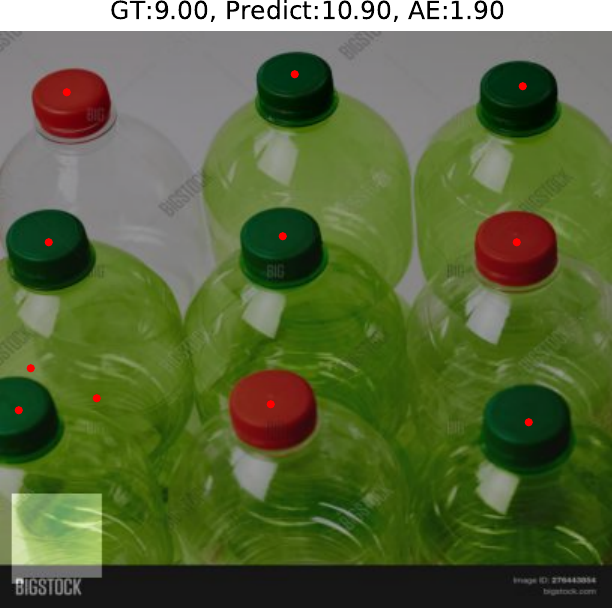}\\ 
\vskip 0.05in
\includegraphics[width=\subFigSz]{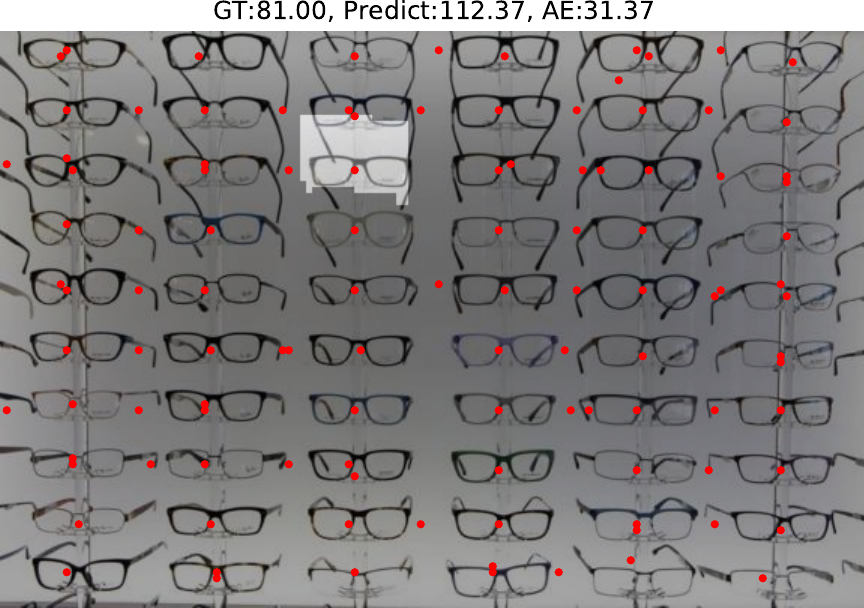} \hfill 
\includegraphics[width=\subFigSz]{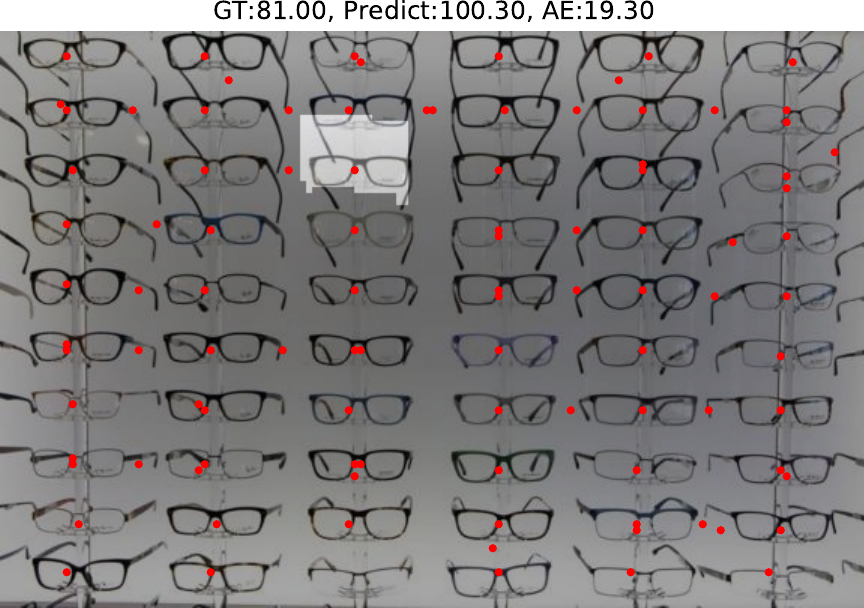} \hfill 
\includegraphics[width=\subFigSz]{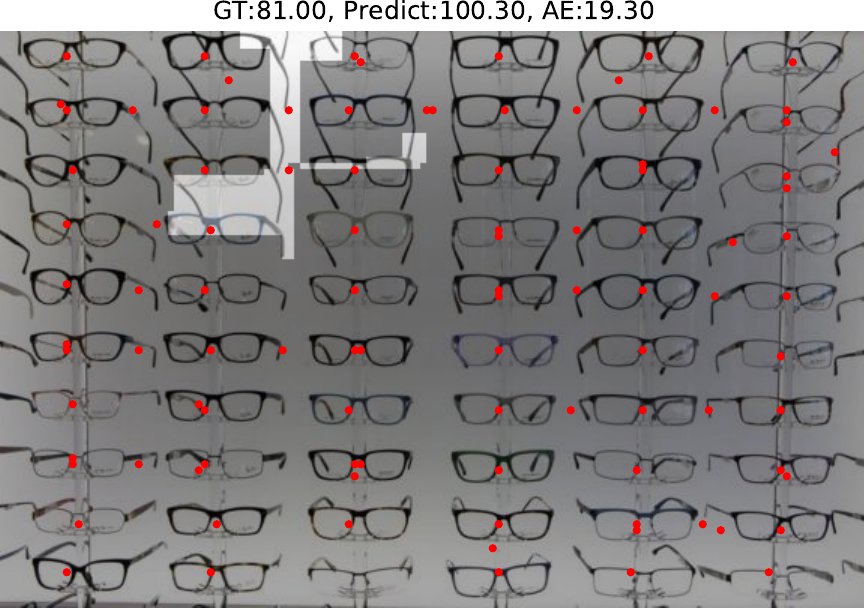} \hfill 
\includegraphics[width=\subFigSz]{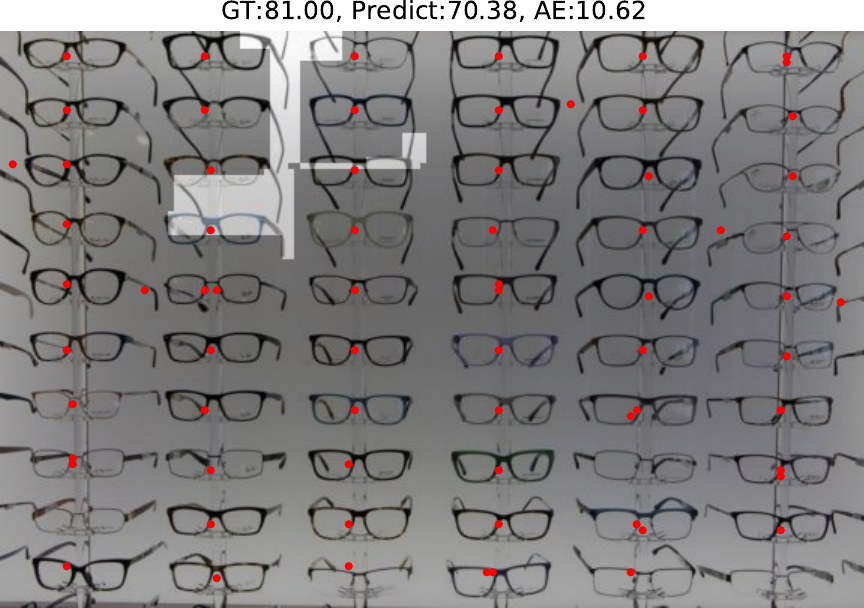}\\ 
\vskip 0.05in
\includegraphics[width=\subFigSz]{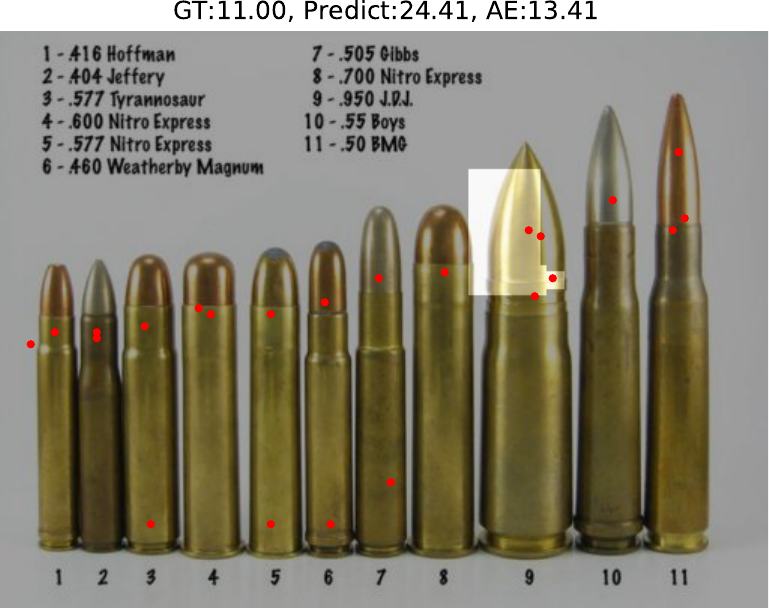} \hfill 
\includegraphics[width=\subFigSz]{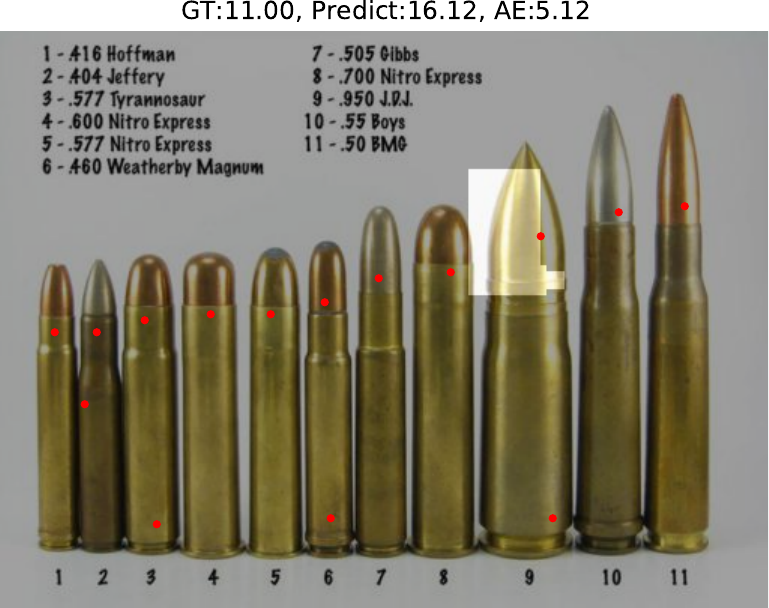} \hfill 
\includegraphics[width=\subFigSz]{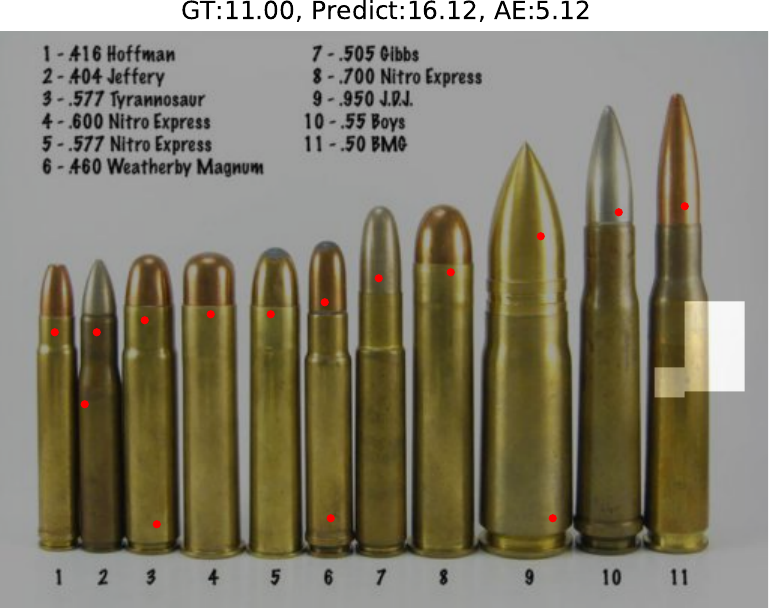} \hfill 
\includegraphics[width=\subFigSz]{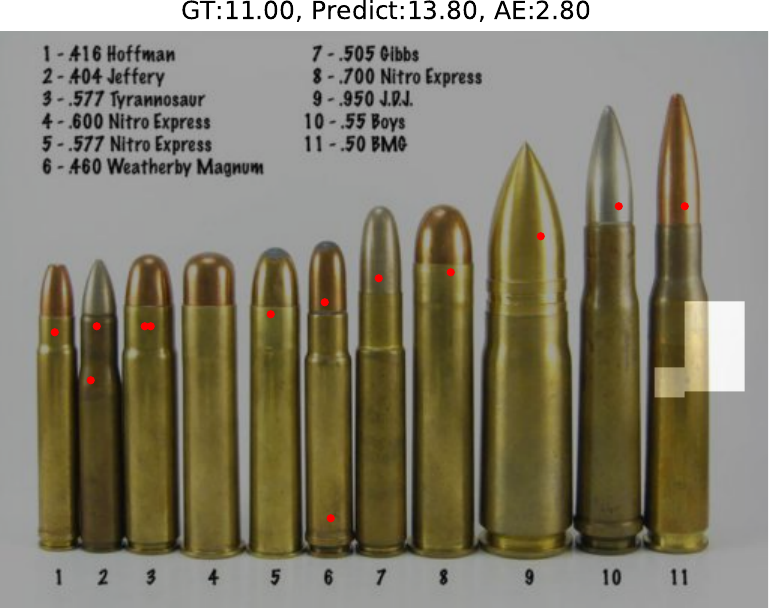}\\ 
\vskip 0.05in
\includegraphics[width=\subFigSz]{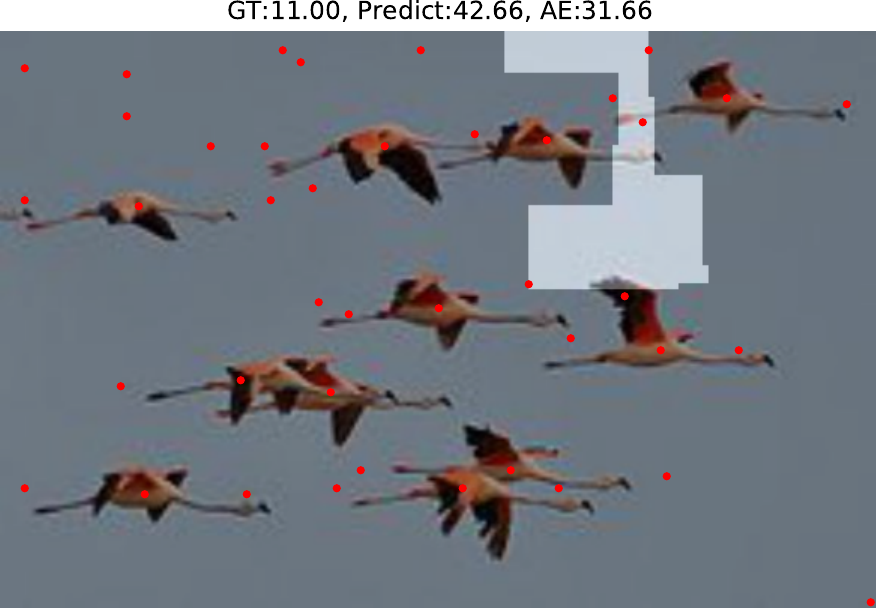} \hfill 
\includegraphics[width=\subFigSz]{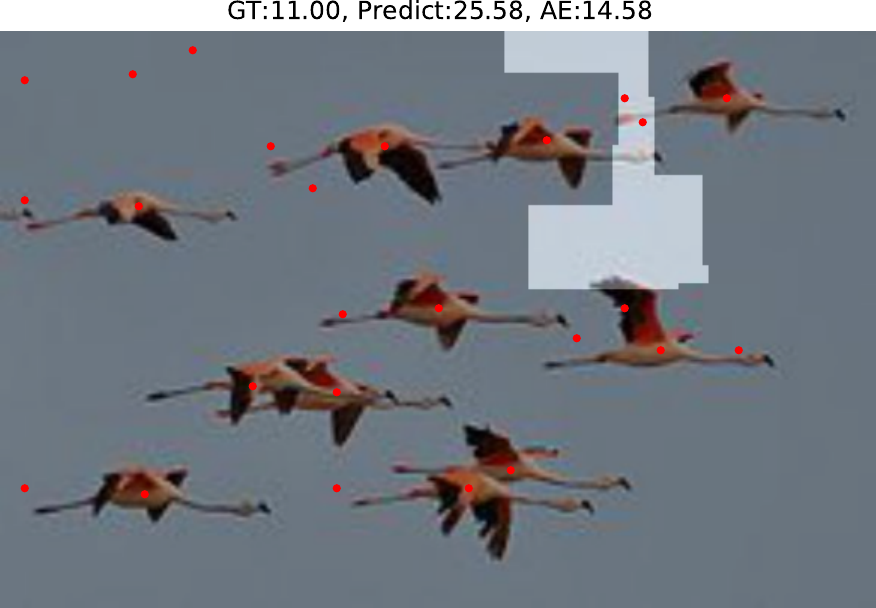} \hfill 
\includegraphics[width=\subFigSz]{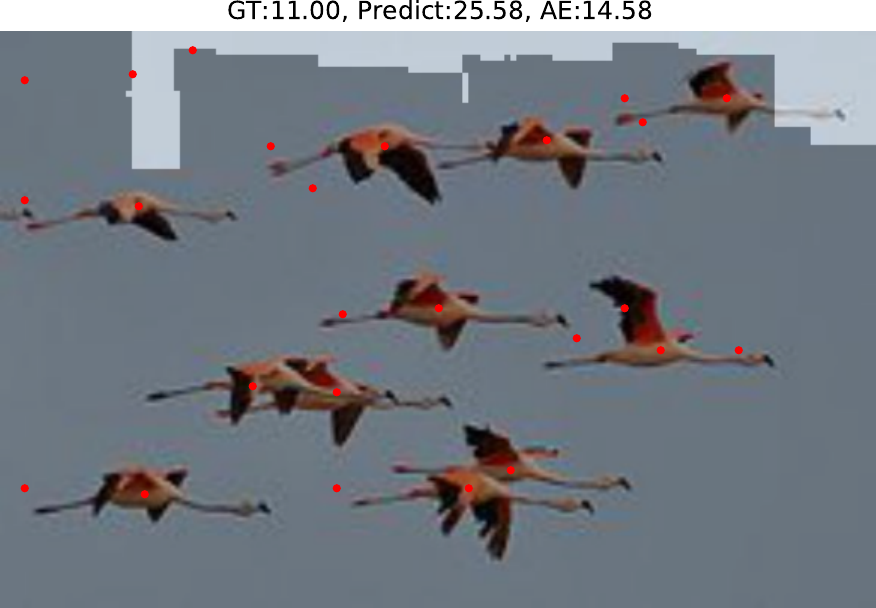} \hfill 
\includegraphics[width=\subFigSz]{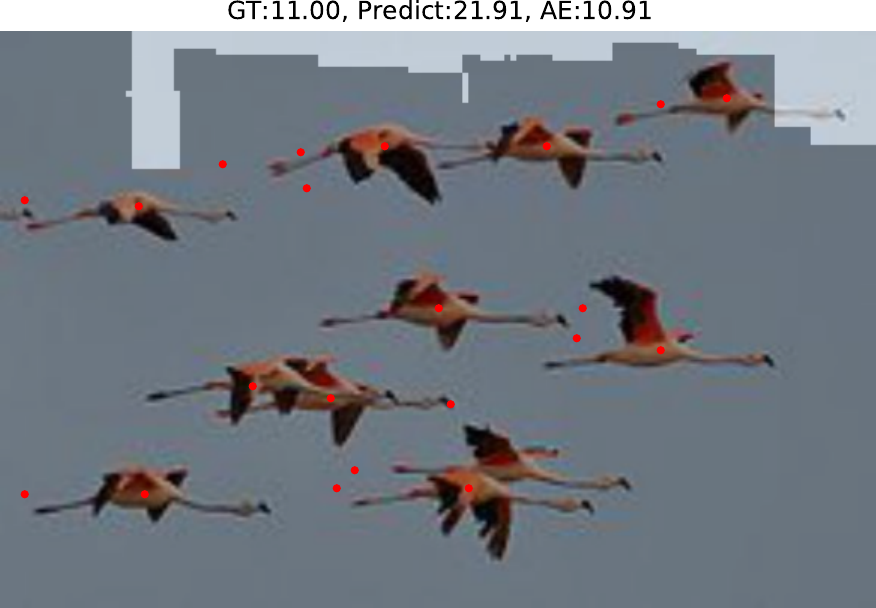}\\ 
\vskip 0.05in
\includegraphics[width=\subFigSz]{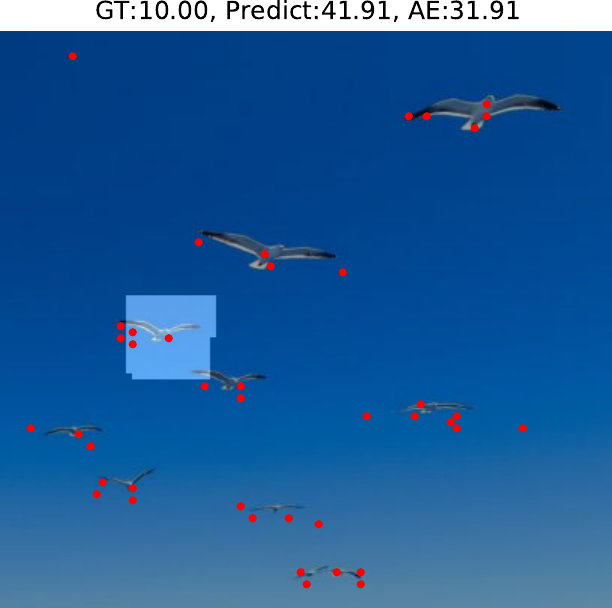} \hfill 
\includegraphics[width=\subFigSz]{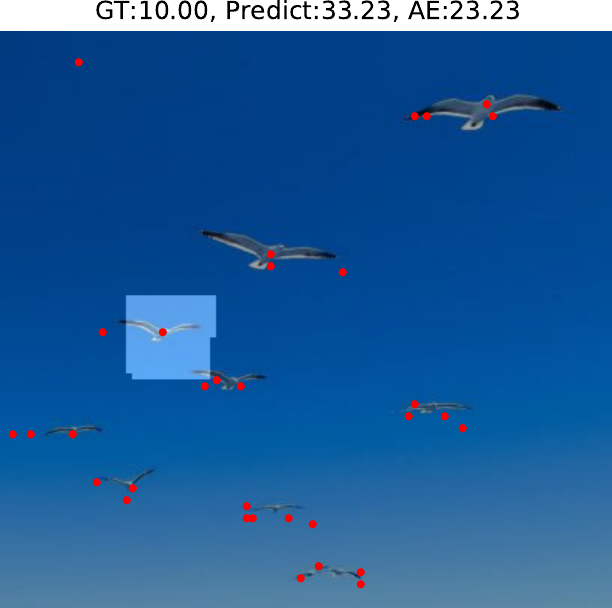} \hfill 
\includegraphics[width=\subFigSz]{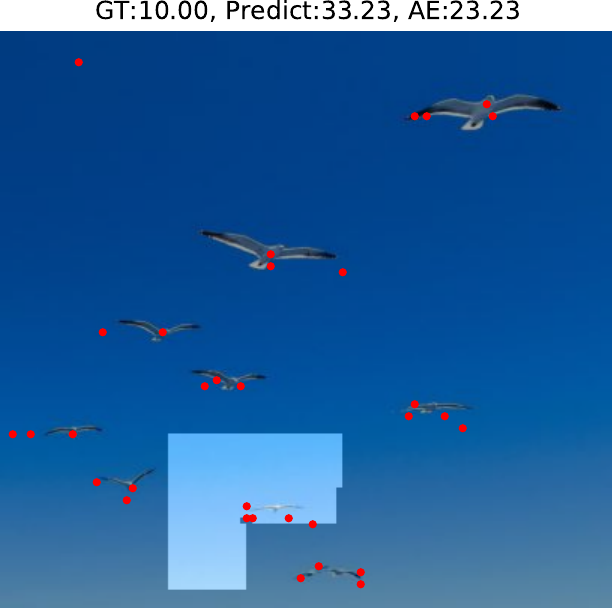} \hfill 
\includegraphics[width=\subFigSz]{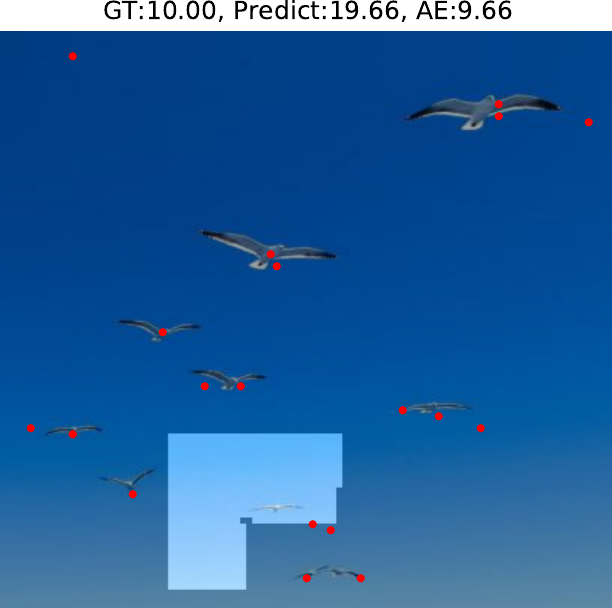}\\ 
\vskip 0.05in
\includegraphics[width=\subFigSz]{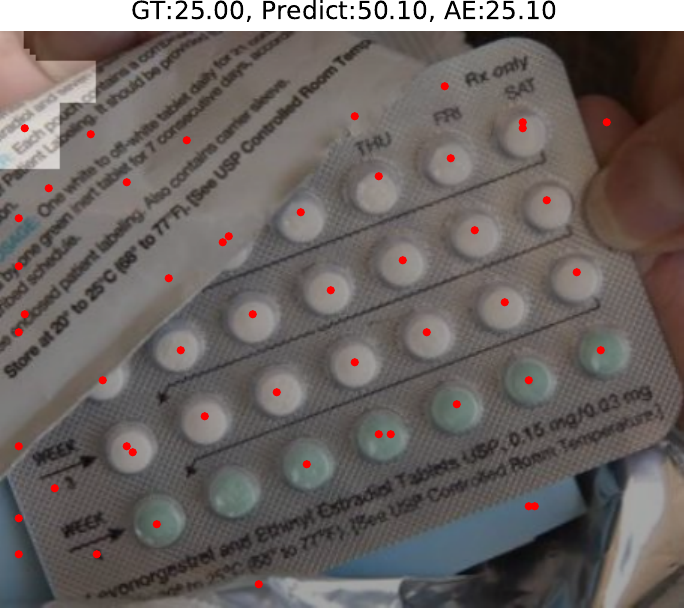} \hfill 
\includegraphics[width=\subFigSz]{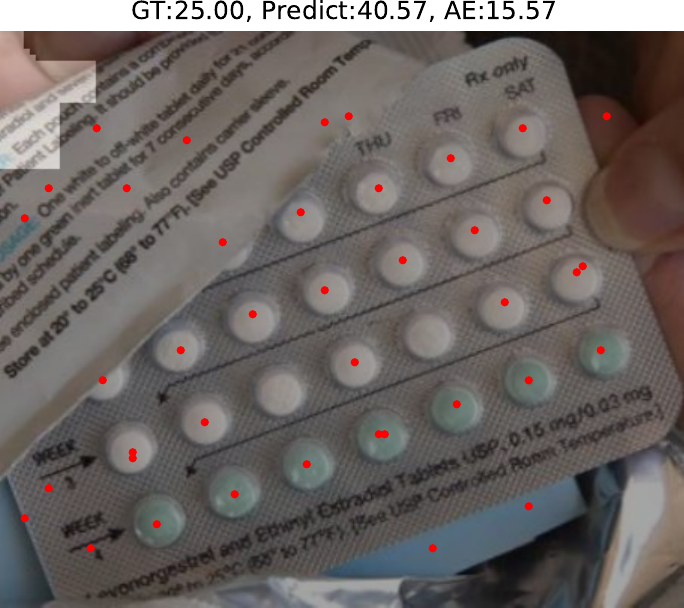} \hfill 
\includegraphics[width=\subFigSz]{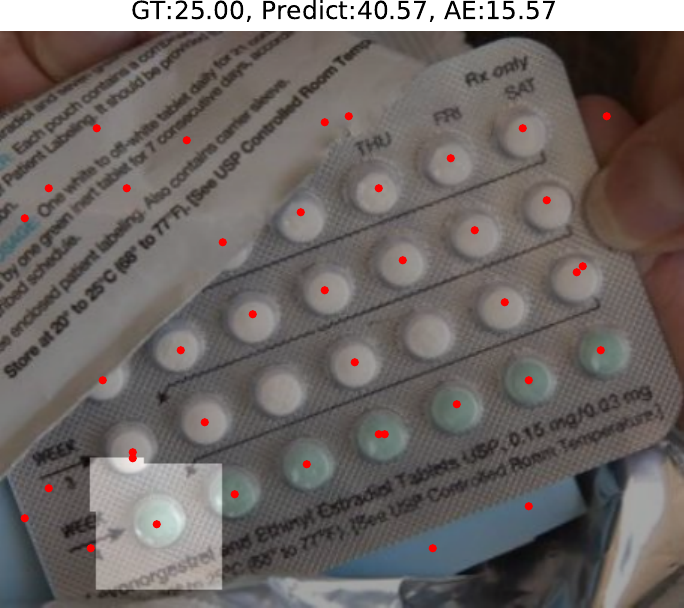} \hfill 
\includegraphics[width=\subFigSz]{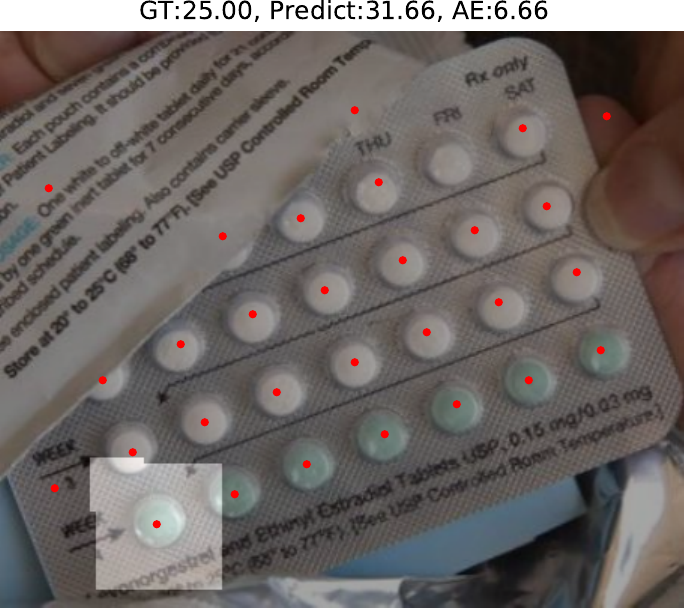}\\ 
\caption{Additional qualitative results. The examples are from FSC-147 with FamNet as the visual counter. }
\label{fig:ADDITIONAL_QUAL2}
\end{figure*}

{\small
\bibliographystyle{ieee_fullname}
\bibliography{longstrings,m_pubs_autogen,egbib}
}

\end{document}